\documentclass[
11pt,              % was 12 originally
letterpaper,       % legalpaper, a4paper, b5paper, ..., landscape
fleqn,
headnosepline,	   % headsepline line separating header from text body
oneside,           % twoside
onecolumn,         % twocolumn
openright,         % openany (new chapter w/any page rather than right)
noonelinecaption,
cleardoubleempty,
smallheadings]
{scrartcl}         % {scrbook}, {scrreprt}, {scrlttr2}
\addtokomafont{caption}{\sffamily\footnotesize}
\setkomafont{pagenumber}{\sffamily\footnotesize}
\addtokomafont{pagenumber}{\sffamily\footnotesize}
\setkomafont{captionlabel}{\sffamily\bfseries\footnotesize}
\setcapindent{0em} % no indentation

\usepackage{graphicx}
\usepackage{psfrag}
\usepackage{amsmath}
\usepackage{amssymb}
\usepackage{color}
\usepackage{xcolor}
\usepackage{rotate}
\usepackage{makeidx}
\usepackage{cite}
\usepackage{eufrak}
\usepackage[varumlaut]{yfonts}
\usepackage{helvet}   % Helvetica font as default sans serif
\usepackage{mathpazo}
\normalfont
\usepackage[T1]{fontenc}
\usepackage{textcomp}
\usepackage{longtable} 
\usepackage[only,llbracket,rrbracket]{stmaryrd}
\usepackage{bm}
\begin{document}
%%%%%%%%%%%%%%%%%%%%%%%%%%%%%%%%%%%%%%%%%%%%%%%%%%%%%%%%%%%%%%%%%%%%%%%%
\newcommand{\beq} {\begin{equation}}
\newcommand{\eeq} {\end{equation}}
\newcommand{\D}   {\displaystyle}
\newcommand{\divg}{\mbox{\rm{div}}\,}
\newcommand{\clearemptydoublepage}{\newpage{\pagestyle{empty}\cleardoublepage}}
\newcommand{\Divg}{\mbox{\rm{Div}}\,}
%%%%%%%%%%%%%%%%%%%%%%%%%%%%%%%%%%%%%%%%%%%%%%%%%%%%%%%%%%%%%%%%%%%%%%%%
\newtheorem{remark}      {\bf{\sffamily{Remark}}}
\newtheorem{definition}  {\bf{\sffamily{Definition}}}
%\numberwithin{remark}    {section}
%\numberwithin{definition}{chapter}
%\numberwithin{equation}  {section} 
%\renewcommand\figurename {\bf{\sffamily\footnotesize{Figure}}}
%\renewcommand\tablename  {\bf{\sffamily\footnotesize{Table}}}
\renewcommand{\sc}{}
\renewcommand{\Psi}{\psi}
\renewcommand{\varrho}{\vartheta}
\renewcommand{\arraystretch}{1.3}
\sloppy
%%%%%%%%%%%%%%%%%%%%%%%%%%%%%%%%%%%%%%%%%%%%%%%%%%%%%%%%%%%%%%%%%%%%%%%%
\def\sca   #1{\mbox{\rm #1}{}}
\def\mat   #1{\mbox{\bf #1}{}}
\def\vec   #1{\mbox{\boldmath $#1$}{}}
\def\ten   #1{\mbox{\boldmath $#1$}{}}
\def\scas  #1{\mbox{{\scriptsize{${\rm{#1}}$}}}{}}
\def\vecs  #1{\mbox{{\boldmath{\scriptsize{$#1$}}}}{}}
\def\tens  #1{\mbox{{\boldmath{\scriptsize{$#1$}}}}{}}
\def\up    #1{^{\mbox{\rm{\footnotesize{#1}}}}}
\def\down  #1{_{\mbox{\rm{\footnotesize{#1}}}}}
\def\ltr   #1{\mbox{\sf{#1}}}
\def\bltr  #1{\mbox{\sffamily{\bfseries{{#1}}}}}
%%%%%%%%%%%%%%%%%%%%%%%%%%%%%%%%%%%%%%%%%%%%%%%%%%%%%%%%%%%%%%%%%%%%%%%%%
\vspace*{0.6cm}
\begin{center}
{\sffamily\bfseries\Large{A new family of Constitutive Artificial Neural Networks}}\\[4pt]
{\sffamily\bfseries\Large{towards automated model discovery}}\\
%\end{center}
%%%%%%%%%%%%%%%%%%%%%%%%%%%%%%%%%%%%%%%%%%%%%%%%%%%%%%%%%%%%%%%%%%%%%%%%%
%%%%%%%%%%%%%%%%%%%%%%%%%%%%%%%%%%%%%%%%%%%%%%%%%%%%%%%%%%%%%%%%%%%%%%%%
\vspace*{0.6cm}
%\begin{center}
Kevin Linka \& Ellen Kuhl\\
Department of Mechanical Engineering \\
Stanford University, Stanford, California, United States \\
\vspace*{0.6cm}
{\small{\it{
We dedicate this manuscript to our Continuum Mechanics teachers\\
whose insights and passion for Continuum Mechanics have stimulated the ideas of this work,\\
Wolfgang Ehlers,
Mikhail Itskov,
Christian Miehe,
Michael Ortiz, \\
J\"org Schr\"oder,
Erwin Stein, and
Paul Steinmann}}}
\end{center}
\vspace*{0.4cm}
%%%%%%%%%%%%%%%%%%%%%%%%%%%%%%%%%%%%%%%%%%%%%%%%%%%%%%%%%%%%%%%%%%%%%%%%
{\sffamily{\bfseries{Abstract.}}}
%%%%%%%%%%%%%%%%%%%%%%%%%%%%%%%%%%%%%%%%%%%%%%%%%%%%%%%%%%%%%%%%%%%%%%%%
For more than 100 years, chemical, physical, and material scientists have proposed competing constitutive models to best characterize the behavior of natural and man-made materials in response to mechanical loading. 
%%%
Now, computer science offers a universal solution: Neural Networks. 
%%%
Neural Networks are powerful function approximators that can learn constitutive relations from large data without any knowledge of the underlying physics. 
%%%
However, classical Neural Networks entirely ignore a century of research in constitutive modeling, violate thermodynamic considerations, and fail to predict the behavior outside the training regime.
%%%
Here we design a new family of Constitutive Artificial Neural Networks that inherently satisfy common kinematical, thermodynamical, and physical constraints and, at the same time, constrain the design space of admissible functions to create robust approximators, even in the presence of sparse data. 
%%%
Towards this goal we revisit the non-linear field theories of mechanics and reverse-engineer the network input to account for material objectivity, material symmetry and incompressibility; 
the network output to enforce thermodynamic consistency; 
the activation functions to implement physically reasonable restrictions; and 
the network architecture to ensure polyconvexity. 
%%%
We demonstrate that this new class of models is a generalization of the classical neo Hooke, Blatz Ko, Mooney Rivlin, Yeoh, and Demiray models and that the network weights have a clear physical interpretation in the form of shear moduli, stiffness-like parameters, and exponential coefficients. 
%%%
When trained with classical benchmark data for rubber under uniaxial tension, biaxial extension, and pure shear, our network autonomously selects the best constitutive model and learns its set of parameters. 
%%%
Our findings suggests that Constitutive Artificial Neural Networks have the potential to induce a paradigm shift in constitutive modeling, from user-defined model selection to automated model discovery. 
%%%
Our source code, data, and examples are available 
%upon request. 
at 
https:/\!/github.com/LivingMatterLab/CANN.

\vspace*{0.5cm}

\clearpage
%%%%%%%%%%%%%%%%%%%%%%%%%%%%%%%%%%%%%%%%%%%%%%%%%%%%%%%%%%%%%%%%%%%
\section{Motivation}
%%%%%%%%%%%%%%%%%%%%%%%%%%%%%%%%%%%%%%%%%%%%%%%%%%%%%%%%%%%%%%%%%%%
\noindent
{\it{``What can your Neural Network tell you about the underlying physics?''}} is the most common question when we apply Neural Networks to study the behavior of materials and {\it{``Nothing.''}}\,is\,the\,honest and disappointing answer. \\[6.pt]
%%%
This manuscript challenges the notion that Neural Networks can teach us nothing about the  physics of a material. It seeks to integrate more than a century of knowledge in continuum mechanics \cite{antman05,ball77,holzapfel00book,ogden72,planck97,spencer71,truesdellnoll65,truesdell69} and modern machine learning \cite{karniadakis21,linka22,raissi19} to create a new family of Constitutive Artificial Neural Networks that inherently satisfy kinematical, thermodynamical, and physical constraints, and constrain the space of admissible functions to train robustly, even when data are space.
While this general idea is by no means new and builds on several important recent discoveries \cite{asad22,klein22,linka21,masi21}, the true novelty of our Constitutive Artificial Neural Networks is that they {\it{autonomously discover a constitutive model}}, and, at the same time, {\it{learn a set of physically meaningful  parameters}} associated with~it. \\[6.pt]
%%%
Interestingly, the first Neural Network for constitutive modeling approximates the incremental principal strains in concrete from known principal strains, stresses, and stress increments and is more than three decades old \cite{ghaboussi91}. In the early days, Neural Networks served merely as regression operators and were commonly viewed as a black box. The lack of transparency is probably the main reason why these early approaches never really generated momentum in the constitutive modeling community. More than 20 years later, data-driven constitutive modeling gained new traction, in part powered by a new computing paradigm, which directly uses experimental data and bypasses constitutive modeling altogether \cite{kirchdoerfer16}. While data-driven elasticity builds on a transparent and rigorous mathematical foundation \cite{conti18}, it can also become fairly complex, especially when expanding the theory to anisotropic \cite{denli22} or history-dependent \cite{eggersmann19} materials. Rather than following this path and eliminate the constitutive model entirely, here we attempt to build our prior physical knowledge into the Neural Network and learn something about the constitutive response \cite{alber19}.\\[6.pt]
%%%
Two successful but fundamentally different strategies have emerged to integrate physical knowledge into network modeling, {\it{Physics-Informed Neural Networks}} that add physics equations as additional terms to the loss function \cite{karniadakis21} and {\it{Constitutive Artificial Neural Networks}} that explicitly modify the network input, output, and architecture to hardwire physical constraints into the network design \cite{linka21}. The former approach is more general and typically works well for incorporating ordinary \cite{linka22} or partial \cite{raissi19} differential equations, while the latter is specifically tailored towards constitutive equations \cite{linka22a}. In fact, one such  Neural Network, with strain invariants as input, free energy functions as output, and a single hidden layer with logistic activation functions in between, has been proposed for rubber materials almost two decades ago \cite{shen04} and recently regained attention in the constitutive modeling community \cite{zopf17}. While these Constitutive Artificial Neural Networks generally provide excellent fits to experimental data \cite{blatz62,mooney40,treloar44}, exactly how they should integrate thermodynamic constraints remains a question of ongoing debate. \\[6.pt]
%%%
Thermodynamics-based Artificial Neural Networks a priori build the first and second law of thermodynamics into the network architecture and select specific activation functions to ensure compliance with thermodynamic constraints \cite{masi21}. Recent studies suggest that this approach can successfully reproduce the constitutive behavior of rubber-like materials \cite{ghaderi20}. Alternative approaches use a regular Artificial Neural Network and ensure thermodynamic consistency a posteriori via a pseudo-potential based correction in a post processing step \cite{kalina22}. To demonstrate the versatility of these different approaches, several recent studies have successfully embedded Neural Networks within a Finite Element Analysis, for example, to model plane rubber sheets \cite{linka21} or entire tires \cite{shen04}, the numerical homogenization of discrete lattice structures \cite{masi22}, the deployment of parachutes \cite{asad22}, or the anisotropic response of skin in reconstructive surgery \cite{tac22}. Regardless of all these success stories, one limitation remains: the lack of an intuitive interpretation of the model and its parameters \cite{klein22}. \\[6.pt]
%%%
The general idea of this manuscript is to reverse-engineer a new family of Constitutive Artificial Neural Networks that are, by design, a generalization of widely used and commonly accepted constitutive models \cite{blatz62,demiray72,mooney40,rivlin48,treloar48,yeoh93}
with well-defined physical parameters \cite{mahnken22,steinmann12}. Towards this goal,
we review the underlying kinematics in Section \ref{sec_kinematics} and discuss constitutive constraints in Section \ref{sec_const}. We then introduce classical Neural Networks in Section \ref{sec_NN} and our new family of Constitutive Artificial Neural Networks in Section \ref{sec_CANN}. In Section \ref{sec_homdef}, we briefly review the three special homogeneous deformation modes that we use to train our model in Section \ref{sec_results}. We discuss our results, limitations, and future directions in Section \ref{sec_discussion} and close with a brief conclusion in Section~\ref{sec_conclusion}.
%%%%%%%%%%%%%%%%%%%%%%%%%%%%%%%%%%%%%%%%%%%%%%%%%%%%%%%%%%%%%%%%%%%
%we revisit the non-linear field theories of mechanics \cite{truesdellnoll65,truesdell69}
%and constrain 
%the network {\it{output}} to enforce thermodynamic consistency;
%the network {\it{input}} to enforce material objectivity, and, if desired, material symmetry and incompressibility;
%the {\it{activation functions}} to implement physically reasonable constitutive restrictions; and the network {\it{architecture}} to ensure polyconvexity.
%We train the model with classical benchmark data of rubber and demonstrate its features compared to a classical Neural Network with no additional physical information.
%%%%%%%%%%%%%%%%%%%%%%%%%%%%%%%%%%%%%%%%%%%%%%%%%%%%%%%%%%%%%%%%%%%
\section{Kinematics}\label{sec_kinematics}
%%%%%%%%%%%%%%%%%%%%%%%%%%%%%%%%%%%%%%%%%%%%%%%%%%%%%%%%%%%%%%%%%%%
\noindent
We begin by characterizing the motion of a body and introduce the deformation map 
$\vec{\varphi}$ that, at any point in time $t$, maps material particles $\vec{X}$ from the undeformed configuration to particles, $\vec{x}=\vec{\varphi}(\vec{X},t)$, in the deformed configuration \cite{antman05}. To characterize relative deformations within the body, we introduce the deformation gradient $\ten{F}$, the gradient of the deformation map $\vec{\varphi}$ with respect to the undeformed coordinates $\vec{X}$, and its Jacobian $J$,
\beq
\ten{F} 
= \nabla_{\vecs{X}} \vec{\varphi}
\qquad \mbox{with} \qquad
J = \det (\ten{F}) > 0 \,.
\eeq
%The condition that the Jacobian has to remain strictly positive is associated with the {\it{principle of impenetrability of matter}} that ensures that the local ratio of the deformed and undeformed volumes can never be reduced to zero \cite{antman05}.
Multiplying $\ten{F}$ with its transpose $\ten{F}^{\scas{t}}$, either from the left or the right, introduces the right and left Cauchy Green deformation tensors $\ten{C}$ and $\ten{b}$,
\beq
  \ten{C} 
= \ten{F}^{\scas{t}} \cdot \ten{F}
\qquad \mbox{and} \qquad
  \ten{b} 
= \ten{F} \cdot \ten{F}^{\scas{t}} \,.
\eeq
In the undeformed state, all three tensors are identical to the unit tensor, $\ten{F}=\ten{I}$, $\ten{C}=\ten{I}$, and $\ten{b}=\ten{I}$, and the Jacobian is one, $J=1$. A Jacobian smaller than one, $0<J<1$, denotes compression and a Jacobian larger than one, $1<J$, denotes extension.\\[6.pt]
%%%%%%%%%%%%%%%%%%%%%%%%%%%%%%%%%%%%%%%%%%%%%%%%%%%%%%%%%%%%%%%%%%%
{\bf{\sffamily{Isotropy.}}} 
%%%%%%%%%%%%%%%%%%%%%%%%%%%%%%%%%%%%%%%%%%%%%%%%%%%%%%%%%%%%%%%%%%%
To characterize an {\it{isotropic}} material, 
we introduce the three principal invariants
$I_1$, $I_2$, $I_3$, either in terms of the deformation gradient $\ten{F}$,
\beq
\begin{array}{ @{\hspace*{0.0cm}}
              r@{\hspace*{0.2cm}}c@{\hspace*{0.2cm}}
              l@{\hspace*{0.6cm}}c@{\hspace*{0.6cm}}
              l@{\hspace*{0.05cm}}
              l@{\hspace*{0.2cm}}c@{\hspace*{0.2cm}}
              l@{\hspace*{0.6cm}}c@{\hspace*{0.6cm}}
              r@{\hspace*{0.2cm}}c@{\hspace*{0.2cm}}
              l@{\hspace*{0.5cm}}l@{\hspace*{0.05cm}}
              l@{\hspace*{0.2cm}}c@{\hspace*{0.2cm}}
              l@{\hspace*{0.5cm}}}
  I_1 &=& \ten{F} : \ten{F}
      & &\partial_{\tens{F}}  I_1&=&2\,\ten{F} \\
  I_2 &=& \frac{1}{2} \; [ I_1^2 - 
          [\, \ten{F}^{\scas{t}} \cdot \ten{F} \,] : 
          [\, \ten{F}^{\scas{t}} \cdot \ten{F} \,] ]
      &\mbox{with}&
          \partial_{\tens{F}}  I_2&=&2 \, [\, I_1 \, \ten{F}- \ten{F} \cdot \ten{F}^{\scas{t}} \cdot \ten{F} \,]\\
  I_3 &=& \mbox{det} \, (\ten{F}^{\scas{t}} \cdot \ten{F}) = J^2
      & &\partial_{\tens{F}}  I_3&=&2\, I_3 \, \ten{F}^{\scas{-t}} \,, \\
\end{array}
\label{invariants}
\eeq
or, equivalently, 
in terms of the right or left Cauchy Green deformation tensors 
$\ten{C}$ or  $\ten{b}$,
\beq
\begin{array}{ @{\hspace*{0.0cm}}
              r@{\hspace*{0.2cm}}c@{\hspace*{0.2cm}}
              l@{\hspace*{0.5cm}}l@{\hspace*{0.05cm}}
              l@{\hspace*{0.2cm}}c@{\hspace*{0.2cm}}
              l@{\hspace*{0.6cm}}c@{\hspace*{0.6cm}}
              r@{\hspace*{0.2cm}}c@{\hspace*{0.2cm}}
              l@{\hspace*{0.5cm}}l@{\hspace*{0.05cm}}
              l@{\hspace*{0.2cm}}c@{\hspace*{0.2cm}}
              l@{\hspace*{0.5cm}}}
  I_1 &=& \mbox{tr} \, (\ten{C}) = \ten{C} : \ten{I}
        &\partial_{\tens{C}}&   I_1&=&\ten{I}
&&I_1 &=& \mbox{tr} \, (\ten{b}) = \ten{b} : \ten{I}
        &\partial_{\tens{b}}&   I_1&=&\ten{I}\\
  I_2 &=& \frac{1}{2} \; [ I_1^2 - \ten{C}:\ten{C}]
        &\partial_{\tens{C}}&   I_2&=&I_1 \, \ten{I}-\ten{C}
      &\mbox{or}& 
  I_2 &=& \frac{1}{2} \; [ I_1^2 - \ten{b}:\ten{b}]
        &\partial_{\tens{b}}&   I_2&=&I_1 \, \ten{I}-\ten{b}\\
  I_3 &=& \mbox{det} \, (\ten{C}) = J^2
        &\partial_{\tens{C}}&   I_3&=&I_3 \, \ten{C}^{\scas{-t}}
&&I_3 &=& \mbox{det} \, (\ten{b}) = J^2
        &\partial_{\tens{b}}&   I_3&=&I_3 \, \ten{b}^{\scas{-t}} \,.\\
\end{array}
\eeq
In the undeformed state, $\ten{F}=\ten{I}$, and the three invariants are equal to three and one,
$I_1 = 3$, $I_2 =3$, and $I_3 = 1$.\\[6.pt]
%%%%%%%%%%%%%%%%%%%%%%%%%%%%%%%%%%%%%%%%%%%%%%%%%%%%%%%%%%%%%%%%%%%
{\bf{\sffamily{Near incompressibility.}}} 
%%%%%%%%%%%%%%%%%%%%%%%%%%%%%%%%%%%%%%%%%%%%%%%%%%%%%%%%%%%%%%%%%%%
To characterize an {\it{isotropic}}, {\it{nearly incompressible}} material, 
we perform a multiplicative decomposition of deformation gradient,
$\ten{F} = J^{1/3} \, \ten{I} \cdot \bar{\ten{F}}$,
into 
a volumetric part, $J^{1/3} \, \ten{I}$, and 
an isochoric part, $\bar{\ten{F}}$  \cite{flory61},
\beq
\bar{\ten{F}} = J^{-1/3} \ten{F} 
\qquad \mbox{and} \qquad
\bar{J} = \det(\bar{\ten{F}}) = 1 \,,
\eeq
and introduce the isochoric right and left Cauchy Green deformation tensors $\bar{\ten{C}}$ and $\bar{\ten{b}}$,
\beq
  \bar{\ten{C}} 
= \bar{\ten{F}}^{\scas{t}} \cdot \bar{\ten{F}}
= J^{-2/3} \, \ten{C}
  \qquad \mbox{and} \qquad
  \bar{\ten{b}} 
= \bar{\ten{F}} \cdot \bar{\ten{F}}^{\scas{t}}
= J^{-2/3} \, \ten{b}\,.
\eeq
We can then introduce an alternative set of invariants for nearly incompressible materials,
$\bar{I}_1$, $\bar{I}_2$, $J$,
in terms of the deformation gradient $\bar{\ten{F}}$,
\beq
\begin{array}{ @{\hspace*{0.0cm}}l@{\hspace*{0.2cm}}c@{\hspace*{0.2cm}}
              l@{\hspace*{0.3cm}}c@{\hspace*{0.3cm}}
              l@{\hspace*{0.2cm}}c@{\hspace*{0.2cm}}l}
  \bar{I}_1 &=& {I_1}/{J^{2/3}}   = \ten{F}:\ten{F} / {J^{2/3}} 
&&\partial_{\tens{F}} \bar{I}_1 
            &=& 2 / J^{2/3} \, \ten{F}  - \mbox{$\frac{2}{3}$} \, \bar{I}_1 \,\ten{F}^{\rm{-t}}   \\
  \bar{I}_2 &=& {I_2}/{J^{4/3}}   
             = \frac{1}{2} [ \bar{I}_1  - [\ten{F}^{\scas{t}} \cdot \ten{F}]:[\ten{F}^{\scas{t}} \cdot \ten{F}] / {J^{4/3}}]           
&\mbox{with}& 
  \partial_{\tens{F}} \bar{I}_2 
            &=& 2 / J^{2/3} \, \bar{I}_1 \, \ten{F}
             -  2 / J^{4/3} \ten{F} \cdot \ten{F}^{\rm{t}} \cdot \ten{F} 
             - \mbox{$\frac{4}{3}$} \, \bar{I}_2 \, \ten{F}^{\rm{-t}}\\
  J         &=& \det({\ten{F}}) 
&&\partial_{\tens{F}} J         
            &=& J \, \ten{F}^{\rm{-t}} \,, \\
\end{array}
\label{invariants_nearlyincomp}
\eeq
or, equivalently, in terms of the right and left Cauchy Green deformation tenors
$\ten{C}$ or $\ten{b}$,
\beq
\begin{array}{ @{\hspace*{0.0cm}}l@{\hspace*{0.2cm}}c@{\hspace*{0.2cm}}
              l@{\hspace*{0.6cm}}c@{\hspace*{0.6cm}}
              l@{\hspace*{0.2cm}}c@{\hspace*{0.2cm}}l}
  \bar{I}_1 &=& {I_1}/{J^{2/3}}   = \ten{C}:\ten{I} / {J^{2/3}} 
&&\bar{I}_1 &=& {I_1}/{J^{2/3}}   = \ten{b}:\ten{I} / {J^{2/3}}   \\
  \bar{I}_2 &=& {I_2}/{J^{4/3}}   
             = \frac{1}{2} [ \bar{I}_1  - \ten{C}:\ten{C} / {J^{4/3}}]
            & \mbox{or} &
  \bar{I}_2 &=& {I_2}/{J^{4/3}}   
             = \frac{1}{2} [ \bar{I}_1  - \ten{b}:\ten{b} / {J^{4/3}}] \\
  J         &=& \det^{1/2}({\ten{C}}) 
&&J         &=& \det^{1/2}({\ten{b}}) \,. \\
\end{array}
\eeq
%%%%%%%%%%%%%%%%%%%%%%%%%%%%%%%%%%%%%%%%%%%%%%%%%%%%%%%%%%%%%%%%%%%
{\bf{\sffamily{Perfect incompressibility.}}} 
%%%%%%%%%%%%%%%%%%%%%%%%%%%%%%%%%%%%%%%%%%%%%%%%%%%%%%%%%%%%%%%%%%%
To characterize an {\it{isotropic, perfectly incompressible}} material, we recall that the third invariant always remains identical to one, $I_3=J^2=1$. This implies that the principal and isochoric invariants are identical, $I_1 = \bar{I}_1$ and $I_2 = \bar{I}_2$, 
and that the set of invariants reduces to only these two. \\[6.pt]
%%%%%%%%%%%%%%%%%%%%%%%%%%%%%%%%%%%%%%%%%%%%%%%%%%%%%%%%%%%%%%%%%%%
{\bf{\sffamily{Transverse isotropy.}}} 
%%%%%%%%%%%%%%%%%%%%%%%%%%%%%%%%%%%%%%%%%%%%%%%%%%%%%%%%%%%%%%%%%%%
To characterize a {\it{transversely isotropic}} material 
with one pronounced direction with unit normal vector $\vec{n}$, we introduce a fourth invariant \cite{spencer71}, 
\beq        
  I_4  
= \vec{n} \cdot \ten{F}^{\scas{t}} \cdot \ten{F} \cdot \vec{n}
= \ten{C} : \ten{N} 
= \lambda_n^2
  \qquad \mbox{with} \qquad 
  \partial_{\tens{C}}   I_4
= \vec{n} \otimes \vec{n}
= \ten{N} \,.
\eeq
Here $\ten{N}= \vec{n} \otimes \vec{n}$ denotes the structural tensor associated with the pronounced direction $\vec{n}$,
with a unit length of $||\,\vec{n}\,||=1$ in the reference configuration
and a stretch of $\lambda_n = ||\,\ten{F} \cdot \vec{n} \,||$ in the deformed configuration. 
In the undeformed state, $\ten{F}=\ten{I}$, and the stretch and the fourth invariant are one, $\lambda_n=1$ and $I_4 = 1$.
%%%%%%%%%%%%%%%%%%%%%%%%%%%%%%%%%%%%%%%%%%%%%%%%%%%%%%%%%%%%%%%%%%%
\section{Constitutive equations}\label{sec_const}
%%%%%%%%%%%%%%%%%%%%%%%%%%%%%%%%%%%%%%%%%%%%%%%%%%%%%%%%%%%%%%%%%%%
%%%%%%%%%%%%%%%%%%%%%%%%%%%%%%%%%%%%%%%%%%%%%%%%%%%%%%%%%%%%%%%%%%%
% https://www.walter-fendt.de/html5/men/derivative12_en.htm
% https://www.brown.edu/Departments/Engineering/Courses/En221/Notes/Elasticity/Elasticity.htm
%%%%%%%%%%%%%%%%%%%%%%%%%%%%%%%%%%%%%%%%%%%%%%%%%%%%%%%%%%%%%%%%%%%
\noindent
In the most general form, constitutive equations in solid mechanics are tensor-valued tensor functions that define the relation between a stress, for example the Piola or nominal stress, 
$ \ten{P} 
= \rm{lim}_{{\scas{d}} \vecs{A} \rightarrow \vecs{0}} \, 
(\,\rm{d} \vec{f} / \rm{d}\vec{A} \,)$, as the force 
$ \rm{d}\vec{f}$ per undeformed area 
$ \rm{d}\vec{A}$, and a deformation measure, for example the deformation gradient $\ten{F}$ \cite{holzapfel00book,truesdellnoll65}, 
\beq
\ten{P} = \ten{P} (\ten{F}) \,.
\label{constitutive}
\eeq
Conceptually, we could use any Neural Network as a function approximator to simply learn the functional relation between $\ten{P}$ and $\ten{F}$ and many approaches in the literature actually do exactly that \cite{ghaboussi91,masi21,schulte22}. However, the functions $\ten{P}(\ten{F})$ that we learn through this approach might be too generic and violate well-known thermodynamical arguments and widely-accepted physical constraints \cite{ghaderi20}. Also, for limited amounts of data, the tensor-valued tensor function $\ten{P}(\ten{F})$ can be difficult to learn and there is a high risk of overfitting \cite{klein22}. Our objective is therefore to design a Constitutive Artificial Neural Network that a priori guarantees thermodynamic consistency of the function $\ten{P}(\ten{F})$, and, at the same time, conveniently limits the space of admissible functions to ensure robustness and prevent overfitting when available data are sparse. \\[6.pt]
%%%%%%%%%%%%%%%%%%%%%%%%%%%%%%%%%%%%%%%%%%%%%%%%%%%%%%%%%%%%%%%%%%%
{\bf{\sffamily{Thermodynamic consistency.}}} 
%%%%%%%%%%%%%%%%%%%%%%%%%%%%%%%%%%%%%%%%%%%%%%%%%%%%%%%%%%%%%%%%%%%
As a first step towards this goal, we ensure {\it{thermodynamically consistency}} and guarantee that the Piola stress $\ten{P}$ inherently satisfies the second law of thermodynamics, the entropy or Clausius-Duhem inequality \cite{planck97}, 
${\cal{D}} 
= \ten{P} : \dot{\ten{F}} - \dot{\psi} \ge 0$.
It states that, for any thermodynamic process, the total change in entropy, the dissipation ${\cal{D}}$,
should always remain greater than or equal to zero, ${\cal{D}}\ge 0$.
To a priori satisfy the dissipation inequality, we introduce the Helmholtz free energy as a function of the deformation gradient,
$\psi = \psi (\ten{F})$ such that
$\dot{\psi} = \partial \psi (\ten{F}) / \partial \ten{F} : \dot{\ten{F}}$,
and rewrite the dissipation inequality following the Coleman-Noll entropy principle  \cite{truesdellnoll65} as
${\cal{D}} 
= [\, \ten{P} - \partial \psi / \partial{\ten{F}} \,] : \dot{\ten{F}} \ge 0$.
For the {\it{hyperelastic}} case with ${\cal{D}} \doteq 0$, 
for all possible $\dot{\ten{F}}$,
the entropy inequality reduces to 
$\ten{P} - \partial \psi / \partial{\ten{F}} \doteq \ten{0}$.
The condition of thermodynamically consistency implies that the Piola stress $\ten{P}$ of a hyperelastic or Green-elastic material is a thermodynamically conjugate function of the deformation gradient $\ten{F}$ \cite{truesdell69}, 
\beq
  \ten{P}
= \frac{\partial \psi (\ten{F})}{\partial \ten{F}} \,.
\label{thermodynamic}
\eeq
For our Neural Network, this implies that, rather than approximating the nine  stress components $\ten{P}(\ten{F})$ as nine generic functions of the nine components of the deformation gradient $\ten{F}$, we train the network to learn the free energy function $\psi(\ten{F})$ and derive the stress $\ten{P}$ in a post-processing step to a priori satisfy the second law of thermodynamics. As such, satisfying thermodynamic consistency according to equation (\ref{thermodynamic}) directly affects the {\it{output}} of the Neural Network. \\[6.pt]
%%%%%%%%%%%%%%%%%%%%%%%%%%%%%%%%%%%%%%%%%%%%%%%%%%%%%%%%%%%%%%%%%%%
{\bf{\sffamily{Material objectivity and frame indifference.}}} 
%%%%%%%%%%%%%%%%%%%%%%%%%%%%%%%%%%%%%%%%%%%%%%%%%%%%%%%%%%%%%%%%%%%
Second, we further constrain the choice of the free energy function $\psi$ to satisfy {\it{material objectivity}} or {\it{frame indifference}} to ensure that the constitutive laws do not depend on the external frame of reference \cite{noll58}. Mathematically speaking, the constitutive equations have to be invariant under rigid body motions, 
$\psi(\ten{F}) = \psi(\ten{Q}\cdot\ten{F})$, for all proper orthogonal tensors $\ten{Q}\in\rm{SO}(3)$. The condition of objectivity implies that the stress response functions are independent of rotations and must be functions of the right Cauchy Green deformation tensor $\ten{C}$ \cite{truesdellnoll65}, 
\beq
  \ten{P}
= \frac{\partial \psi(\ten{C})}{\partial \ten{F}} 
= \frac{\partial \psi(\ten{C})}{\partial \ten{C}} : 
  \frac{\partial      \ten{C} }{\partial \ten{F}}
= 2 \, \ten{F} \cdot \frac{\partial \psi(\ten{C})}{\partial \ten{C}} \,.
\label{objectivity}
\eeq
For our Neural Network, this implies that rather than using the nine independent components of the deformation gradient $\ten{F}$ as input, we constrain the input to the six independent components of the symmetric right Cauchy Green deformation tensor,
$\ten{C} = \ten{F}^{\scas{t}} \cdot \ten{F}$. As such, satisfying material objectivity according to equation (\ref{objectivity}) directly affects the {\it{input}} of the Neural Network.\\[6.pt]
%%%%%%%%%%%%%%%%%%%%%%%%%%%%%%%%%%%%%%%%%%%%%%%%%%%%%%%%%%%%%%%%%%%
{\bf{\sffamily{Material symmetry and isotropy.}}} 
%%%%%%%%%%%%%%%%%%%%%%%%%%%%%%%%%%%%%%%%%%%%%%%%%%%%%%%%%%%%%%%%%%%
Third, we further constrain the choice of the free energy function $\psi$ to include constraints of {\it{material symmetry}}, which implies that the material response remains unchanged under transformations of the reference configuration,
$\psi(\ten{F}) = \psi(\ten{F} \cdot \ten{Q})$.
Here we consider the special case of {\it{isotropy}} for which the material response remains unchanged under proper orthogonal transformations of the reference configuration, 
$ \psi(\ten{F}^{\scas{t}} \cdot \ten{F}) 
= \psi(\ten{Q}^{\scas{t}} \cdot \ten{F}^{\scas{t}} \cdot \ten{F} \cdot \ten{Q})$, 
for {\it{all}} proper orthogonal tensors $\ten{Q}\in\rm{SO}(3)$ \cite{antman05}. 
The condition of isotropy implies that the stress response functions, 
$\psi(\ten{C})=\psi(\ten{b})$, must be functions of the left Cauchy Green deformation tensor, $\ten{b}=\ten{F} \cdot \ten{F}^{\scas{t}}$, 
and, together with the condition of objectivity,
$\psi(\ten{b}) = \psi(\ten{Q}^{\scas{t}} \cdot \ten{b} \cdot \ten{Q})$,
that the stress response functions must be functions of the 
{\it{invariants}} of $\ten{C}$ and $\ten{b}$, for example $\psi(I_1,I_2,I_3)$ using the set of invariants from equation~(\ref{invariants}). 
%In the most general form, we can expand the free energy as an infinite series in terms of the invariants \cite{mooney40},
%$ \psi (I_1,I_2,I_3)
%= \sum_{j,k,l=0}^{\infty} \,
%  a_{jkl}\,
%  [I_1-3]^j[I_2-3]^k[I_3-1]^l $
%where $a_{jkl}$ are material constants.
The Piola stress for hyperelastic isotropic materials then becomes
\beq
  \ten{P} 
= \frac{\partial \psi(I_1,I_2,I_3)}{\partial \ten{F}} 
= \frac{\partial \psi}{\partial I_1} \frac{\partial I_1}{\ten{F}}
+ \frac{\partial \psi}{\partial I_2} \frac{\partial I_2}{\ten{F}}
+ \frac{\partial \psi}{\partial I_3} \frac{\partial I_3}{\ten{F}}
= 2 \left[\frac{\partial \psi}{\partial I_1} 
    + I_1 \frac{\partial \psi}{\partial I_2} \right] \ten{F}
- 2 \frac{\partial \psi}{\partial I_2}
    \ten{F} \cdot \ten{F}^{\scas{t}} \cdot \ten{F} 
+ 2 I_3 \frac{\partial \psi}{\partial I_3} \ten{F}^{\scas{-t}} \,.
\label{Psi_of_I}
\eeq
For the case of {\it{near incompressibility}}, instead of using the invariants $I_1$, $I_2$, $I_3$, we can express the energy and stress as functions of the invariants $\bar{I}_1$, $\bar{I}_2$, $J$ from equation~(\ref{invariants_nearlyincomp}) \cite{holzapfel00book},
\beq
   \ten{P}
=  \frac{\partial \psi(\bar{I}_1,\bar{I}_2,J)}{\partial \ten{F}} 
=2 \frac{1}{J^{2/3}}
   \left[ \frac{\partial  \psi}{\partial \bar{I}_1} 
    + \bar{I}_1 \frac{\partial  \psi}{\partial \bar{I}_2} \right] \ten{F}
-2 \frac{1}{J^{4/3}}
   \frac{\partial  \psi}{\partial \bar{I}_2}
   \ten{F} \cdot \ten{F}^{\scas{t}} \cdot \ten{F}  
-  \frac{2}{3}
   \left[ \bar{I}_1 \frac{\partial  \psi}{\partial \bar{I}_1}
   + 2 \bar{I}_2 \frac{\partial  \psi}{\partial \bar{I}_2}  \right]
   \ten{F}^{\scas{-t}}
+\,J \frac{\partial  \psi}{\partial J} \ten{F}^{\scas{-t}} \,.
\label{Psi_of_Ibar}
\eeq
For our Neural Network, this implies that rather than using the six independent components of the symmetric right Cauchy Green deformation tensor
$\ten{C}$ as input, we constrain the input to a set of three invariants of the right and left Cauchy Green deformation tensors $\ten{C}$ and $\ten{b}$, 
either 
$I_1$, $I_2$, $I_3$ or
$\bar{I}_1$, $\bar{I}_2$, $J$.
In essence, considering materials with known symmetry classes according to equations (\ref{Psi_of_I}) or (\ref{Psi_of_Ibar}) directly affects, and ideally reduces, the {\it{input}} of the Neural Network.\\[6.pt]
%%%%%%%%%%%%%%%%%%%%%%%%%%%%%%%%%%%%%%%%%%%%%%%%%%%%%%%%%%%%%%%%%%%
{\bf{\sffamily{Incompressibility.}}} 
%%%%%%%%%%%%%%%%%%%%%%%%%%%%%%%%%%%%%%%%%%%%%%%%%%%%%%%%%%%%%%%%%%%
Fourth, we can further constrain the choice of the free energy function $\psi$ for the special case of {\it{perfect incompressibility}} for which the Jacobian remains one, $I_3 = J^2 = 1$.
The condition of perfect incompressibility implies that
equations (\ref{Psi_of_I}) and (\ref{Psi_of_Ibar}) 
simplify to an expression in terms of ony the first two invariants $I_1$ and $I_2$,
\beq
  \ten{P} 
= \frac{\partial \psi}{\partial I_1} \frac{\partial I_1}{\ten{F}}
+ \frac{\partial \psi}{\partial I_2} \frac{\partial I_2}{\ten{F}}
=2 \left[\frac{\partial \psi}{\partial I_1} 
   + I_1 \frac{\partial \psi}{\partial I_2} \right] \ten{F}
-2 \frac{\partial \psi}{\partial I_2}
   \ten{F} \cdot \ten{F}^{\scas{t}} \cdot \ten{F} .
\label{incompressibility}   
\eeq
For our Neural Network, this implies that rather than using the set of three invariants of the right and left Cauchy Green deformation tensors, either 
$I_1$, $I_2$, $I_3$ or
$\bar{I}_1$, $\bar{I}_2$, $J$ as input, 
we reduce the input to a set of only two invariants, $I_1$ and $I_2$.
Considering materials with perfect incompressibility according to equation (\ref{incompressibility}) further reduces the {\it{input}} of the Neural Network.
\\[6.pt]
%%%%%%%%%%%%%%%%%%%%%%%%%%%%%%%%%%%%%%%%%%%%%%%%%%%%%%%%%%%%%%%%%%%
{\bf{\sffamily{Physically reasonable constitutive restrictions.}}} 
%%%%%%%%%%%%%%%%%%%%%%%%%%%%%%%%%%%%%%%%%%%%%%%%%%%%%%%%%%%%%%%%%%%
Fifth, in addition to systematically reducing the parameterization of the free energy $\psi$ from the nine components of the non-symmetric deformation gradient $\ten{F}$, via the six components of the symmetric right Cauchy Green deformation tensor $\ten{C}$, to three or even two scalar-valued invariants $I_1$, $I_2$, $I_3$ and possibly $I_1$, $I_2$, we can restrict the functional form of the free energy $\psi$ by including additional constitutive restrictions that are both physically reasonable and mathematically convenient \cite{antman05}:\\[6.pt]
(i) The free energy $\psi$ is {\it{non-negative}} for all deformation states,
\beq
\psi(\ten{F}) \ge 0 
\quad \forall \quad
\ten{F}\,.
\label{physics_all}
\eeq
(ii) The free energy $\psi$ is {\it{zero}} in the reference configuration, also known as the {\it{growth condition}}, and it a priori ensures a {\it{stress-free reference configuration}},
\beq
\psi    (\ten{F}) \doteq 0
\quad \mbox{for} \quad
\ten{P} (\ten{F}) \doteq \ten{0}
\quad \mbox{at} \quad
\ten{F}=\ten{I}\,.
\label{physics_zero}
\eeq 
(iii) The free energy $\psi$ is {\it{infinite}} at the extreme states of  infinite compression, $J \rightarrow 0$, and infinite expansion, $J \rightarrow \infty$, 
\beq
\psi(\ten{F}) \rightarrow \infty 
\quad \mbox{for} \quad
J \rightarrow 0
\quad \mbox{or} \quad
J \rightarrow \infty \,.
\label{physics_infty}
\eeq
%%%
In addition, it seems reasonable to require that an increase in a component of the strain should be accompanied by an increase in the corresponding component of the stress and that extreme deformations for which an eigenvalue of the strain is zero or infinite should result in infinite stresses. 
For our Neural Network, to facilitate a stress-free reference configuration according to equation (\ref{physics_zero}), instead of using the invariants $I_1$, $I_2$, $I_3$ themselves as input, we use their deviation from the energy- and stress-free reference state, $[\,I_1-3\,]$, $[\,I_2-3\,]$, $[\,I_3-1\,]$, as input. In addition, from all possible activation functions, we select functional forms that comply with conditions (i), (ii), and (iii). 
As such, satisfying physical considerations according to equations (\ref{physics_all}), (\ref{physics_zero}), and (\ref{physics_infty}) directly affects the {\it{activation functions}} of the Neural Network, especially those between the last hidden layer and the output layer.\\[6.pt]
%%%%%%%%%%%%%%%%%%%%%%%%%%%%%%%%%%%%%%%%%%%%%%%%%%%%%%%%%%%%%%%%%%%
{\bf{\sffamily{Polyconvexity.}}} 
%%%%%%%%%%%%%%%%%%%%%%%%%%%%%%%%%%%%%%%%%%%%%%%%%%%%%%%%%%%%%%%%%%%
Sixth, to guide the selection of the functional forms for the free energy function $\psi$, and ultimately the selection of appropriate activation functions for our Neural Network, we consider {\it{polyconvexity}} requirements \cite{ball77}. 
From the general representation theorem we know that in its most generic form, the free energy of an isotropic material can be expressed as an infinite series of products of powers of the invariants \cite{rivlin51},
$ \psi (I_1,I_2,I_3)
= \sum_{j,k,l=0}^{\infty} \,
  a_{jk}\,
  [I_1-3]^j[I_2-3]^k[I_3-1]^l$,
where $a_{jkl}$ are material constants.
Importantly, mixed products of convex functions are generally not convex, and it is easier to show that the sum of specific convex subfunction usually is \cite{hartmann03}.
This motivates a special subclass of free energy functions in which the free energy is the sum of three individual polyconvex subfunctions $\psi_1$, $\psi_2$, $\psi_3$, such that
$ \psi  (\ten{F})
= \psi_1(I_1)
+ \psi_2(I_2)
+ \psi_3(I_3)$,
is polyconvex by design and the stresses take the following form,
\beq
  \ten{P} 
= \frac{\partial \psi  }{\partial \ten{F}} 
= \frac{\partial \psi_1}{\partial I_1}
  \frac{\partial I_1}{\partial \ten{F}}
+ \frac{\partial \psi_2}{\partial I_2}
  \frac{\partial I_2}{\partial \ten{F}}
+ \frac{\partial \psi_3}{\partial I_3}
  \frac{\partial I_3}{\partial \ten{F}} \,. 
\label{polyconvexity}   
\eeq
Popular polyconvex subfunctions are the power functions,
$\psi_1(I_1) = [I_1^k - 3^k]^i$ and
$\psi_2(I_2) = [I_2^{3k/2} - (3\sqrt{3})^k]^i$ and
$\psi_3(I_3) = [I_3 -1]^k$,
the exponential functions,
$\psi_1(I_1) = {\rm{exp}}(\varphi_1(I_1)) - 1$ and
$\psi_2(I_2) = {\rm{exp}}(\varphi_2(I_2)) - 1$,
and the logarithmic function, % J^2 = I_3, J = sqrt(I_3)
$\psi_3(I_3) = I_3 - 2\,\rm{ln}((I_3)^{1/2}) + 4\, (\rm{ln}(I_3)^{1/2})^2$,
for {\it{non-negative coefficients}}, $i,k \ge 1$.
For our Neural Network, this implies that we can either select polyconvex activation functions from a set of algorithmically predefined activation functions~\cite{klein22} or custom-design our own activations functions from known polyconvex subfunctions $\psi_1$,$\psi_2$,$\psi_3$ \cite{asad22}. In addition, polyconvexity requirements suggest that we should carefully consider using a fully-connected network architecture, in which mixed products of the invariants $I_1$, $I_2$, $I_3$ emerge naturally. Rather, polyconvexity points towards network architectures in which the three inputs $I_1$, $I_2$, $I_3$ are decoupled and only combined additively when we collect the entries of last hidden layer into the free energy function, $\psi=\psi_1+\psi_2+\psi_3$. As such, satisfying polyconvexity, for example according to equation (\ref{polyconvexity}), generally enforces {\it{non-negative network weights}} \cite{asad22} and directly affects the {\it{architecture}} and connectedness of the Neural Network \cite{klein22}.
%%%%%%%%%%%%%%%%%%%%%%%%%%%%%%%%%%%%%%%%%%%%%%%%%%%%%%%%%%%%%%%%%%%%%%%%
\section{Classical Neural Networks}\label{sec_NN}
%%%%%%%%%%%%%%%%%%%%%%%%%%%%%%%%%%%%%%%%%%%%%%%%%%%%%%%%%%%%%%%%%%%%%%%%
\noindent
Classical Neural Networks are versatile function approximators that are capable of learning any nonlinear function~\cite{mcculloch43}. However, as we will see, conventional off-the-shelf Neural Networks may violate the conditions of thermodynamic consistency, material objectivity, material symmetry, incompressibility, constitutive restrictions, and polyconvexity. In this section, we briefly summarize the {\it{input}}, {\it{output}}, {\it{architecture}}, and {\it{activation functions}} of classical Neural Networks to then, in the following section, modify these four elements as we design a new family of Constitutive Artificial Neural Networks that a priori satisfy the fundamental laws of physics. 
%%%%%%%%%%%%%%%%%%%%%%%%%%%%%%%%%%%%%%%%%%%%%%%%%%%%%%%%%%%%%%%%%%%%%%%%
\begin{figure}[h]
\centering
\includegraphics[width=0.64\linewidth]{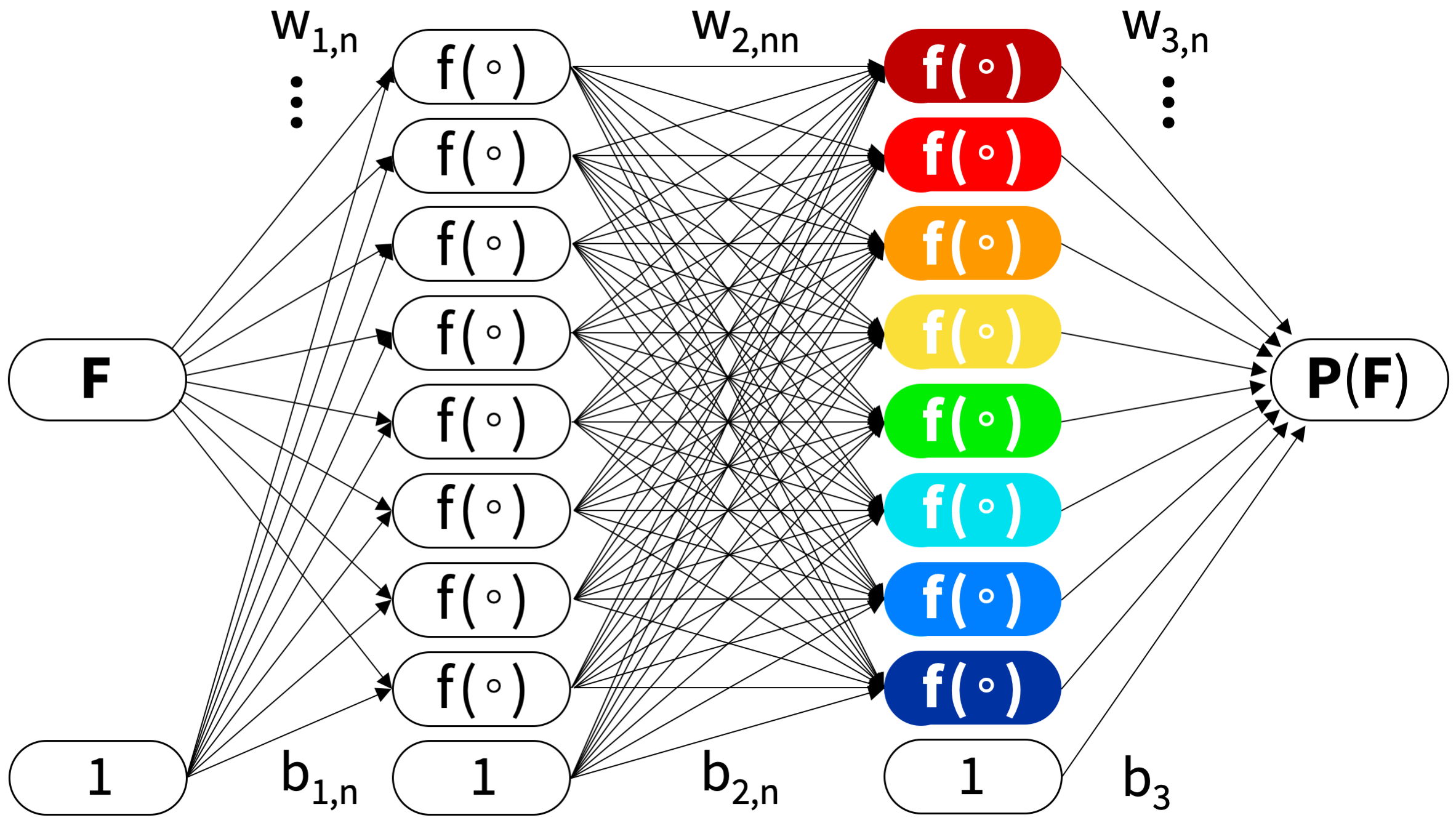}
\caption{{\bf{\sffamily{Classical Neural Network.}}} Example of a fully connected feed forward Neural Network with two hidden layers and eight nodes per layer to approximate the nine components of the tensor-valued Piola stress $\ten{P}(\ten{F})$ as a function of the nine components of the tensor-valued deformation gradient $\ten{F}$. The upper arrows originate from the network nodes and are associated with the weights ${\mat{w}}$, the lower arrows originate from the values one and are associated with the biases ${\mat{b}}$. The total number of arrows defines the number of network parameters we need to learn during the training process. The network in this example has $n_{\scas{w}} = 80$ weights, $n_{\scas{b}}=17$ biases, and a total number of $n_{\theta}=97$ parameters.}
\label{fig01}
\end{figure}\\[10.pt]
%%%%%%%%%%%%%%%%%%%%%%%%%%%%%%%%%%%%%%%%%%%%%%%%%%%%%%%%%%%%%%%%%%%%%%%%
%%%%%%%%%%%%%%%%%%%%%%%%%%%%%%%%%%%%%%%%%%%%%%%%%%%%%%%%%%%%%%%%%%%%%%%%
{\bf{\sffamily{Neural Network input and output.}}} 
%%%%%%%%%%%%%%%%%%%%%%%%%%%%%%%%%%%%%%%%%%%%%%%%%%%%%%%%%%%%%%%%%%%%%%%%
In constitutive modeling, we can use Neural Networks as universal function approximators to map a second order tensor, the deformation gradient $\ten{F}$ or any other strain measure, onto another second order tensor, the Piola stress $\ten{P}$ or any other stress measure, according to equation (\ref{constitutive}). Figure \ref{fig01} illustrates a classical Neural Network with the nine components of the deformation gradient $\ten{F}$ as input and the nine components of the nominal or Piola stress $\ten{P}$ as output. \\[6.pt]
%%%%%%%%%%%%%%%%%%%%%%%%%%%%%%%%%%%%%%%%%%%%%%%%%%%%%%%%%%%%%%%%%%%%%%%%
{\bf{\sffamily{Neural Network architecture.}}} 
%%%%%%%%%%%%%%%%%%%%%%%%%%%%%%%%%%%%%%%%%%%%%%%%%%%%%%%%%%%%%%%%%%%%%%%%
The architecture of the Neural Network determines how we approximate the relation between network input and output, in our case deformation gradient $\ten{F}$ and Piola stress \ten{P}. The simplest architecture is a {\it{feed forward}} Neural Network in which information moves only in one direction--forward--from the input nodes, without any cycles or loops, to the output nodes. Between input and output, the information passes through one or multiple {\it{hidden layers}}. Each hidden layer consists of multiple {\it{nodes or neurons}}. In the simplest case of a {\it{fully connected}} feed forward Neural Network, all nodes of a layer receive information from all nodes of the previous layer, each multiplied by an individual {\it{weight}}, all summed up and modulated by a {\it{bias}}.  
Figure \ref{fig01} illustrates the example of a fully connected feed forward Neural Network with an input layer composed of the deformation gradient $\ten{F}$, two hidden layers with eight nodes per layer, and an output layer composed of the Piola stress $\ten{P}$. Let us denote the input as $\mat{z}_0$, the nodal values of the hidden layer $k$ as $\mat{z}_{k}$, and the output as $\mat{z}_{k+1}$. For the example in Figure \ref{fig01} with two hidden layers, $k=1,2$, we calculate the values of each new layer from the  values of the previous layer according to the following set of equations,
\beq
\begin{array}{l@{\hspace*{0.2cm}}c@{\hspace*{0.2cm}}
              l@{\hspace*{0.2cm}}l@{\hspace*{0.2cm}}
              l@{\hspace*{0.2cm}}l}
\mat{z}_0 &=&      &                          &  &  \mat{F}       \\ 
\mat{z}_1 &=& f\,( & {\mat{w}_1} \, \mat{z}_0 &+ & {\mat{b}_1}\,)  \\ 
\mat{z}_2 &=& f\,( & {\mat{w}_2} \, \mat{z}_1 &+ & {\mat{b}_2}\,)  \\ 
\mat{z}_3 &=&      & {\mat{w}_3} \, \mat{z}_2 &+ & {\mat{b}_3} 
           \approx \mat{P} (\mat{F})\,.
\end{array}
\label{neuralnetwork}
\eeq
Here, 
${\mat{w}}$ are the set of network weights, 
${\mat{b}}$ are the network biases, and 
$f(\circ)$ are the activation functions.
In Figure \ref{fig01},
the upper arrows that originate from the nodes of the previous layer and are associated with the weights ${\mat{w}}$, the lower arrows that originate from the values one and are associated with the biases ${\mat{b}}$. The total number of arrows defines the number of network parameters we need to learn during the training process. 
For the fully connected feed forward Neural Network in Figure \ref{fig01}
with two hidden layers with eight nodes each, 
$\mat{w}_1 \in \mathbb{R}^{1 \times 8}$,
$\mat{w}_2 \in \mathbb{R}^{8 \times 8}$,
$\mat{w}_3 \in \mathbb{R}^{8 \times 1}$,
and 
$\mat{b}_1 \in \mathbb{R}^{8}$,
$\mat{b}_2 \in \mathbb{R}^{8}$,
$\mat{b}_3 \in \mathbb{R}^{1}$,
resulting in
$n_{\scas{w}} = 8+8\times8+8 = 80$ weights and 
$n_{\scas{b}} = 8+8+1 =17$ biases, 
and a total number of $n_{\theta}=97$ network parameters.\\[6.pt]
%%%%%%%%%%%%%%%%%%%%%%%%%%%%%%%%%%%%%%%%%%%%%%%%%%%%%%%%%%%%%%%%%%%
{\bf{\sffamily{Activation functions.}}} 
% https://www.walter-fendt.de/html5/men/derivative12_en.htm
%%%%%%%%%%%%%%%%%%%%%%%%%%%%%%%%%%%%%%%%%%%%%%%%%%%%%%%%%%%%%%%%%%%
Activation functions translate 
the sum of the weighted inputs to each node into 
an output signal that will be fed into the next layer \cite{mcculloch43}. 
In analogy to the brain that processes input signals and decides whether a neuron should fire or not \cite{budday15}, activation functions decide whether the nodal input is important or not in the process of approximating the final function, in our case the stress $\ten{P}(\ten{F})$. 
%%%%%%%%%%%%%%%%%%%%%%%%%%%%%%%%%%%%%%%%%%%%%%%%%%%%%%%%%%%%%%%%%%%%%%%%
\begin{figure}[h]
\centering
\includegraphics[width=0.72\linewidth]{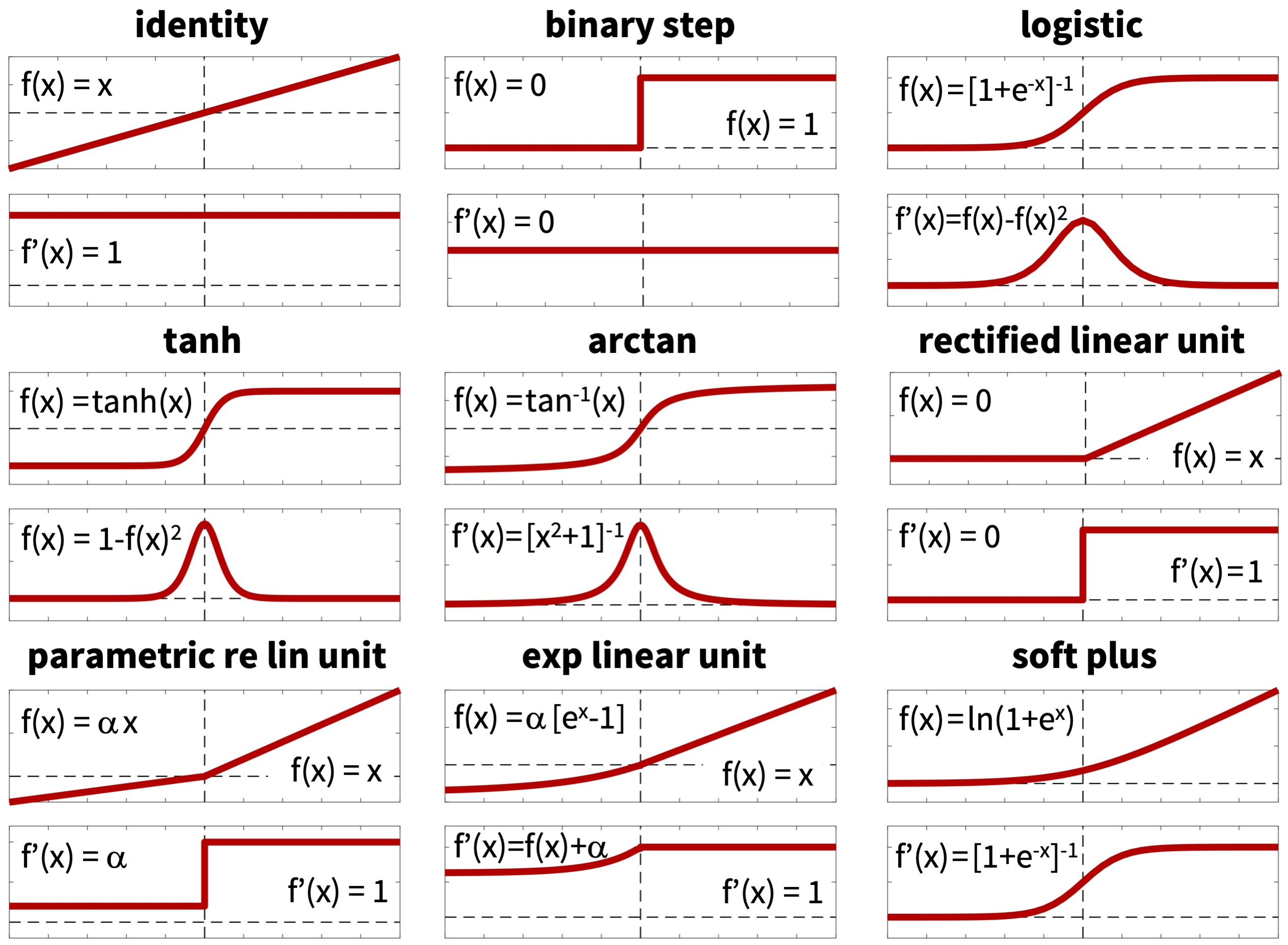}
\caption{{\bf{\sffamily{Activation functions for Classical Neural Networks.}}} 
Popular activation functions $f(x)$ along with their derivatives $f'(x)$ include the identity, binary step, logistic or soft step, hyperbolic tangent, inverse tangent, rectified linear unit or ReLU, parametric rectified linear unit or PReLU, exponential linear unit or ELU, and soft plus functions.
Activation functions can be continuous or discontinuous, linear or nonlinear, and bounded or unbounded.}
\label{fig02}
\end{figure}\\[6.pt]
%%%%%%%%%%%%%%%%%%%%%%%%%%%%%%%%%%%%%%%%%%%%%%%%%%%%%%%%%%%%%%%%%%%%%%%%
Figure \ref{fig02} illustrates the nine most popular activation functions $f(x)$ in Neural Network modeling along with their derivatives $f'(x)$. Depending on the final function we want to approximate, we can select from continuous or discontinuous, linear or nonlinear, and bounded or unbounded activation functions. In classical Neural Networks, all hidden layers typically use the same activation function, whereas the final output layer often uses a different activation function. 
%%%
For the simple example of a feed forward fully connected Neural Network similar to Figure \ref{fig01}, with
one input $z_0=F_{1}$, one output $z_3=P_{1}$,
and two hidden layers with two nodes per layer,
$\mat{z}_1 = [\,z_{11}, z_{12}\,]$ and
$\mat{z}_2 = [\,z_{21}, z_{22}\,]$,
the system of equations (\ref{neuralnetwork}) 
with activation functions of hyperbolic tangent type,
$f(x) = \rm{tanh}(x)$,
results in the following explicit expressions,
\beq
\begin{array}{l@{\hspace*{0.1cm}}c@{\hspace*{0.1cm}}c@{\hspace*{0.1cm}}l}
z_0    &=& F_{11} \\
z_{11} &=& \tanh (&  {w_{111}} \cdot  F_{11} + {b_{11}} \,) \\
z_{12} &=& \tanh (&  {w_{112}} \cdot  F_{11} + {b_{12}} \,) \\
z_{21} &=& \tanh (&  {w_{211}} \cdot z_{11} +  {w_{212}} \cdot z_{12} +  {b_{21}} \,) \\
z_{22} &=& \tanh (&  {w_{221}} \cdot z_{11} +  {w_{222}} \cdot z_{12} +  {b_{22}} \,) \\
z_{3}  &=&        &  {w_{321}} \cdot z_{21} +  {w_{322}} \cdot z_{22} +  {b_{31}}  \\
P_{11} &\approx&  & 
 {w_{321}} \cdot (
  \tanh ( {w_{211}} \cdot \tanh ( {w_{111}} \cdot  F_{11} +  {b_{11}})) 
+ \tanh ( {w_{212}} \cdot \tanh ( {w_{112}} \cdot  F_{11} +  {b_{12}})) +  {b_{21}}) \\
&&  + & 
 {w_{322}} \cdot (
  \tanh ( {w_{221}} \cdot \tanh ( {w_{111}} \cdot  F_{11} +  {b_{11}})) 
+ \tanh ( {w_{222}} \cdot \tanh ( {w_{112}} \cdot  F_{11} +  {b_{12}})) +  {b_{22}})+ {b_{31}}\, , \\
\end{array}
\label{neuralnetwork_tanh}
\eeq
where the output of the last layer $z_3$ approximates the true solution,
$P_{1} \approx z_3$. 
This specific Neural Network has
$\mat{w}_1 = [\,{w_{111}}, {w_{112}}\,]$,
$\mat{w}_2 = [\,{w_{211}}, {w_{212}}, {w_{221}}, {w_{222}}\,]$,
$\mat{w}_3 = [\,{w_{321}}, {w_{322}}\,]$,
and 
$\mat{b}_1 = [\,{b_{11}}, {b_{12}}\,]$,
$\mat{b}_2 = [\,{b_{21}}, {b_{22}}\,]$,
$\mat{b}_3 = [\,{b_{31}}\,]$,
resulting in
$n_{\scas{w}} = 2+2\times2+2 = 8$ weights and 
$n_{\scas{b}} = 2+2+1 =5$ biases, 
and a total number of $n_{\theta}=13$ network parameters. The set of equations (\ref{neuralnetwork_tanh}) illustrates that, for every hidden layer, we add one more level of nested activation functions, in this case $\rm{tanh}(\circ)$. The final approximated stress stretch relation $P_{1}(F_{1})$ is fairly complex, inherently nonlinear, and difficult if not impossible to invert explicitly. From the set of equations (\ref{neuralnetwork_tanh}), it is clear that the network weights and biases have {\it{no clear physical interpretation}}. \\[6.pt]
%%%
The selection of appropriate activation functions depends on the type of prediction we expect from our model. In constitutive modeling, where we seek to approximate the stress $\ten{P}$ as a function of the deformation gradient $\ten{F}$, we can immediately rule out some of the common activation functions in Figure \ref{fig02}--at least for the final output layer--when considering the physically reasonable constitutive restrictions (\ref{physics_all}), (\ref{physics_zero}), and (\ref{physics_infty}) from Section \ref{sec_const}:
(i) the binary step function is {\it{discontinuous}} at the origin, $f(-0)\ne f(+0)$, which violates our general understanding that the energy $\psi$ and the stress $\ten{P}$ should be smooth and continuous for all hyperelastic deformations;
(ii) the binary step function and rectified linear unit are {\it{constant}} over part or all of the domain, $f(x)=0$ or $f(x)=1$, which violates our general understanding that the stress $\ten{P}$ should not be constant, but rather increase with increasing deformation $\ten{F}$;
(iii) the binary step, logistic, hyperbolic tangent, and inverse tangent functions are {\it{horizontally asymptotic}}, $f(-\infty)=0$ and $f(+\infty)=1$, which violates the physically reasonable constitutive restriction (\ref{physics_infty}) that the energy and stress should not be bounded, but rather become infinite, $\ten{P} \rightarrow \infty$, for extreme deformations, $\ten{F} \rightarrow \infty$;
(iv) the rectified linear unit, parametric rectified linear unit, and exponential linear unit are continuous but {\it{non-differentiable}} at zero, $f'(-0)\neq f'(+0)$, which could be useful to model tension-compression asymmetry, but is not the most elegant choice to model the tension-compression transition at the origin.
At the same time, 
the identity, $f(x)=x$, and
the left branch of the exponential linear unit, $f(x) = \alpha \,[\rm{exp}(x) -1]$, 
remind us of the classical linear neo Hooke \cite{treloar48} and exponential Holzapfel \cite{holzapfel00} models. Together with the soft plus function, $f(x) = \rm{ln} (1+\rm{exp(x)})$, they are the only three functions that are {\it{continuous}}, {\it{differentiable}}, and {\it{polyconvex}} \cite{klein22}.
This motivates the question, can we identify existing activation functions or design our own set of activation functions that mimic known constitutive models, or contributions to them, and, ideally, satisfy polyconvexity requirements by design?\\[6.pt]
%%%%%%%%%%%%%%%%%%%%%%%%%%%%%%%%%%%%%%%%%%%%%%%%%%%%%%%%%%%%%%%%%%%
{\bf{\sffamily{Loss function.}}} 
%%%%%%%%%%%%%%%%%%%%%%%%%%%%%%%%%%%%%%%%%%%%%%%%%%%%%%%%%%%%%%%%%%%
The objective of a classical Neural Network is to learn the network parameters, 
$\vec{\theta}=\{ \mat{w}_k, \mat{b}_k \}\,$, the network weights and biases, by minimizing a loss function $L$ that penalizes the error between model and data. 
We commonly characterize this error as the mean squared error, %MSE, 
the $L_2$-norm of the difference between model 
$\ten{P}(\ten{F}_i)$ and data $\hat{\ten{P}}_i$, 
divided by the number of training points $n_{\rm{trn}}$,
\beq
  L (\vec{\theta} ; \ten{F})
= \frac{1}{n_{\rm{trn}}} \sum_{i=1}^{n_{\rm{trn}}}
|| \, \ten{P}(\ten{F}_i) - \hat{\ten{P}}_i \, ||^2 
\rightarrow \mbox{min}\,.
\label{loss_NN}
\eeq
We train the network by minimizing the loss function (\ref{loss_NN}) and learn the network parameters, $\vec{\theta}= \{ \mat{w}_k, \mat{b}_k \}$, in our case using the ADAM optimizer, a robust adaptive algorithm for gradient-based first-order optimization. 
With appropriate training data, classical Neural Networks can interpolate data well, without any prior knowledge of the underlying physics. However, they typically fail to extrapolate and make informed predictions \cite{alber19}. Since they usually have many degrees of freedom, they are inherently at risk of overfitting, especially if the available data are sparse \cite{peng21}. In addition, they may violate the thermodynamic restrictions of Section \ref{sec_const}. This motivates the question, can we integrate physical information we already know to constrain the function $\ten{P}(\ten{F})$, prevent overfitting, and make the model more predictive?
%%%%%%%%%%%%%%%%%%%%%%%%%%%%%%%%%%%%%%%%%%%%%%%%%%%%%%%%%%%%%%%%%%%%%%%%
\section{Constitutive Artificial Neural Networks}\label{sec_CANN}
%%%%%%%%%%%%%%%%%%%%%%%%%%%%%%%%%%%%%%%%%%%%%%%%%%%%%%%%%%%%%%%%%%%%%%%%
\noindent
We now propose a new family of Constitutive Artificial Neural Networks that satisfy the conditions of thermodynamic consistency, material objectivity, material symmetry, incompressibility, constitutive restrictions, and polyconvexity by design. In the following, we discuss how this guides our selection of network {\it{input}}, {\it{output}}, {\it{architecture}}, and {\it{activation functions}} to a priori satisfy the fundamental laws of physics. We also demonstrate that, for special cases, members of this family reduce to well-known constitutive models, including the neo Hooke \cite{treloar48}, Blatz Ko \cite{blatz62}, Mooney Rivlin \cite{mooney40,rivlin48}, Yeoh \cite{yeoh93}, Demiray \cite{demiray72} and Holzapfel \cite{holzapfel00} models, and that the network weights have a clear physical interpretation.
%%%%%%%%%%%%%%%%%%%%%%%%%%%%%%%%%%%%%%%%%%%%%%%%%%%%%%%%%%%%%%%%%%%%%%%%
\begin{figure}[h]
\centering
\includegraphics[width=0.72\linewidth]{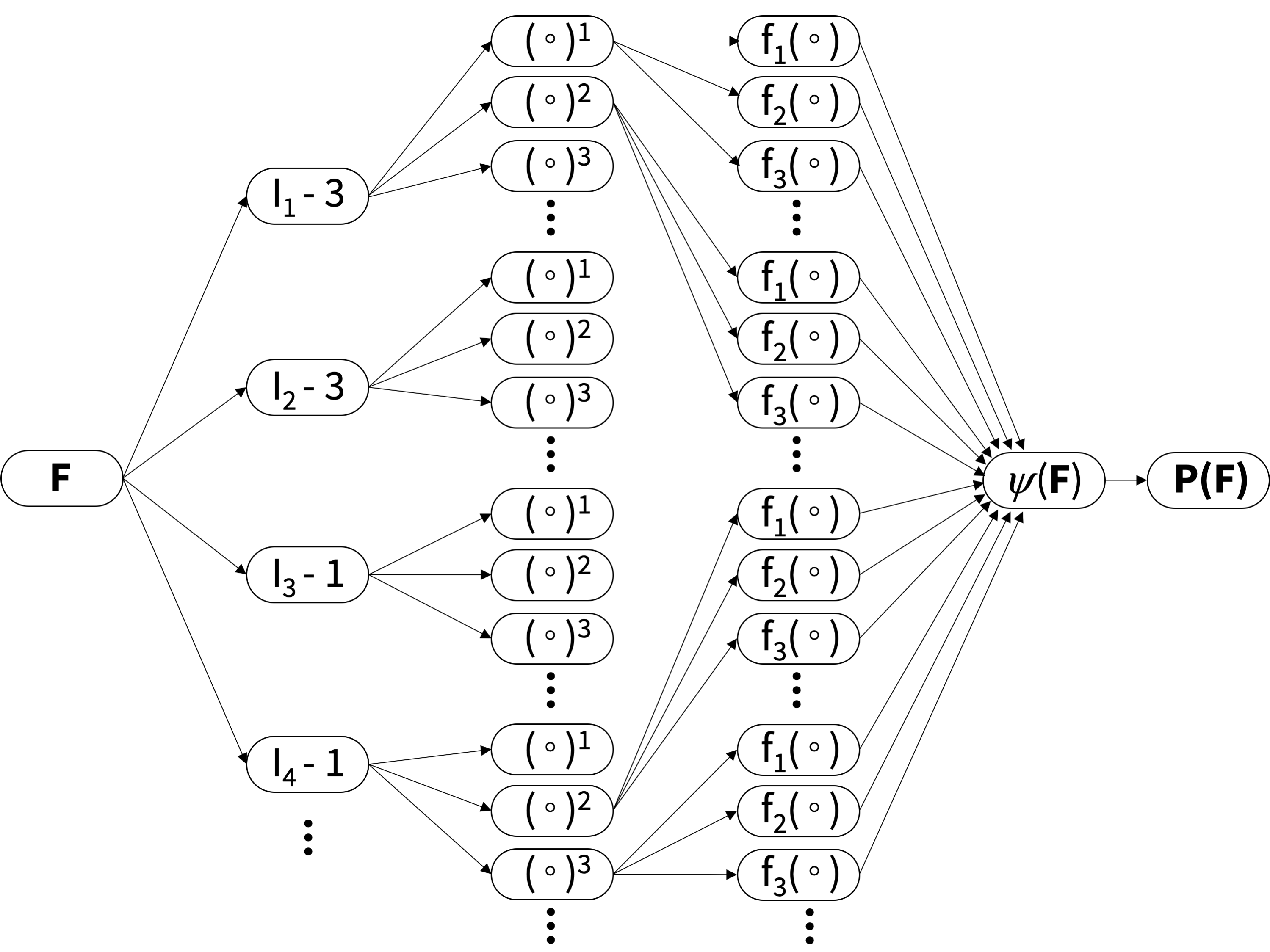}
\caption{{\bf{\sffamily{Constitutive Artificial Neural Network.}}} 
Family of a feed forward Constitutive Artificial Neural Networks with two hidden layers to approximate the single scalar-valued free energy function $\psi(I_1, I_2, I_3, I_4)$ as a function of the scalar-valued invariants $I_1, I_2, I_3, I_4$ of the deformation gradient $\ten{F}$. 
The first layer generates powers $(\circ)$, $(\circ)^2$, $(\circ)^3$ of the network input and the second layer applies thermodynamically admissible activation functions $f(\circ)$ to these powers. Constitutive Artificial Neural Networks are typically not fully connected by design to a priori satisfy the condition of polyconvexity.}
\label{fig03}
\end{figure}\\[6.pt]
%%%%%%%%%%%%%%%%%%%%%%%%%%%%%%%%%%%%%%%%%%%%%%%%%%%%%%%%%%%%%%%%%%%%%%%%
%%%%%%%%%%%%%%%%%%%%%%%%%%%%%%%%%%%%%%%%%%%%%%%%%%%%%%%%%%%%%%%%%%%%%%%%
\noindent{\bf{\sffamily{Constitutive Artificial Neural Network input and output.}}} 
%%%%%%%%%%%%%%%%%%%%%%%%%%%%%%%%%%%%%%%%%%%%%%%%%%%%%%%%%%%%%%%%%%%%%%%%
To ensure thermodynamical consistency, rather than directly approximating the stress $\ten{P}$ as a function of the deformation gradient $\ten{F}$,  
we use the Constitutive Artificial Neural Network as a universal function approximator to map a the scalar-valued invariants $I_1, I_2, I_3, I_4$ onto the scalar-valued free energy function $\psi$ according to equations (\ref{Psi_of_I}). The Piola stress $\ten{P}$ then follows naturally from the second law of thermodynamics as the derivative of the free energy $\psi$ with respect to the deformation gradient $\ten{F}$ according to equations (\ref{constitutive}) and (\ref{Psi_of_I}). Figure \ref{fig03} illustrates a Constitutive Artificial Neural Network with the invariants $I_1$, $I_2$, $I_3$, $I_4$ as input and the the free energy $\psi$ as output.  \\[6.pt]
%%%%%%%%%%%%%%%%%%%%%%%%%%%%%%%%%%%%%%%%%%%%%%%%%%%%%%%%%%%%%%%%%%%%%%%%
{\bf{\sffamily{Constitutive Artificial Neural Network architecture.}}} 
%%%%%%%%%%%%%%%%%%%%%%%%%%%%%%%%%%%%%%%%%%%%%%%%%%%%%%%%%%%%%%%%%%%%%%%%
Since we seek to model a hyperelastic history-independent material, we select a feed forward architecture in which information only moves in one direction, from the input nodes, without any cycles or loops, to the output nodes. 
To control polyconvexity, rather than choosing a fully connected feed forward network, we select a network architecture in which nodes only receive an input from selected nodes of the previous layer. Specifically, according to equation (\ref{polyconvexity}), the nodes of the individual invariants are not connected, such that the free energy function does not contain mixed terms in the invariants.
Figure \ref{fig03} illustrates one possible architecture that attempts to a priori satisfy the polyconvexity condition (\ref{polyconvexity}) by decoupling the information of the individual invariants.
For this particular network architecture, the free energy function that we seek to approximate takes the following format,
\beq
\begin{array}{l@{\hspace*{0.1cm}}l@{\hspace*{0.1cm}}l@{\hspace*{0.1cm}}
              l@{\hspace*{0.1cm}}l@{\hspace*{0.0cm}}l@{\hspace*{0.1cm}}
              l@{\hspace*{0.1cm}}l@{\hspace*{0.1cm}}l@{\hspace*{0.1cm}}
              l@{\hspace*{0.0cm}}c@{\hspace*{0.1cm}}l@{\hspace*{0.1cm}}
              l@{\hspace*{0.1cm}}l@{\hspace*{0.1cm}}l@{\hspace*{0.0cm}}l}
    \psi(I_1,I_2,I_3,I_4)
&=& w_{2,1}  &f_1 (w_{1,1}  & [\, I_1 -3 \,]^1&)
&+& w_{2,2}  &f_2 (w_{1,2}  & [\, I_1 -3 \,]^1&) 
&+& w_{2,3}  &f_3 (w_{1,3}  & [\, I_1 -3 \,]^1&) \\
&+& w_{2,4}  &f_1 (w_{1,4}  & [\, I_1 -3 \,]^2&)
&+& w_{2,5}  &f_2 (w_{1,5}  & [\, I_1 -3 \,]^2&) 
&+& w_{2,6}  &f_3 (w_{1,6}  & [\, I_1 -3 \,]^2&) \\
&+& w_{2,7}  &f_1 (w_{1,7}  & [\, I_1 -3 \,]^3&)
&+& w_{2,8}  &f_2 (w_{1,8}  & [\, I_1 -3 \,]^3&) 
&+& w_{2,9}  &f_3 (w_{1,9}  & [\, I_1 -3 \,]^3&) \\
&+& w_{2,10} &f_1 (w_{1,10} & [\, I_2 -3 \,]^1&)
&+& w_{2,11} &f_2 (w_{1,11} & [\, I_2 -3 \,]^1&) 
&+& w_{2,12} &f_3 (w_{1,12} & [\, I_2 -3 \,]^1&) \\
&+& w_{2,13} &f_1 (w_{1,13} & [\, I_2 -3 \,]^2&)
&+& w_{2,14} &f_2 (w_{1,14} & [\, I_2 -3 \,]^2&) 
&+& w_{2,15} &f_3 (w_{1,15} & [\, I_2 -3 \,]^2&) \\
&+& w_{2,16} &f_1 (w_{1,16} & [\, I_2 -3 \,]^3&)
&+& w_{2,17} &f_2 (w_{1,17} & [\, I_2 -3 \,]^3&) 
&+& w_{2,18} &f_3 (w_{1,18} & [\, I_2 -3 \,]^3&) \\
&+& w_{2,19} &f_1 (w_{1,19} & [\, I_3 -1 \,]^1&)
&+& w_{2,20} &f_2 (w_{1,20} & [\, I_3 -1 \,]^1&) 
&+& w_{2,21} &f_3 (w_{1,21} & [\, I_3 -1 \,]^1&) \\
&+& w_{2,22} &f_1 (w_{1,22} & [\, I_3 -1 \,]^2&)
&+& w_{2,23} &f_2 (w_{1,23} & [\, I_3 -1 \,]^2&) 
&+& w_{2,24} &f_3 (w_{1,24} & [\, I_3 -1 \,]^2&)\\
&+& w_{2,25} &f_1 (w_{1,25} & [\, I_3 -1 \,]^3&)
&+& w_{2,26} &f_2 (w_{1,26} & [\, I_3 -1 \,]^3&) 
&+& w_{2,27} &f_3 (w_{1,27} & [\, I_3 -1 \,]^3&) + ... \,.\\
\end{array}
\label{CANNenergy_generic}
\eeq
%%%%%%%%%%%%%%%%%%%%%%%%%%%%%%%%%%%%%%%%%%%%%%%%%%%%%%%%%%%%%%%%%%%%%%%%
This specific network has 
$4 \times 3 \times 3 + 4 \times 3 \times 3  = 72$ weights for the transversely isotropic case with all four invariants $I_1$, $I_2$, $I_3$, $I_4$ and 
$3 \times 3 \times 3 + 3 \times 3 \times 3  = 54$ weights for the isotropic case with only three invariants $I_1$, $I_2$, $I_3$. \\[6.pt]
%%%%%%%%%%%%%%%%%%%%%%%%%%%%%%%%%%%%%%%%%%%%%%%%%%%%%%%%%%%%%%%%%%%
{\bf{\sffamily{Activation functions.}}} 
% https://www.walter-fendt.de/html5/men/derivative12_en.htm
%%%%%%%%%%%%%%%%%%%%%%%%%%%%%%%%%%%%%%%%%%%%%%%%%%%%%%%%%%%%%%%%%%%
To ensure that our network satsifies basic physically reasonable constitutive restrictions, rather than selecting from the popular pre-defined activation functions in Figure \ref{fig02}, we custom-design our own activation functions to {\it{reverse-engineer}} a free energy function that captures popular forms of constitutive terms. Specifically, we select from linear, quadratic, cubic, and higher order powers for the first layer of the network, and from linear, exponential, or logarithmic functions for the second layer.  
%%%%%%%%%%%%%%%%%%%%%%%%%%%%%%%%%%%%%%%%%%%%%%%%%%%%%%%%%%%%%%%%%%%%%%%%
\begin{figure}[h]
\centering
\includegraphics[width=0.55\linewidth]{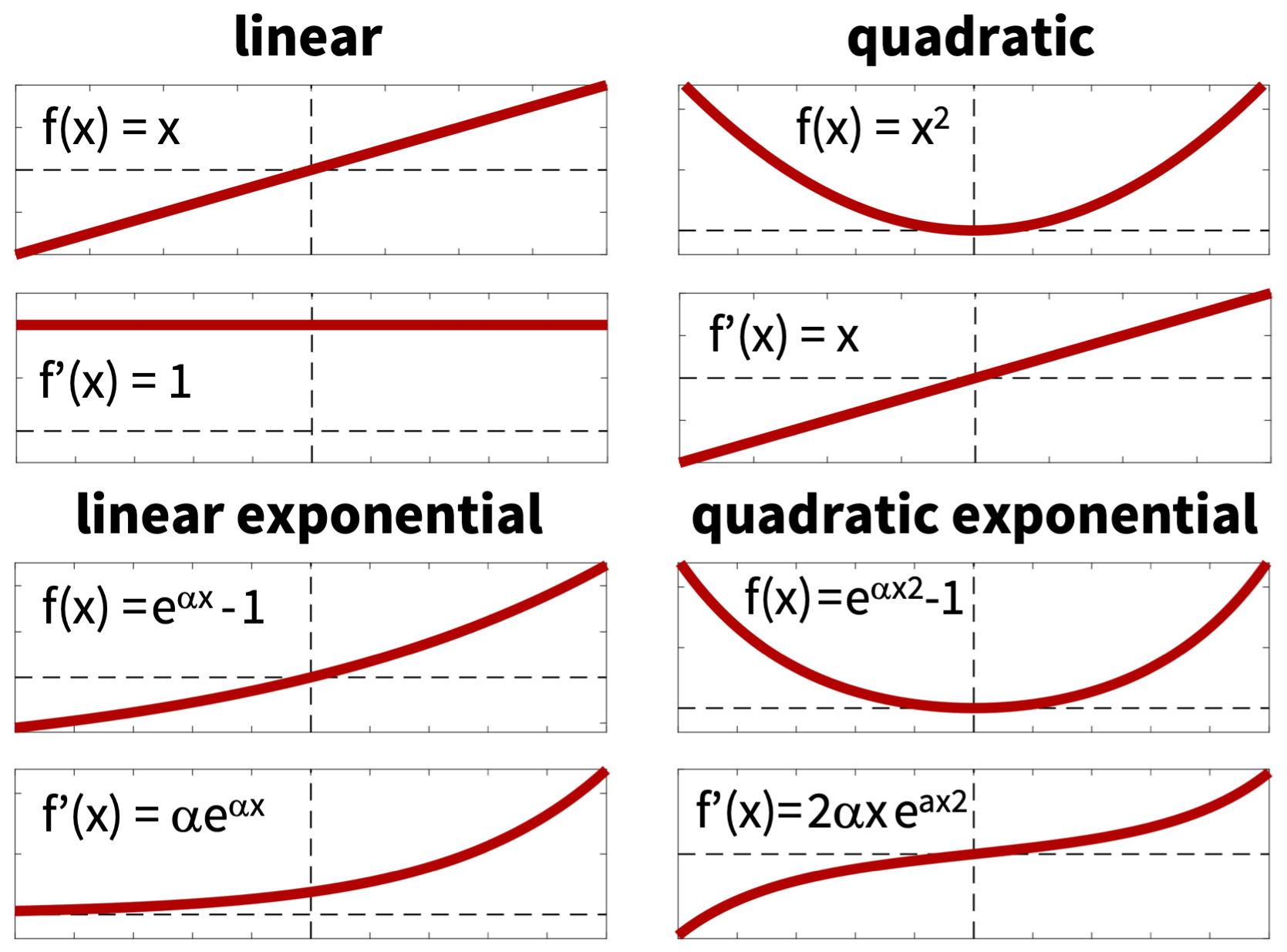}
\caption{{\bf{\sffamily{Activation functions for Constitutive Artificial Neural Networks.}}} 
We use custom-designed activation functions $f(x)$ along with their derivatives $f'(x)$ that include linear and quadratic mappings, either as final activation functions themselves, top rows, or combined with exponential functions, bottom rows, to reverse engineer a free energy function that captures popular functional forms of constitutive terms.}
\label{fig04}
\end{figure}\\[6.pt]
%%%%%%%%%%%%%%%%%%%%%%%%%%%%%%%%%%%%%%%%%%%%%%%%%%%%%%%%%%%%%%%%%%%%%%%%
Figure \ref{fig04} illustrates the four activation functions $f(x)$ along with their derivatives $f'(x)$ that we use throughout the remainder of this work. Notably, in contrast to the activation functions for classical Neural Networks in Figure \ref{fig02}, all four functions are not only
{\it{monotonic}}, $f(x+\varepsilon) \ge f(x)$ for $\varepsilon\ge0$, such that increasing deformations result in increasing stresses, but also 
{\it{continuous}} at the origin, $f(-0)=f(+0)$, 
{\it{continuously differentiable}} and {\it{smooth}} at the origin, $f'(-0)=f'(+0)$,  
zero at the origin, $f(0)=0$, to ensure an energy- and stress-free reference configuration according to equation (\ref{physics_zero}), and
{\it{unbounded}}, $f(-\infty) \rightarrow \infty$ and $f(+\infty) \rightarrow \infty$, to ensure an infinite energy and stress for extreme deformations according to equation~(\ref{physics_infty}).%\\[6.pt]
%%%%%%%%%%%%%%%%%%%%%%%%%%%%%%%%%%%%%%%%%%%%%%%%%%%%%%%%%%%%%%%%%%%%%%%%
\begin{figure}[h]
\centering
\includegraphics[width=0.72\linewidth]{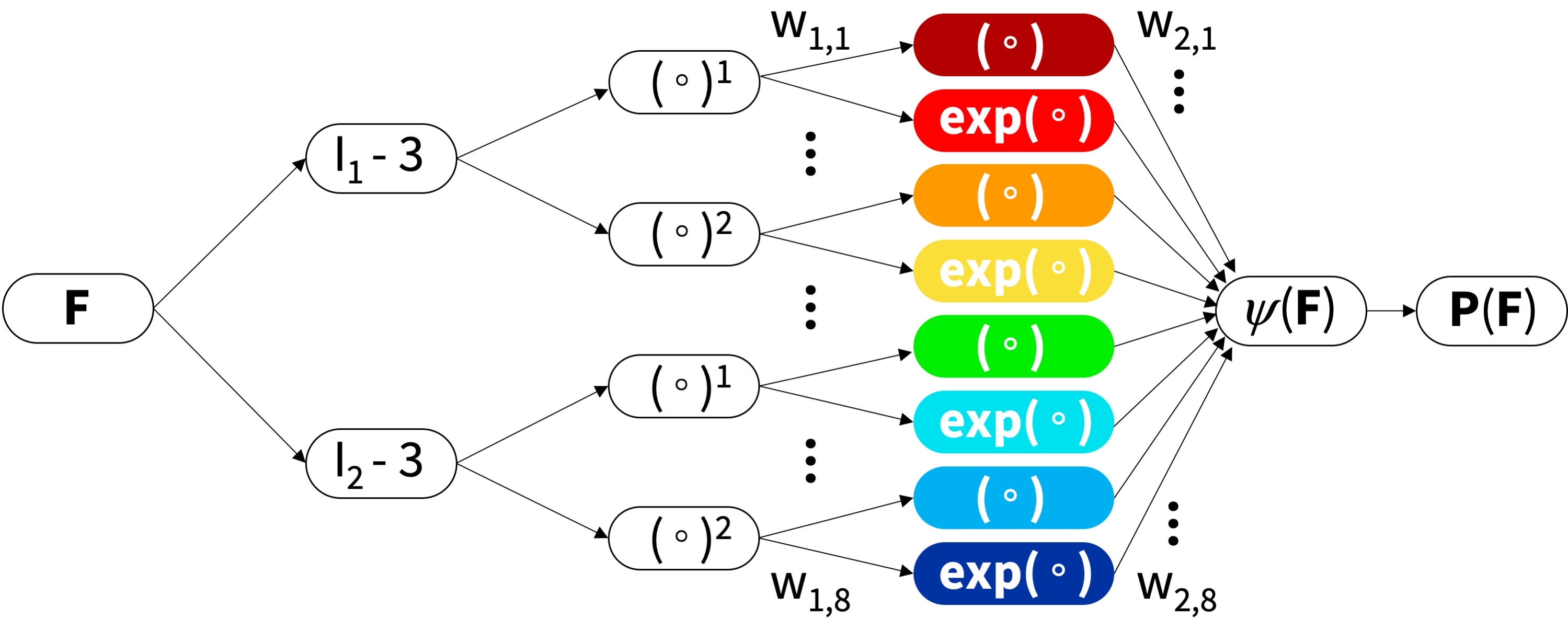}
\caption{{\bf{\sffamily{Constitutive Artificial Neural Network.}}} Example of an isotropic perfectly incompressible Constitutive Artificial Neural Network with with two hidden layers to approximate the single scalar-valued free energy function $\psi(I_1, I_2)$ as a function of the first and second invariants of the deformation gradient $\ten{F}$ using eight terms. The first layer generates powers $(\circ)$ and $(\circ)^2$ of the network input and the second layer applies the identity $(\circ)$ and exponential functions $(\rm{exp}(\alpha(\circ))-1)$ to these powers. The networks is not fully connected by design to a priori satisfy the condition of polyconvexity.}
\label{fig05}
\end{figure}\\[6.pt]
%%%%%%%%%%%%%%%%%%%%%%%%%%%%%%%%%%%%%%%%%%%%%%%%%%%%%%%%%%%%%%%%%%%%%%%%
Figure \ref{fig05} illustrates an example of an isotropic incompressible Constitutive Artificial Neural Network with two hidden layers and four and eight nodes. The first layer generates powers $(\circ)$ and $(\circ)^2$ of the network input and the second layer applies the identity, $(\circ)$, and the exponential function, $(\rm{exp}(\alpha(\circ))-1)$, to these powers. As such, the first and fifths dark red and green inputs to the free energy in Figure \ref{fig05} correspond to the linear activation function in Figure \ref{fig04}, the second and sixths red and light blue inputs correspond to the quadratic activation function, the third and sevenths orange and blue inputs correspond to the linear exponential function, and the fourth and eights yellow and dark blue inputs correspond to the quadratic exponential function. The set of equations for this networks takes the following explicit form, 
%%%%%%%%%%%%%%%%%%%%%%%%%%%%%%%%%%%%%%%%%%%%%%%%%%%%%%%%%%%%%%%%%%%%%%%%
\beq
\begin{array}{l@{\hspace*{0.1cm}}c@{\hspace*{0.1cm}}l@{\hspace*{0.1cm}}
              l@{\hspace*{0.1cm}}c@{\hspace*{0.1cm}}
              l@{\hspace*{0.1cm}}l@{\hspace*{0.1cm}}l@{\hspace*{0.04cm}}
              l@{\hspace*{0.1cm}}c@{\hspace*{0.1cm}}
              l@{\hspace*{0.1cm}}l@{\hspace*{0.1cm}}l@{\hspace*{0.0cm}}l}
    \psi(I_1,I_2)
&=& w_{2,1}w_{1,1} &[\,I_1 - 3\,]
&+& w_{2,2} & [ \, \exp (\,   w_{1,2} & [\, I_1 -3 \,]&)   - 1\,] \\
&+& w_{2,3}w_{1,3} &[\,I_1 - 3\,]^2
&+& w_{2,4} & [ \, \exp (\,   w_{1,4} & [\, I_1 -3 \,]^2&) - 1\,] \\
&+& w_{2,5}w_{1,5} &[\,I_2 - 3\,]
&+& w_{2,6} & [ \, \exp (\,   w_{1,6} & [\, I_2 -3 \,]&)   - 1\,] \\
&+& w_{2,7}w_{1,7} &[\,I_2 - 3\,]^2
&+& w_{2,8} & [ \, \exp (\,   w_{1,8} & [\, I_2 -3 \,]^2&)- 1\,] \,.
\label{CANNenergy}
\end{array}
\eeq
%%%%%%%%%%%%%%%%%%%%%%%%%%%%%%%%%%%%%%%%%%%%%%%%%%%%%%%%%%%%%%%%%%%
For this particular format, one of the first two weights of each row becomes redundant, and we can reduce the set of network parameters to twelve, 
$ \mat{w} = 
[\,(w_{1,1}w_{2,1}), w_{1,2}, w_{2,2}, (w_{1,3}w_{2,3}), w_{1,4}, w_{2,4} 
   (w_{1,5}w_{2,5}), w_{1,6}, w_{2,6}, (w_{1,7}w_{2,7}), w_{1,8}, w_{2,8} \,]$.
Using the second law of thermodynamics, we can derive an explicit expression for the Piola stress from equation (\ref{thermodynamic}), 
$\ten{P} = {\partial \psi}/{\partial \ten{F}}$,
or, more specifically, for the case of perfect incompressibility 
from equation (\ref{incompressibility}),
$\ten{P} = {\partial \psi}/{\partial I_1} \cdot {\partial I_1}/{\partial \ten{F}}
         + {\partial \psi}/{\partial I_2} \cdot {\partial I_2}/{\partial \ten{F}}$,
%          = \partial_{\scas{I_1}} \psi \, \partial_{\tens{F}} I_1
%          + \partial_{\scas{I_2}} \psi \, \partial_{\tens{F}} I_2$,
%%%%%%%%%%%%%%%%%%%%%%%%%%%%%%%%%%%%%%%%%%%%%%%%%%%%%%%%%%%%%%%%%%%
\beq
\begin{array}{l@{\hspace*{0.1cm}}l@{\hspace*{0.1cm}}l@{\hspace*{0.0cm}}
              l@{\hspace*{0.1cm}}l@{\hspace*{0.1cm}}l@{\hspace*{0.0cm}}
              l@{\hspace*{0.0cm}}l@{\hspace*{0.1cm}}l@{\hspace*{0.1cm}}
              l@{\hspace*{0.0cm}}l@{\hspace*{0.1cm}}l@{\hspace*{0.1cm}}
              l@{\hspace*{0.0cm}}l@{\hspace*{0.1cm}}l@{\hspace*{0.1cm}}
              l@{\hspace*{0.1cm}}c@{\hspace*{0.1cm}}
              l@{\hspace*{0.1cm}}l@{\hspace*{0.1cm}}l@{\hspace*{0.0cm}}
              l@{\hspace*{0.1cm}}l}
   \ten{P}
%=  \D{\frac{\partial \psi}{\partial \ten{F}}}
&=&[
  & w_{2,1}   w_{1,1}   &
&+& w_{2,2} & w_{1,2} & \exp (\,   w_{1,2} & [\, I_1 -3 \,]&) \\
%&+& W_{3,1} & W_{3,2} & /[\,1- W_{3,1}&[\, I_1 -3 \,]\,] \\
&+&2\,[\,I_1 - 3\,][& w_{2,3}w_{1,3} &
&+& w_{2,4} & w_{1,4} & \exp (\,   w_{1,4} & [\, I_1 -3 \,]^2&)]
%&+& W_{6,1} & W_{6,2} & /[\,1- W_{6,1}&[\, I_1 -3 \,]^2\,]]]
%&\D{\frac{\partial I_1}{\partial\ten{F}}} \\
%&\D{\partial_{\tens{F}} I_1 }
&\D{\partial I_1}/{\partial \ten{F}}\\
%%%  
&+&[
  & w_{2,5}   w_{1,5}   &
&+& w_{2,6} & w_{1,6} & \exp (\,   w_{1,6} & [\, I_2 -3 \,]&) \\
%&+& W_{9,1} & W_{9,2} & /[\,1- W_{9,1}&[\, I_2 -3 \,]\,] \\
&+&2\,[\,I_2 - 3\,][& w_{2,7}w_{1,7} &
&+& w_{2,8} & w_{1,8}& \exp (\,   w_{1,8} & [\, I_2 -3 \,]^2&)]
%&+& W_{12,1} & W_{12,2}& /[\,1- W_{12,1}&[\, I_2 -3 \,]^2\,]]]
%&\D{\frac{\partial I_2}{\partial\ten{F}}}  \\
%&\D{\partial_{\tens{F}} I_2 }
&\D{\partial I_2}/{\partial \ten{F}}
\label{CANNstress}
\end{array}
\eeq
%%%%%%%%%%%%%%%%%%%%%%%%%%%%%%%%%%%%%%%%%%%%%%%%%%%%%%%%%%%%%%%%%%%
Compared to the stress stretch relation $\ten{P}(\ten{F})$ of classical Neural Networks (\ref{neuralnetwork_tanh}), the stress stretch relation of Constitutive Artificial Neural Networks (\ref{CANNstress}) is fairly simple by design. More importantly, the particular form (\ref{CANNstress}) represents a {\it{generalization}} of many popular constitutive models for incompressible hyperelastic materials. It seems natural to ask whether and how our network parameters $\mat{w}$ relate to common well-known material parameters.   \\[6.pt]
%%%%%%%%%%%%%%%%%%%%%%%%%%%%%%%%%%%%%%%%%%%%%%%%%%%%%%%%%%%%%%%%%%%
%%%%%%%%%%%%%%%%%%%%%%%%%%%%%%%%%%%%%%%%%%%%%%%%%%%%%%%%%%%%%%%%%%%
{\bf{\sffamily{Special types of constitutive equations.}}} 
%\noindent \textcolor{red}{{\bf{Motivation:}} Can we build a constitutive neural network that has the four examples, Neo Hooke, Demiray, Holzapfel, and Gent as special cases?} \\
%%%%%%%%%%%%%%%%%%%%%%%%%%%%%%%%%%%%%%%%%%%%%%%%%%%%%%%%%%%%%%%%%%%
To demonstrate that the family of Constitutive Artificial Neural Networks in Figure \ref{fig03} and the specific example in Figure \ref{fig05} are a {\it{generalization}} of popular constitutive models, we consider several widely used models and systematically compare their material parameters to our network weights $\mat{w}$:\\[6.pt]
%%%
\noindent
The {\it{neo Hooke model}} \cite{treloar48}, the simplest of all models, has a free energy function that is a constant function of only the first invariant, $[\,I_1-3\,]$, scaled by the shear modulus $\mu$,
\beq
  \psi 
= \mbox{$\frac{1}{2}$} \, \mu \, [\,I_1-3\,]
  \qquad \mbox{where} \qquad
  \mu 
= 2\, w_{1,1} w_{2,1} 
  \;\, \mbox{in eq.} (\ref{CANNenergy})\,.
\label{neohooke}
\eeq
The {\it{Blatz Ko model}} \cite{blatz62}, has a free energy function that depends only the second and third invariants, $[\,I_2-3\,]$ and $[\,I_3-1\,]$, scaled by the shear modulus $\mu$,
$ \psi 
= \mbox{$\frac{1}{2}$} \, \mu \, [\,I_2/I_3 + 2\, \sqrt{I_3} - 5\,]$. 
For perfectly incompressible materials, $I_3 = 1$, it simplifies to the following form, 
\beq
  \psi 
= \mbox{$\frac{1}{2}$} \, \mu \, [\,I_2 - 3\,]
  \qquad \mbox{where} \qquad
  \mu 
= 2\, w_{1,5} w_{2,5} 
  \;\, \mbox{in eq.} (\ref{CANNenergy})\,.
\label{blatzko}  
\eeq
The {\it{Mooney Rivlin model}} \cite{mooney40,rivlin48} is a combination of both and accounts for the first and second invariants, $[\,I_1-3\,]$ and $[\,I_2-3\,]$, scaled by the moduli $\mu_1$ and $\mu_2$ that sum up to the overall shear modulus, $\mu = \mu_1 + \mu_2$,
% special case of Ogden model with $\alpha=\pm 2$, $\beta = \pm C_{1,2}$
\beq
  \psi 
= \mbox{$\frac{1}{2}$} \, \mu_1 \, [\,I_1-3\,]
+ \mbox{$\frac{1}{2}$} \, \mu_2 \, [\,I_2-3\,]
 \qquad \mbox{where} \qquad
 \mu_1 = 2\, w_{1,1} w_{2,1} 
 \;\,\mbox{and} \;\,
 \mu_2 = 2\, w_{1,5} w_{2,5} 
 \;\, \mbox{in eq.} (\ref{CANNenergy}) \,.
\label{mooney} 
\eeq 
The {\it{Yeoh model}} \cite{yeoh93} considers linear, quadratic, and cubic terms of only the first invariant, $[I_1-3]$, as
\beq
  \psi 
= a_1 \, [\,{I}_1-3\,]
+ a_2 \, [\,{I}_1-3\,]^2
+ a_3 \, [\,{I}_1-3\,]^3
  \; \mbox{where} \;
  a_1 = 2\, w_{1,1} w_{2,1}
  \, \mbox{and} \,
  a_2 = 2\, w_{1,3} w_{2,3}
  \, \mbox{and} \,
  a_3 = 0
  \, \mbox{in eq.} (\ref{CANNenergy_generic}) \,.
\label{yeoh}  
\eeq  
The {\it{Demiray model}} \cite{demiray72} or {\it{Delfino model}} \cite{delfino97} uses
linear exponentials of the first invariant, $[I_1-3]$, in terms of two parameters $a$ and $b$,
%when expanded into a power series of $[\,I_1-3\,]$, the first two non-zero terms are Neo Hooke, $f=\frac{1}{2}\,\mu\,[\,I_1-3\,]$
\beq
  \psi 
= \frac{1}{2} \, \frac{a}{b} \, 
  [\,\exp (\, b\, [\,I_1-3\,] \,) -1\,]
  \qquad \mbox{where} \qquad
  a = 2\, w_{1,2} w_{2,2} 
  \;\, \mbox{and} \;\,
  b = w_{1,2} 
  \;\, \mbox{in eq.} (\ref{CANNenergy}) \,.
\label{demiray}
\eeq  
The {\it{Treloar model}} \cite{treloar48} and {\it{Mooney Rivlin model}} \cite{mooney40,rivlin48} for nearly incompressible materials both consider a quadratic term of the third invariant, $[\,J-1\,]$, scaled by the bulk modulus $\kappa$, to additionally account for the bulk behavior,
\beq
  \bar{\psi}
= \mbox{$\frac{1}{2}$} \, \kappa \, [\,J-3\,]^2
  \qquad \mbox{where} \qquad
  \kappa = 2\, w_{1,13} w_{2,13} 
   \;\, \mbox{in eq.} (\ref{CANNenergy_generic}) \,.
\eeq
The {\it{Holzapfel model}} \cite{holzapfel00} uses quadratic exponentials of the fourth invariant, 
$[\,I_4-1\,]$, in terms of two parameters  $a$ and $b$ to additionally account for a transversely isotropic behavior,
\beq
  \bar{\psi}
= \frac{1}{2} \, \frac{a}{b} \, 
  [\,\exp (\, b\, [\,I_4-1\,]^2 \,) -1\,]
  \qquad \mbox{where} \qquad
  a = 2\, w_{1,22} w_{2,22} 
  \;\, \mbox{and} \;\,
  b = w_{1,22} 
  \;\, \mbox{in eq.} (\ref{CANNenergy_generic}) \,.
\eeq 
These simple examples demonstrate that we can recover popular constitutive functions for which the network weights gain a well-defined physical meaning.\\[6.pt]
%%%%%%%%%%%%%%%%%%%%%%%%%%%%%%%%%%%%%%%%%%%%%%%%%%%%%%%%%%%%%%%%%%%
\noindent
{\bf{\sffamily{Loss function.}}} 
%%%%%%%%%%%%%%%%%%%%%%%%%%%%%%%%%%%%%%%%%%%%%%%%%%%%%%%%%%%%%%%%%%%
The objective of a Constitutive Artificial Neural Network is to learn the network parameters 
$\vec{\theta}=\{ \mat{w}_k \}\,$, the network weights, 
by minimizing a loss function $L$ that penalizes the error between model and data. 
Similar to classical Neural Networks, we characterize this error as the mean squared error, the $L_2$-norm of the difference between model 
$\ten{P}(\ten{F}_i)$ and data $\hat{\ten{P}}_i$, 
divided by the number of training points $n_{\rm{trn}}$,
\beq
  L (\vec{\theta} ; \ten{F})
= \frac{1}{n_{\rm{trn}}} \sum_{i=1}^{n_{\rm{trn}}}
|| \, \ten{P}(\ten{F}_i) - \hat{\ten{P}}_i \, ||^2 
\rightarrow \mbox{min}\,.
\label{loss_CANN}
\eeq
While this is not the focus of the present work, in the spirit of Physics Informed Neural Networks, we could add additional thermodynamic constraints to the loss function \cite{karniadakis21,linka22}. For the perfectly incompressible hyperelastic materials we consider here, the thermodynamics are already well represented and hardwired into the network through input, output, architecture and activation functions, and we do not need to consider this extra step. 
We train the network by minimizing the loss function (\ref{loss_CANN}) and learn the network parameters $\vec{\theta}= \{ \mat{w} \}$ using the ADAM optimizer, a robust adaptive algorithm for gradient-based first-order optimization, and constrain the network weights to always remain non-negative, $\mat{w} \ge \mat{0}$.
While we could equally well solve the optimization problem (\ref{loss_CANN}) using a different optimization solver, we capitalize on the power and robustness of optimizers developed for machine learning and opt for the widely used ADAM optimizer, rather than implementing this minimization ourselves.\\[6.pt]
With only small amounts of training data, Constitutive Artificial Neural Networks can both interpolate and extrapolate well and make informed predictions within the range of validity of the underlying thermodynamic assumptions. 
Since they limit the number of degrees of freedom, they are less likely to overfit, especially if the available data are sparse. 
By design, Constitutive Artificial Neural Networks are compliant with the thermodynamic restrictions of Section \ref{sec_const}. 
Most importantly, for practical applications, they do not operate as a black box; rather they are a {\it{generalization}} of existing constitutive models and their parameters have a clear {\it{physical interpretation}}. 
%%%%%%%%%%%%%%%%%%%%%%%%%%%%%%%%%%%%%%%%%%%%%%%%%%%%%%%%%%%%%%%%%%%
\section{Special homogeneous deformation modes}\label{sec_homdef}
\begin{figure}[h]
\centering
\includegraphics[width=0.72\linewidth]{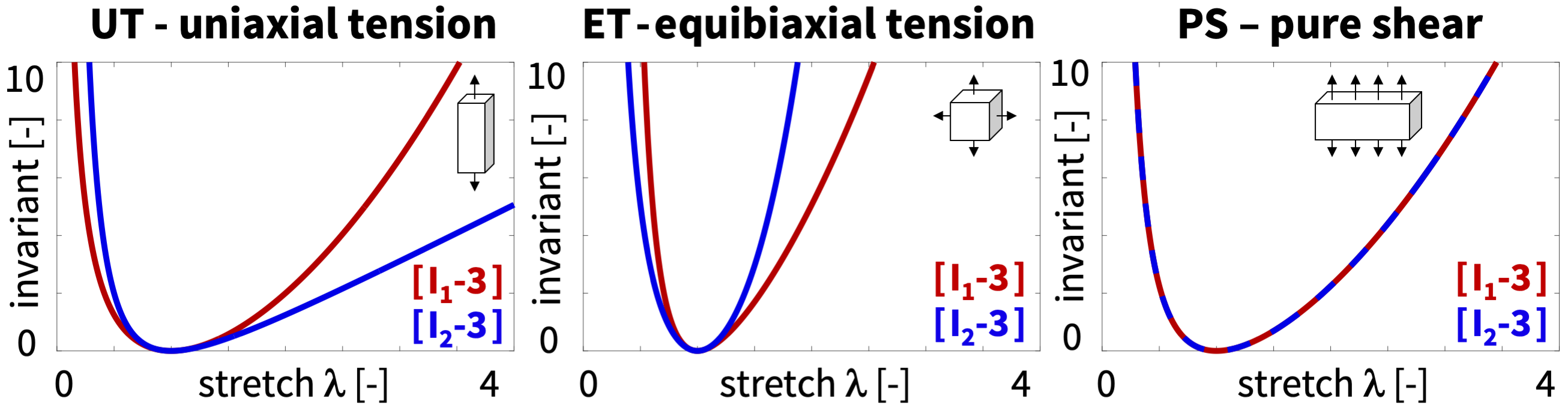}
\caption{{\bf{\sffamily{Special homogeneous deformation modes.}}} 
Invariant-stretch relations for the special modes of perfectly incompressible uniaxial tension with 
$  \ten{F} 
= {\rm{diag}} \, \{ \; \lambda, \lambda^{-1/2}, \lambda^{-1/2}\}$, 
equibiaxial tension with 
$  \ten{F} 
= {\rm{diag}} \, \{ \; \lambda, \lambda, \lambda^{-2}\}$, and 
pure shear with 
$  \ten{F} 
= {\rm{diag}} \, \{ \; \lambda, 1, \lambda^{-1}\}$.}
\label{fig06}
\end{figure}

%%%%%%%%%%%%%%%%%%%%%%%%%%%%%%%%%%%%%%%%%%%%%%%%%%%%%%%%%%%%%%%%%%%
\noindent
To demonstrate the features of our thermodynamically consistent Constitutive Artificial Neural Networks, we consider an {\it{isotropic}}, {\it{perfectly incompressible}} material for which the principal stretches $\lambda_i$ and nominal stresses $P_i$ are related via
\beq
  P_i 
= \frac{\partial \psi}{\partial \lambda_i} 
- \frac{1}{\lambda_i} \, p
\qquad \forall \qquad i = 1,2,3,
\label{P_lambda}
\eeq
where $p$ denotes the hydrostatic pressure. 
Using the chain rule, we can reparameterize equation (\ref{P_lambda}) in terms of the invariants $I_1$ and $I_2$, recalling the incompressibility constraint $I_3=1$, such that
\beq
  P_i 
= \frac{\partial \psi}{\partial I_1} \,
  \frac{\partial I_1} {\partial \lambda_i} 
+ \frac{\partial \psi}{\partial I_2} \,
  \frac{\partial I_2} {\partial \lambda_i} 
- \frac{1}{\lambda_i} \, p
\qquad \forall \qquad i = 1,2,3.
\label{stress}
\eeq
In the following, we summarize the deformation gradients $\ten{F}$, the invariants $I_1$ and $I_2$, their derivatives $\partial I_1/\partial \lambda$ and $\partial I_2/\partial \lambda$, and the resulting nominal stress $\ten{P}$ for the special homogeneous deformation modes of incompressible {\it{uniaxial tension}}, {\it{equibiaxial tension}}, and {\it{pure shear}} \cite{ogden72}. Figure \ref{fig06} summarizes the stretch-invariant relationship for all three cases. 
%%%%%%%%%%%%%%%%%%%%%%%%%%%%%%%%%%%%%%%%%%%%%%%%%%%%%%%%%%%%%%%%%%%%%%%%
\\[6.pt]
%%%%%%%%%%%%%%%%%%%%%%%%%%%%%%%%%%%%%%%%%%%%%%%%%%%%%%%%%%%%%%%%%%%%%%%%
%%%%%%%%%%%%%%%%%%%%%%%%%%%%%%%%%%%%%%%%%%%%%%%%%%%%%%%%%%%%%%%%%%%
\noindent
{\bf{\sffamily{Uniaxial tension.}}} 
%%%%%%%%%%%%%%%%%%%%%%%%%%%%%%%%%%%%%%%%%%%%%%%%%%%%%%%%%%%%%%%%%%%
For the special case of incompressible uniaxial tension, we stretch the specimen in one direction, $\lambda_1 = \lambda$.
From isotropy and incompressibility,
$I_3 = \lambda_1^2 + \lambda_2^2 + \lambda_3^2 = 1$,
we conclude that the stretches orthogonal to this direction are
the same and equal to the square root of the stretch,
$\lambda_2=\lambda_3=\lambda^{-1/2}$.
The deformation gradient $\ten{F}$ and Piola stress $\ten{P}$ for incompressible uniaxial tension follow as
\beq
  \ten{F} 
= {\rm{diag}} \, \{ \; \lambda, \lambda^{-1/2}, \lambda^{-1/2}\}
  \quad \mbox{and} \quad
  \ten{P} 
= {\rm{diag}} \, \{ \; P_{1}, 0, 0\, \} \,.
\eeq
We can use the explicit expressions of the
first and second invariants and their derivatives,
\beq
  I_1 
= \lambda^2 + \frac{2}{\lambda}
  \quad \mbox{and} \quad
  I_2 
= 2\lambda + \frac{1}{\lambda^2}
  \quad \mbox{with} \quad
  \frac{\partial I_1}{\partial  \lambda} 
= 2 \, \left[\lambda - \frac{1}{\lambda^2} \right]
  \quad \mbox{and} \quad
  \frac{\partial I_2}{\partial  \lambda} 
= 2 \, \left[1 - \frac{1}{\lambda^3}\right] \,,
\eeq
to determine the pressure $p$ from the zero stress condition in the transverse directions, $P_2 = 0$ and $P_3 = 0$, using equation (\ref{stress}),
\beq
  p 
= \frac{2}{\lambda} \, 
  \frac{\partial \psi}{\partial I_1}
+ 2 
  \left[
  \lambda+\frac{1}{\lambda^2}
  \right] \,
  \frac{\partial \psi}{\partial I_2} \,,
\eeq
and obtain an explicit analytical expression for the nominal stress $P_1$ in terms of the stretch $\lambda$ from equation (\ref{stress}),
\beq
  P_1 
= 2 \, \left[ 
  \frac{\partial \psi}{\partial I_1}
+ \frac{1}{\lambda}
  \frac{\partial \psi}{\partial I_2}
  \right]
  \left[
  \lambda - \frac{1}{\lambda^2}
  \right]\,.
\label{stressUT}  
\eeq
%%%%%%%%%%%%%%%%%%%%%%%%%%%%%%%%%%%%%%%%%%%%%%%%%%%%%%%%%%%%%%%%%%%
{\bf{\sffamily{Equibiaxial tension.}}} 
%%%%%%%%%%%%%%%%%%%%%%%%%%%%%%%%%%%%%%%%%%%%%%%%%%%%%%%%%%%%%%%%%%%
For the special case of incompressible equibiaxial tension, we stretch the specimen equally in two directions, 
$\lambda_1 = \lambda_2 =\lambda$.
From the incompressibility condition,
$I_3 = \lambda_1^2 + \lambda_2^2 + \lambda_3^2 = 1$,
we conclude that the stretch in the third direction is
$\lambda_3=\lambda^{-2}$.
The deformation gradient $\ten{F}$ and Piola stress $\ten{P}$ for incompressible equibiaxial tension follow as
\beq
  \ten{F} 
= {\rm{diag}} \, \{ \; \lambda, \lambda, \lambda^{-2}\}
  \quad \mbox{and} \quad
  \ten{P} 
= {\rm{diag}} \, \{ \; P_{1}, P_{2}, 0\, \} \,.
\eeq
Using the explicit expressions of the
first and second invariants and their derivatives,
\beq
  I_1 
= 2\lambda^2 + \frac{1}{\lambda^4} 
  \quad \mbox{and} \quad
  I_2 
= \lambda^4 + \frac{2}{\lambda^2} 
  \quad \mbox{with} \quad
  \frac{\partial I_1}{\partial  \lambda} 
= 2 \, \left[\lambda - \frac{1}{\lambda^5} \right]
  \quad \mbox{and} \quad
  \frac{\partial I_2}{\partial  \lambda} 
= 2 \, \left[\lambda^3 - \frac{1}{\lambda^3}\right] \,,
\eeq
we determine the pressure $p$ from the zero stress condition in the third direction, $P_3 = 0$, using equation (\ref{stress}),
\beq
  p 
= \frac{2}{\lambda^4} \, 
  \frac{\partial \psi}{\partial I_1}
+ \frac{4}{\lambda^2} \,
  \frac{\partial \psi}{\partial I_2}
\eeq
and obtain an explicit analytical expression for the nominal stresses $P_1$ and $P_2$ in terms of the stretch $\lambda$ from equation~(\ref{stress}),
\beq
  P_1 
= P_2  
= 2 \, \left[ 
  \frac{\partial \psi}{\partial I_1}
+ \lambda^2
  \frac{\partial \psi}{\partial I_2}
  \right]
  \left[
  \lambda - \frac{1}{\lambda^5}
  \right]\,.
\label{stressET}  
\eeq
%%%%%%%%%%%%%%%%%%%%%%%%%%%%%%%%%%%%%%%%%%%%%%%%%%%%%%%%%%%%%%%%%%%
{\bf{\sffamily{Pure shear.}}} 
%%%%%%%%%%%%%%%%%%%%%%%%%%%%%%%%%%%%%%%%%%%%%%%%%%%%%%%%%%%%%%%%%%%
For the special case of incompressible pure shear, we stretch a long rectangular specimen along its short axis, $\lambda_1 = \lambda$, and assume that it remains undeformed along its long axis, $\lambda_2 = 1$.
From the incompressibility condition,
$I_3 = \lambda_1^2 + \lambda_2^2 + \lambda_3^2 = 1$,
we conclude that the stretch in the third direction is
$\lambda_3=\lambda^{-1}$.
The deformation gradient $\ten{F}$ and Piola stress $\ten{P}$ for incompressible pure shear are
\beq
  \ten{F} 
= {\rm{diag}} \, \{ \; \lambda, 1, \lambda^{-1}\}
  \quad \mbox{and} \quad
  \ten{P} 
= {\rm{diag}} \, \{ \; P_{1}, P_{2}, 0\, \} \,.
\eeq
Using the explicit expressions of the
first and second invariants and their derivatives,
\beq
  I_1 
= I_2  
= \lambda^2 + 1 + \frac{1}{\lambda^2} 
  \quad \mbox{with} \quad
  \frac{\partial I_1}{\partial  \lambda} 
= \frac{\partial I_2}{\partial  \lambda} 
= 2 \, \left[\lambda - \frac{1}{\lambda^3} \right]
\eeq
we determine the pressure $p$ from the zero stress condition in the third direction, $P_3 = 0$, using equation (\ref{stress}),
\beq
  p 
= \frac{2}{\lambda^2} \, 
  \frac{\partial \psi}{\partial I_1}
+ 2\,
  \left[ 1 + \frac{1}{\lambda^2} \right] \,
  \frac{\partial \psi}{\partial I_2} \,,
\eeq
and obtain explicit analytical expressions for the nominal stresses $P_1$ and $P_2$ in terms of the stretch $\lambda$ from equation~(\ref{stress}),
\beq
  P_1   
= 2 \, \left[ 
  \frac{\partial \psi}{\partial I_1}
+ \frac{\partial \psi}{\partial I_2}
  \right]
  \left[
  \lambda - \frac{1}{\lambda^3}
  \right]
  \quad \mbox{and} \quad
  P_2   
= 2 \, \left[ 
  \frac{\partial \psi}{\partial I_1}
+ \lambda^2
  \frac{\partial \psi}{\partial I_2}
  \right]
  \left[
  1 - \frac{1}{\lambda^2}
  \right] \,.
\label{stressPS}  
\eeq
%%%%%%%%%%%%%%%%%%%%%%%%%%%%%%%%%%%%%%%%%%%%%%%%%%%%%%%%%%%%%%%%%%%%%%%%

\begin{figure}[h]
\centering
\includegraphics[width=0.72\linewidth]{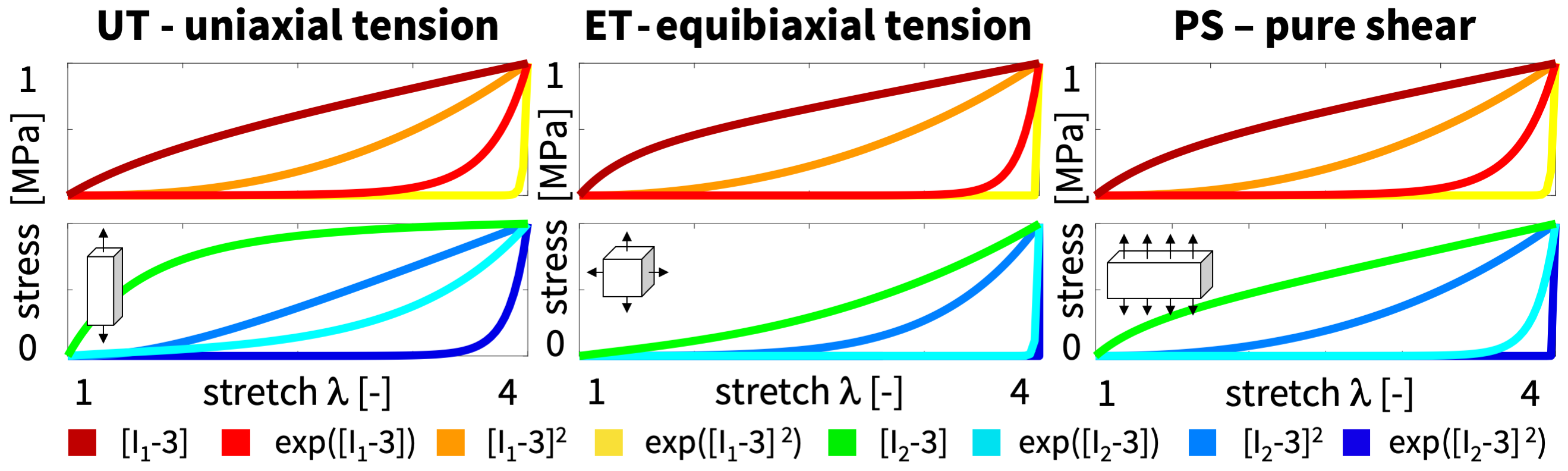}
\caption{{\bf{\sffamily{Special homogeneous deformation modes.}}} 
Stress-stretch relations for the example of the free energy function $\psi$ in equation (\ref{CANNstress}) for the special modes of perfectly incompressible uniaxial tension, 
equibiaxial tension, and pure shear.
The eight curves highlight the linear, quadratic, linear exponential, and quadratic exponential contributions of the first and second invariants $I_1$ and $I_2$ to the final 
final stress function $P_1(\lambda)$ in equations (\ref{stressUT}), (\ref{stressET}), and (\ref{stressPS}). The color-code agrees with the nodes of the Constitutive Artificial Neural Network in Figure \ref{fig05}.}
\label{fig14}
\end{figure}

\noindent
Figure \ref{fig14} illustrates the stress-stretch relations for the example of the free energy function $\psi(\lambda)$ in equation (\ref{CANNenergy})
for the special homogeneous deformation modes of perfectly incompressible uniaxial tension, equibiaxial tension, and pure shear.
The eight curves highlight the linear, quadratic, linear exponential, and quadratic exponential contributions of the first invariant $I_1$, top row, and second invariant $I_2$, bottom row, to the final stress function $P_1(\lambda)$ in equations
(\ref{stressUT}), (\ref{stressET}), and (\ref{stressPS}). For comparison, all curves are scaled to unity. Their color code corresponds to the eight nodes of the Constitutive Artificial Neural Network in Figure \ref{fig05}. 
%%%%%%%%%%%%%%%%%%%%%%%%%%%%%%%%%%%%%%%%%%%%%%%%%%%%%%%%%%%%%%%%%%%%%%%%
 %\\
%%%%%%%%%%%%%%%%%%%%%%%%%%%%%%%%%%%%%%%%%%%%%%%%%%%%%%%%%%%%%%%%%%%%%%%%
The stress contributions of the first invariant take a comparable shape for all three deformation modes: The linear term, $[\,I_1-3\,]$, is concave for all three modes, whereas the other three terms are convex. The terms of the second invariant behave similarly under uniaxial tension and pure shear: The linear term, $[\,I_2-3\,]$, is concave and the other three terms are convex. For equibiaxial tension, however, all four terms, including the $[\,I_2-3\,]$ term, are convex. Notably, both quadratic exponential terms increase rapidly for all six cases. In the following section, when we train our Constitutive Artificial Neural Network with real data, we will explore how linear combinations of these eight terms, scaled by the learnt twelve network weights $\mat{w}$, make up the free energy function $\psi(\lambda)$, and with it the stress $P(\lambda)$ that best approximates the data $\hat{P}$.
%%%%%%%%%%%%%%%%%%%%%%%%%%%%%%%%%%%%%%%%%%%%%%%%%%%%%%%%%%%%%%%%%%%%%%%%
\section{Results}\label{sec_results}
\begin{figure}[h]
\centering
\includegraphics[width=0.72\linewidth]{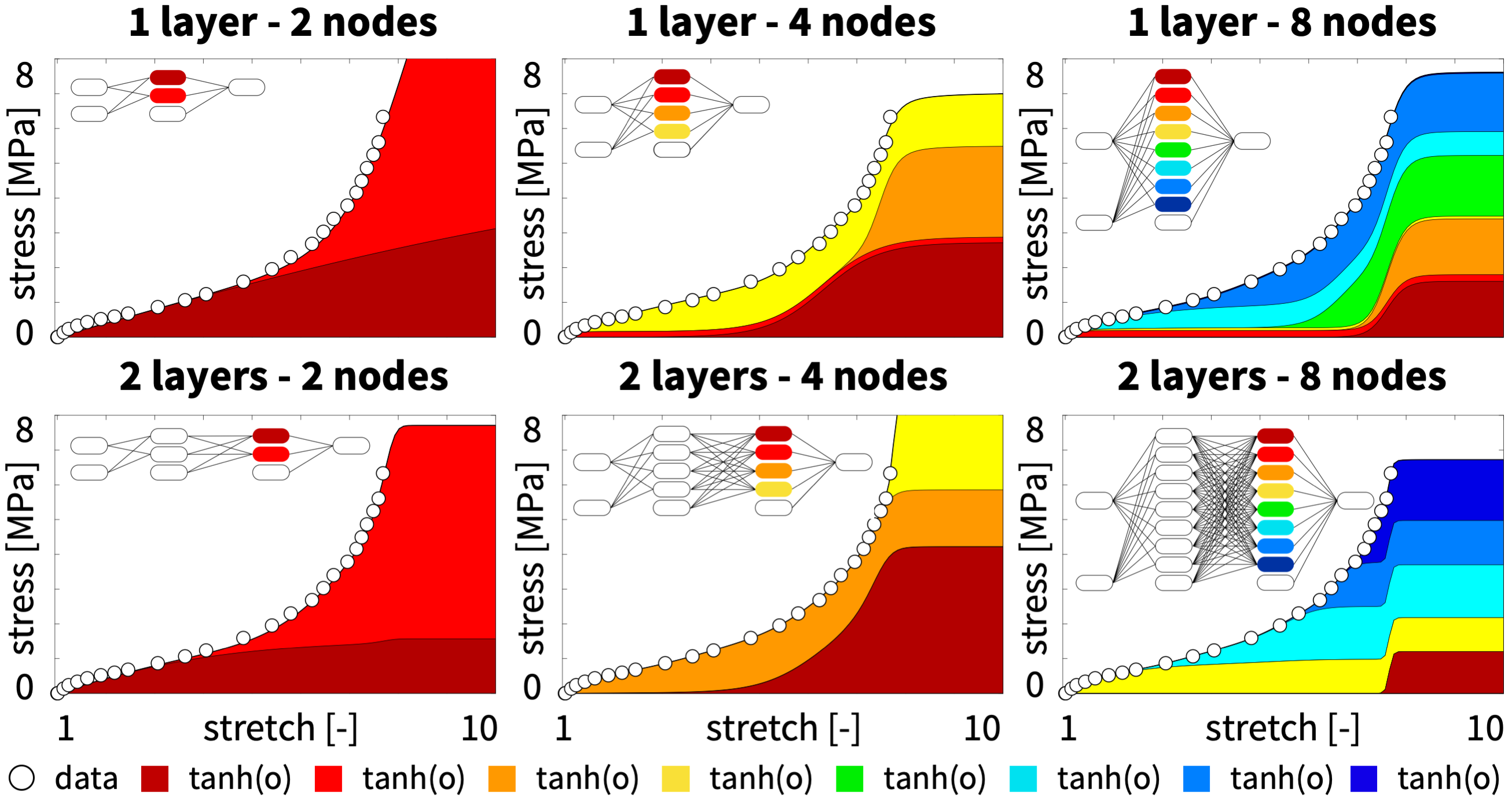}
\caption{{\bf{\sffamily{Classical Neural Networks. Effect of network depth and breadth.}}} Piola stress $P(\lambda)$ as a function of the stretch $\lambda$ for six fully connected feed forward Neural Networks with one and two layers and two, four, and eight nodes using the hyperbolic tangent activation function from Figure \ref{fig02}.
Dots illustrate the uniaxial tension data $\hat{P}$ for rubber at 20$^{\circ}$ \cite{treloar44} from Tables \ref{tab01} and \ref{tab02}; color-coded areas highlight the contributions of the color-coded nodes to the final stress function $P(\lambda)$.}
\label{fig07}
\end{figure}

\begin{figure}[h]
\centering
\includegraphics[width=0.72\linewidth]{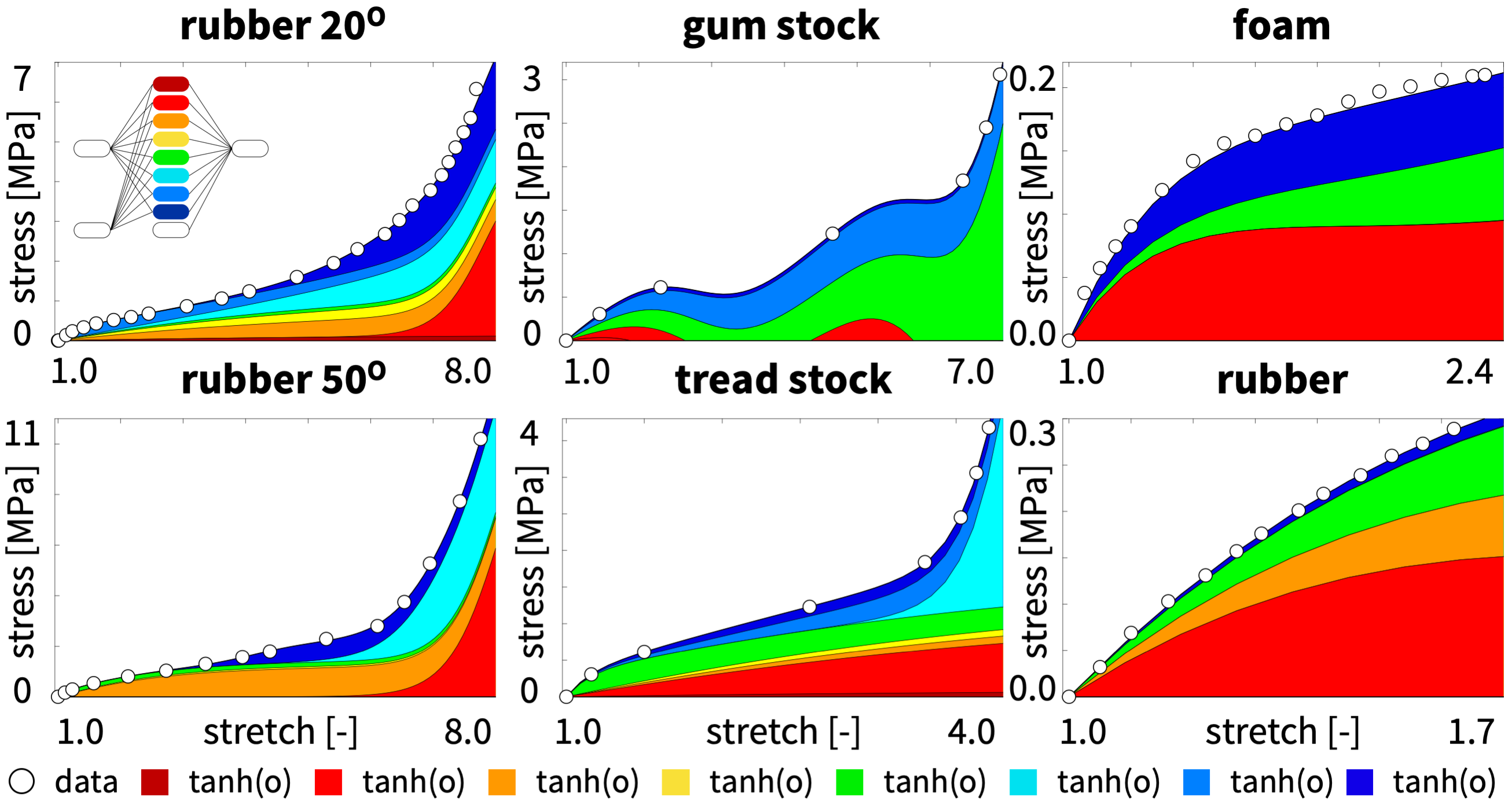}
\caption{{\bf{\sffamily{Classical Neural Network. Uniaxial tension.}}} 
Piola stress $P(\lambda)$ as a function of the stretch $\lambda$ for a fully connected feed 
forward Neural Network with one layer, eight nodes, 16 weights, and nine biases, using the hyperbolic tangent activation function from Figure \ref{fig02}.
Dots illustrate the uniaxial tension data $\hat{P}$ for 
rubber at 20$^{\circ}$ and 50$^{\circ}$ \cite{treloar44}, 
gum stock and tread stock \cite{mooney40}, and
polymeric foam and rubber \cite{blatz62} from Table \ref{tab01}; color-coded areas highlight the contributions of the color-coded nodes to the final stress function $P(\lambda)$.}
\label{fig08}
\end{figure}

\begin{figure}[h]
\centering
\includegraphics[width=0.72\linewidth]{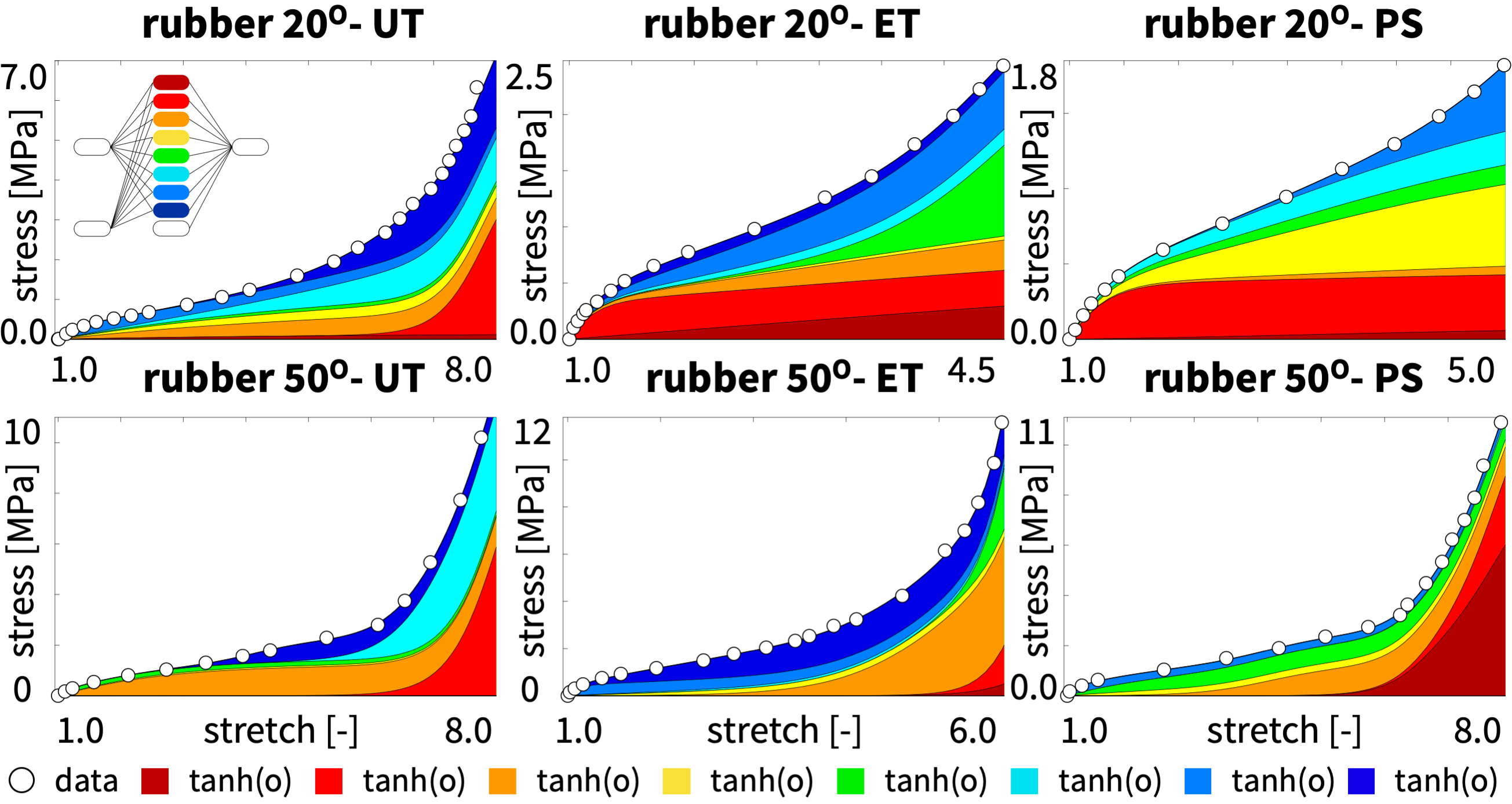}
\caption{{\bf{\sffamily{Classical Neural Network. Uniaxial tension, equibiaxial tension, and pure shear.}}} 
Piola stress $P(\lambda)$ as a function of the stretch $\lambda$ for a fully connected feed forward Neural Network with one layer, eight nodes, 16 weights, and nine biases, using the hyperbolic tangent activation function from Figure \ref{fig02}.
Dots illustrate the uniaxial tension, equibiaxial tension, and pure shear data $\hat{P}$ for rubber at 20$^{\circ}$ and 50$^{\circ}$ \cite{treloar44} from Table \ref{tab02}; color-coded areas highlight the contributions of the color-coded nodes to the final stress function $P(\lambda)$ for individual single-mode training..}
\label{fig09}
\end{figure}

\begin{figure}[t]
\centering
\includegraphics[width=0.72\linewidth]{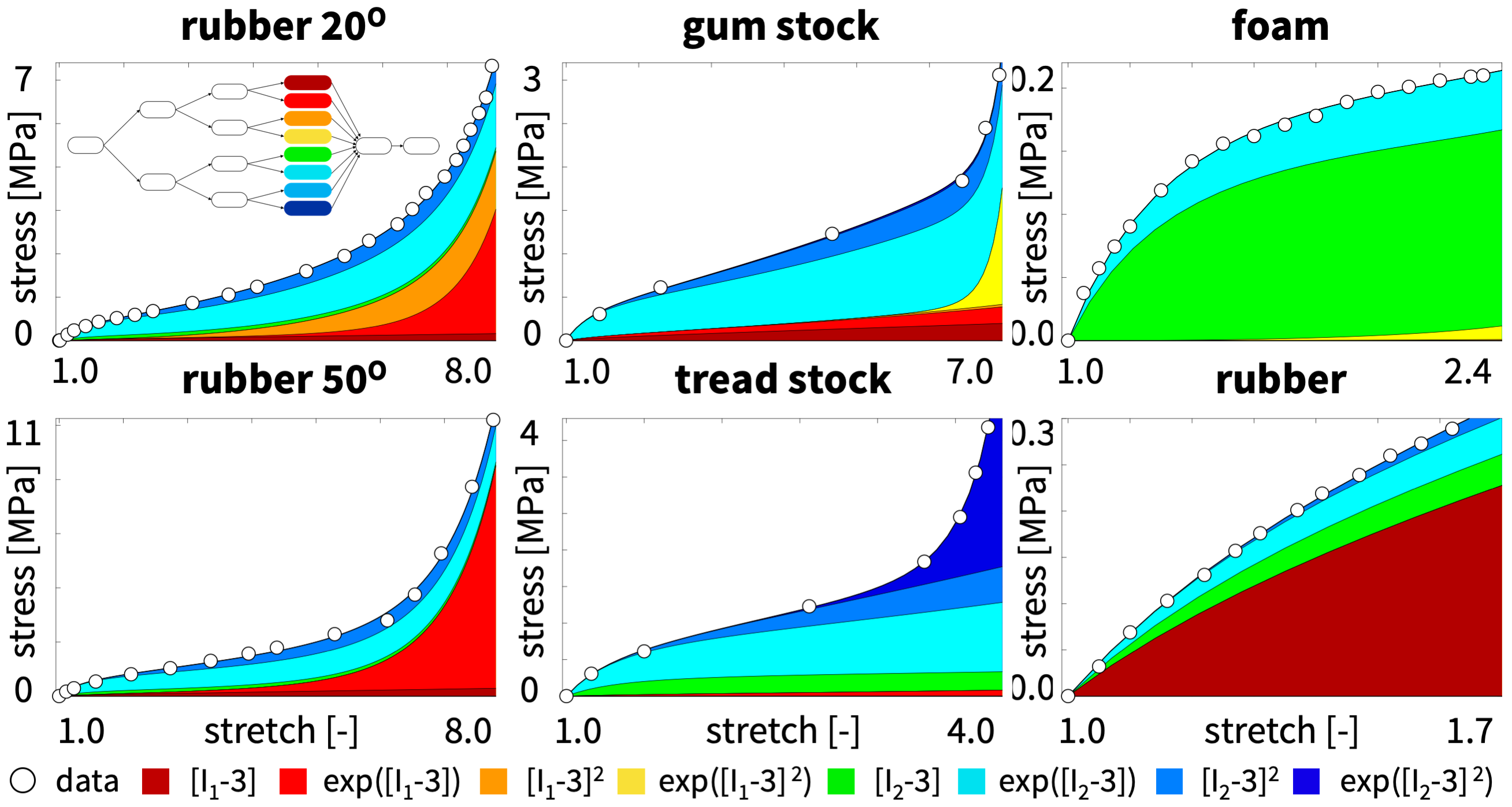}
\caption{{\bf{\sffamily{Constitutive Artificial Neural Network. Uniaxial tension.}}} 
Piola stress $P(\lambda)$ as a function of the stretch $\lambda$ for the 
feed forward Constitutive Artificial Neural Network from Figure \ref{fig05}
with two layers, eight nodes, and twelve weights using the custom-designed activation functions from Figure \ref{fig04}.
Dots illustrate the uniaxial tension data $\hat{P}$ for 
rubber at 20$^{\circ}$ and 50$^{\circ}$ \cite{treloar44}, 
gum stock and tread stock \cite{mooney40}, and
polymeric foam and rubber \cite{blatz62} from Table \ref{tab01}; color-coded areas highlight the contributions of the color-coded nodes to the final stress function $P(\lambda)$.}
\label{fig10}
\end{figure}

\begin{figure}[h]
\centering
\includegraphics[width=0.72\linewidth]{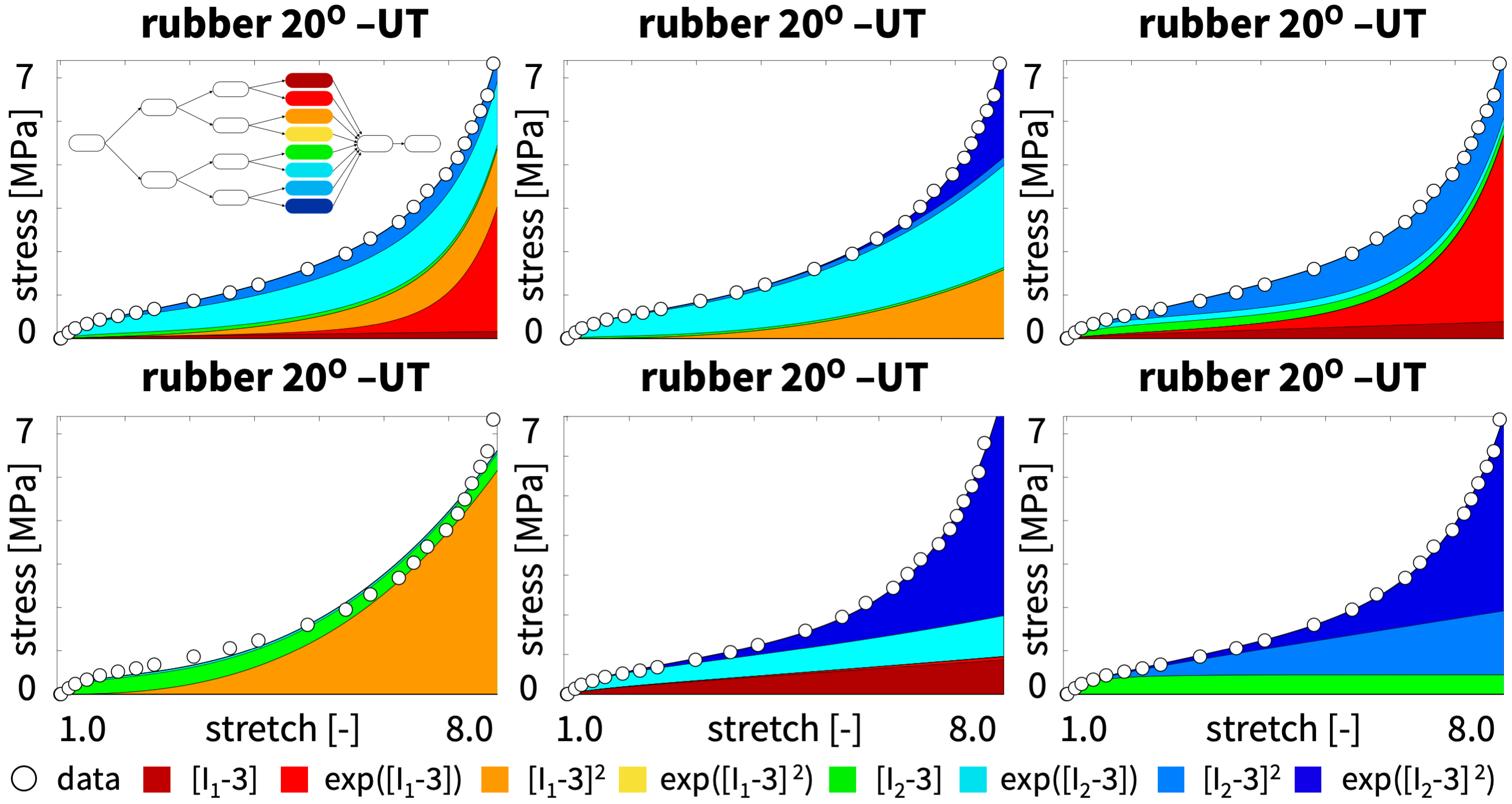}
\caption{{\bf{\sffamily{Constitutive Artificial Neural Network. Effect of initial conditions and non-uniqueness.}}} 
Six Piola stresses $P(\lambda)$ as functions of the stretch $\lambda$ for the 
feed forward Constitutive Artificial Neural Network from Figure \ref{fig05}
with two layers, eight nodes, and 12 weights, initialized with six different sets of initial conditions.
Dots illustrate the uniaxial tension data $\hat{P}$ for 
rubber at 20$^{\circ}$ \cite{treloar44} from Table \ref{tab01}; color-coded areas highlight the contributions of the color-coded nodes to the stress functions $P(\lambda)$ for six different sets of initial conditions.}
\label{fig11}
\end{figure}

\begin{figure}[h]
\centering
\includegraphics[width=0.72\linewidth]{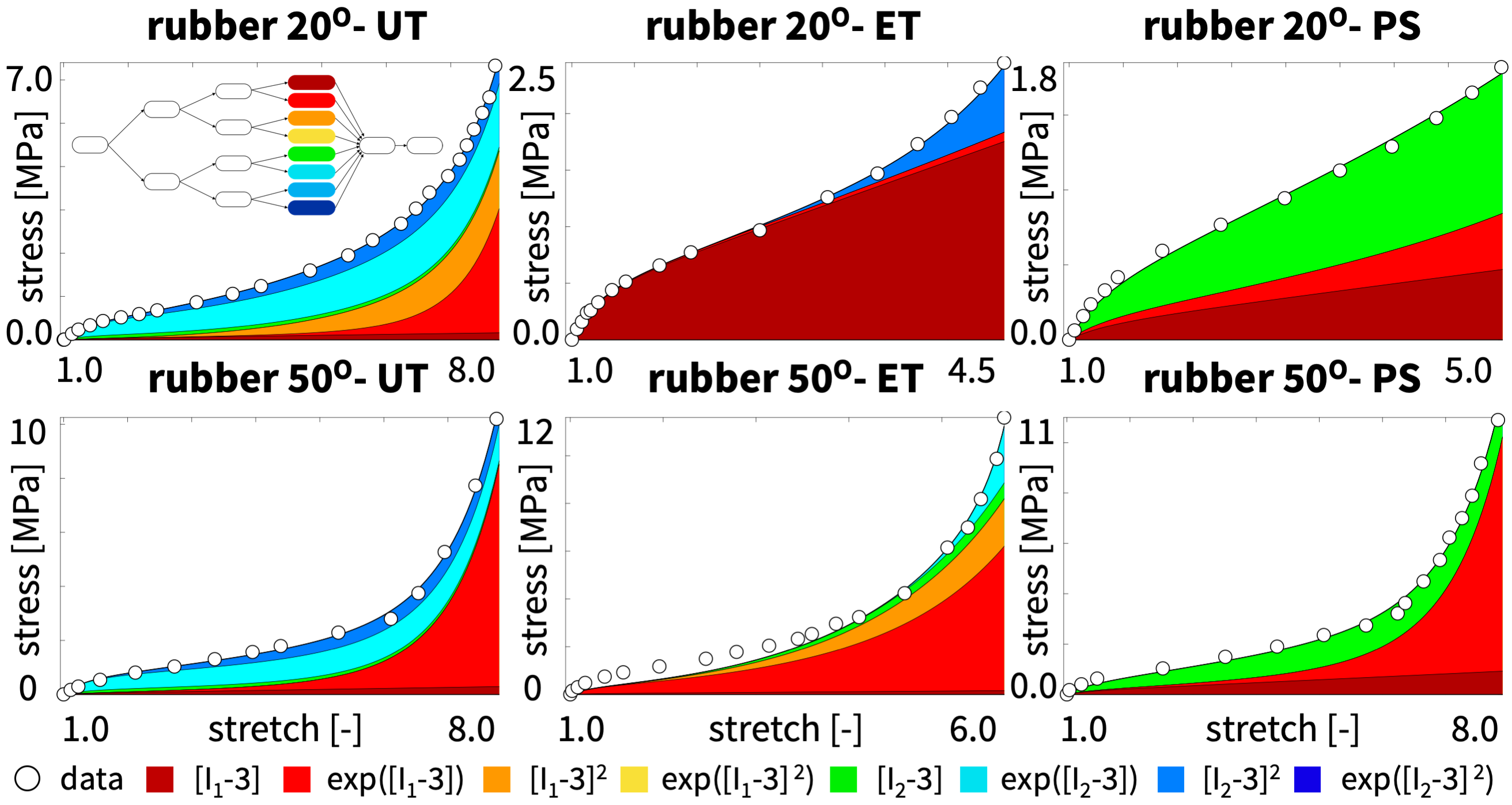}
\caption{{\bf{\sffamily{Constitutive Artificial Neural Network. Uniaxial tension, equibiaxial tension, and pure shear.}}} 
Piola stress $P(\lambda)$ as a function of the stretch $\lambda$ for the 
feed forward Constitutive Artificial Neural Network from Figure \ref{fig05}
with two layers, eight nodes, and twelve weights using the custom-designed activation functions from Figure \ref{fig04}.
Dots illustrate the uniaxial tension, equibiaxial tension, and pure shear data $\hat{P}$ for rubber at 20$^{\circ}$ and 50$^{\circ}$ \cite{treloar44} from Table \ref{tab02}; color-coded areas highlight the contributions of the color-coded nodes to the final stress function $P(\lambda)$ for individual single-mode training.}
\label{fig12}
\end{figure}

\begin{figure}[h]
\centering
\includegraphics[width=0.72\linewidth]{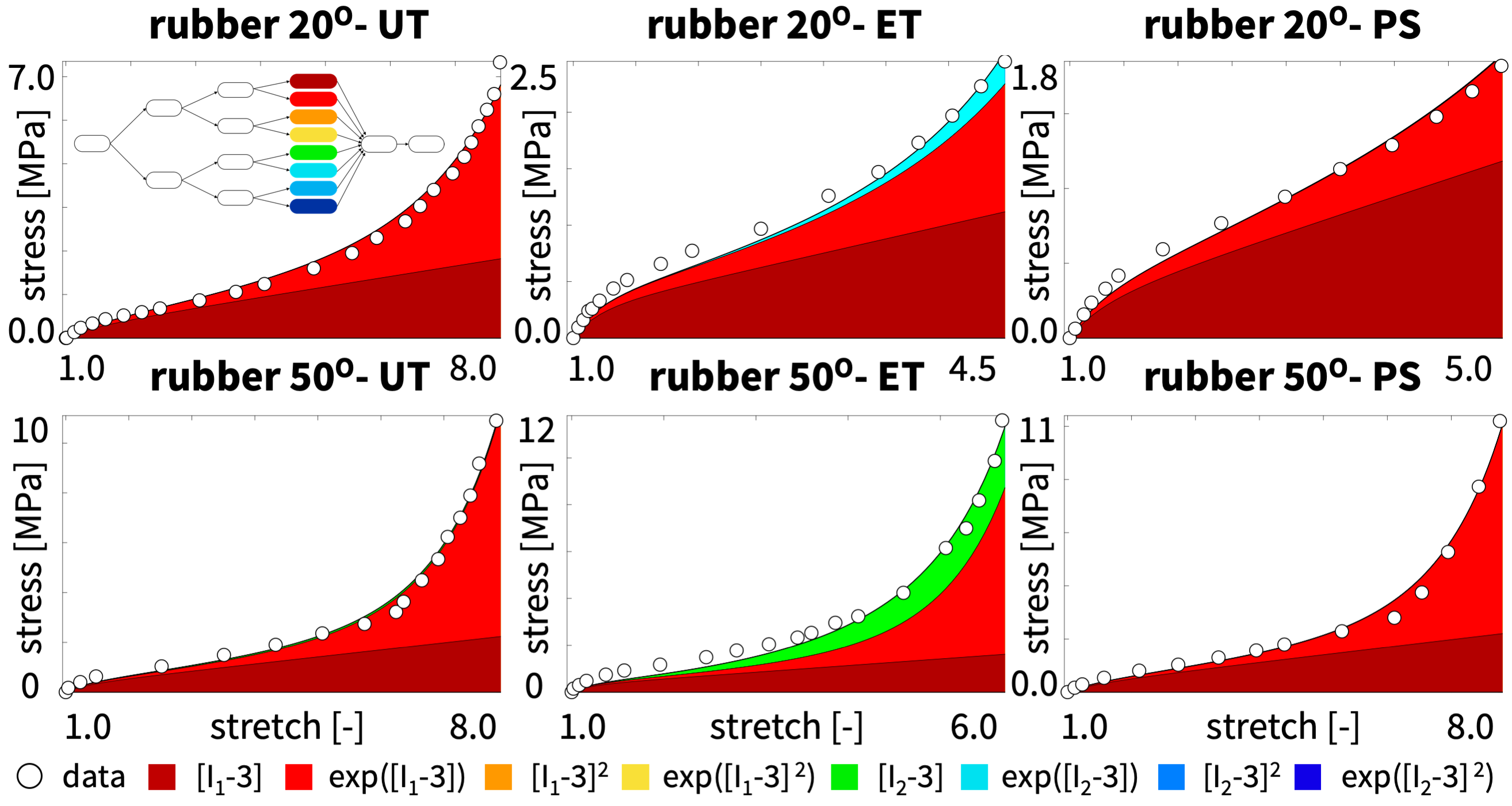}
\caption{{\bf{\sffamily{Constitutive Artificial Neural Network. Uniaxial tension, equibiaxial tension, and pure shear.}}} 
Piola stress $P(\lambda)$ as a function of the stretch $\lambda$ for the 
feed forward Constitutive Artificial Neural Network from Figure \ref{fig05}
with two layers, eight nodes, and twelve weights using the custom-designed activation functions from Figure \ref{fig04}.
Dots illustrate the uniaxial tension, equibiaxial tension, and pure shear data $\hat{P}$ for rubber at 20$^{\circ}$ and 50$^{\circ}$ \cite{treloar44} from Table \ref{tab02}; color-coded areas highlight the contributions of the color-coded nodes to the final stress function $P(\lambda)$ for simultaneous multi-mode training.}
\label{fig13}
\end{figure}

%%%%%%%%%%%%%%%%%%%%%%%%%%%%%%%%%%%%%%%%%%%%%%%%%%%%%%%%%%%%%%%%%%%%%%%%
\noindent
To demonstrate the performance of our new family of Constitutive Artificial Neural Networks, we perform a systematic side-by-side comparison of classical Neural Networks and Constitutive Neural Networks using widely-used benchmark data for rubber elasticity. Specifically, we train and compare the fully connected two-layer eight-term Neural Network from Figure \ref{fig01} and the two-layer eight-term Constitutive Artificial Neural Networks for isotropic perfectly incompressible materials from Figure \ref{fig05}. We consider two training scenarios, {\it{single-mode training}} and {\it{multi-mode training}}, for the special homogeneous deformation modes of uniaxial tension, biaxial tension, and pure shear. 
%%%%%%%%%%%%%%%%%%%%%%%%%%%%%%%%%%%%%%%%%%%%%%%%%%%%%%%%%%%%%%%%%%%%%%%%
\begin{table}[h]
\centering
\caption{{\bf{\sffamily{Benchmark stress-stretch data for single-mode training.}}} 
Uniaxial tension (UT) experiments
for rubber at 20$^{\circ}$ and 50$^{\circ}$ \cite{treloar44}, 
gum stock and tread stock \cite{mooney40}, and
polymeric foam and rubber \cite{blatz62}. 
All reported stresses are converted from their initial units 
[kg/cm$^2$] \cite{treloar44},
[kg/$2.5\cdot3.2$mm$^2$] \cite{mooney40}, and
[Psi] \cite{blatz62} into the unified unit [MPa].}
\vspace*{0.2cm} 
\footnotesize
\renewcommand{\arraystretch}{0.9}
\label{tab01}
\begin{tabular}{|c|c||c|c||c|c|
                |c|c||c|c||c|c|} \hline               
%%%%%%%%%%%%%%%%%%%%%%%%%%%%%%%%%%%%%%%%%%%%%%%%%%%%%%%%%%%%%%%%%%%%%%%%
  \multicolumn{2}{|c||}{\bf{\sffamily{UT}}}
& \multicolumn{2} {c||}{\bf{\sffamily{UT}}} 
& \multicolumn{2} {c||}{\bf{\sffamily{UT}}} 
& \multicolumn{2} {c||}{\bf{\sffamily{UT}}} 
& \multicolumn{2} {c||}{\bf{\sffamily{UT}}} 
& \multicolumn{2} {c|} {\bf{\sffamily{UT}}} \\ 
  \multicolumn{2}{|c||}{\bf{\sffamily{rubber 20$^{\circ}$}}}
& \multicolumn{2} {c||}{\bf{\sffamily{rubber 50$^{\circ}$}}}
& \multicolumn{2} {c||}{\bf{\sffamily{gum stock}}} 
& \multicolumn{2} {c||}{\bf{\sffamily{tread stock}}} 
& \multicolumn{2} {c||}{\bf{\sffamily{foam}}} 
& \multicolumn{2} {c|} {\bf{\sffamily{rubber}}} \\
  \multicolumn{2}{|c||}{Treloar \cite{treloar44}}
& \multicolumn{2} {c||}{Treloar \cite{treloar44}} 
& \multicolumn{2} {c||}{Mooney  \cite{mooney40}} 
& \multicolumn{2} {c||}{Mooney  \cite{mooney40}} 
& \multicolumn{2} {c||}{Blatz Ko \cite{blatz62}} 
& \multicolumn{2} {c|} {Blatz Ko \cite{blatz62}} \\ \hline \hline
%%%%%%%%%%%%%%%%%%%%%%%%%%%%%%%%%%%%%%%%%%%%%%%%%%%%%%%%%%%%%%%%%%%%%%%%
  $\lambda$ & P & $\lambda$ & P & $\lambda$ & P 
& $\lambda$ & P & $\lambda$ & P & $\lambda$ & P  \\
  \;[-]\; & [MPa] & \;[-]\; & [MPa] & \;[-]\; & [MPa] 
& \;[-]\; & [MPa] & \;[-]\; & [MPa] & \;[-]\; & [MPa]  \\ \hline \hline
%%%%%%%%%%%%%%%%%%%%%%%%%%%%%%%%%%%%%%%%%%%%%%%%%%%%%%%%%%%%%%%%%%%%%%%%
1.00  &  0.00  &  1.00  &  0.00  &  1.00  &  0.00  &  1.00  &  0.00  &  1.00  &  0.00  &  1.00  &  0.00 \\
1.01  &  0.00  &  1.11  &  0.17  &  1.46  &  0.31  &  1.16  &  0.31  &  1.05  &  0.04  &  1.05  &  0.03 \\
1.13  &  0.14  &  1.23  &  0.29  &  2.30  &  0.61  &  1.50  &  0.61  &  1.10  &  0.06  &  1.10  &  0.07 \\
1.23  &  0.24  &  1.57  &  0.54  &  4.66  &  1.23  &  2.56  &  1.23  &  1.15  &  0.07  &  1.16  &  0.10 \\
1.41  &  0.33  &  2.12  &  0.80  &  6.45  &  1.84  &  3.30  &  1.84  &  1.20  &  0.09  &  1.22  &  0.13 \\
1.61  &  0.43  &  2.73  &  1.03  &  6.77  &  2.45  &  3.53  &  2.45  &  1.30  &  0.12  &  1.27  &  0.16 \\
1.89  &  0.52  &  3.36  &  1.30  &  6.96  &  3.06  &  3.63  &  3.06  &  1.40  &  0.14  &  1.31  &  0.18 \\
2.17  &  0.59  &  3.95  &  1.57  &    &    &  3.71  &  3.68  &  1.50  &  0.16  &  1.37  &  0.20 \\
2.45  &  0.68  &  4.39  &  1.79  &    &    &    &    &  1.60  &  0.16  &  1.41  &  0.22 \\
3.06  &  0.87  &  5.29  &  2.29  &    &    &    &    &  1.70  &  0.17  &  1.47  &  0.24 \\
3.62  &  1.06  &  6.11  &  2.80  &    &    &    &    &  1.80  &  0.18  &  1.52  &  0.26 \\
4.06  &  1.24  &  6.54  &  3.75  &    &    &    &    &  1.90  &  0.19  &  1.57  &  0.27 \\
4.82  &  1.60  &  6.95  &  5.27  &    &    &    &    &  2.00  &  0.20  &  1.62  &  0.29 \\
5.41  &  1.95  &  7.43  &  7.73  &    &    &    &    &  2.10  &  0.20  &    &   \\
5.79  &  2.30  &  7.76  &  \!\!\!10.21  &    &    &    &    &  2.20  &  0.21  &    &   \\
6.23  &  2.68  &    &    &    &    &    &    &  2.30  &  0.21  &    &   \\
6.46  &  3.03  &    &    &    &    &    &    &  2.34  &  0.21  &    &   \\
6.67  &  3.40  &    &    &    &    &    &    &    &    &    &   \\
6.96  &  3.78  &    &    &    &    &    &    &    &    &    &   \\
7.14  &  4.16  &    &    &    &    &    &    &    &    &    &   \\
7.25  &  4.49  &    &    &    &    &    &    &    &    &    &   \\
7.36  &  4.86  &    &    &    &    &    &    &    &    &    &   \\
7.49  &  5.24  &    &    &    &    &    &    &    &    &    &   \\
7.60  &  5.60  &    &    &    &    &    &    &    &    &    &   \\
7.69  &  6.33  &    &    &    &    &    &    &    &    &    &   \\ \hline
%%%%%%%%%%%%%%%%%%%%%%%%%%%%%%%%%%%%%%%%%%%%%%%%%%%%%%%%%%%%%%%%%%%
\end{tabular}
\end{table} %\\[6.pt]
%%%%%%%%%%%%%%%%%%%%%%%%%%%%%%%%%%%%%%%%%%%%%%%%%%%%%%%%%%%%%%%%%%%
%%%%%%%%%%%%%%%%%%%%%%%%%%%%%%%%%%%%%%%%%%%%%%%%%%%%%%%%%%%%%%%%%%%
\begin{table}[h]
\centering
\caption{{\bf{\sffamily{Benchmark stress-stretch data for multi-mode training.}}} 
Uniaxial tension (UT), equibiaxial tension (ET), and pure shear (PS) experiments
for rubber at 20$^{\circ}$ and 50$^{\circ}$ \cite{treloar44}. 
Equibiaxial stresses are multiplied by their stretches and
all stresses are converted from their initial unit 
[kg/cm$^2$] into the unified unit [MPa].}
\vspace*{0.2cm} 
\footnotesize
\renewcommand{\arraystretch}{0.9}
\label{tab02}
\begin{tabular}{|c|c||c|c||c|c|
                |c|c||c|c||c|c|} \hline               
%%%%%%%%%%%%%%%%%%%%%%%%%%%%%%%%%%%%%%%%%%%%%%%%%%%%%%%%%%%%%%%%%%%
  \multicolumn{2}{|c||}{\bf{\sffamily{UT}}}
& \multicolumn{2} {c||}{\bf{\sffamily{ET}}} 
& \multicolumn{2} {c||}{\bf{\sffamily{PS}}} 
& \multicolumn{2} {c||}{\bf{\sffamily{UT}}} 
& \multicolumn{2} {c||}{\bf{\sffamily{ET}}} 
& \multicolumn{2} {c|} {\bf{\sffamily{PS}}} \\ 
  \multicolumn{2}{|c||}{\bf{\sffamily{rubber 20$^{\circ}$}}}
& \multicolumn{2} {c||}{\bf{\sffamily{rubber 20$^{\circ}$}}}
& \multicolumn{2} {c||}{\bf{\sffamily{rubber 20$^{\circ}$}}}
& \multicolumn{2} {c||}{\bf{\sffamily{rubber 50$^{\circ}$}}} 
& \multicolumn{2} {c||}{\bf{\sffamily{rubber 50$^{\circ}$}}} 
& \multicolumn{2} {c|} {\bf{\sffamily{rubber 50$^{\circ}$}}} \\
  \multicolumn{2}{|c||}{Treloar \cite{treloar44}}
& \multicolumn{2} {c||}{Treloar \cite{treloar44}} 
& \multicolumn{2} {c||}{Treloar \cite{treloar44}} 
& \multicolumn{2} {c||}{Treloar \cite{treloar44}} 
& \multicolumn{2} {c||}{Treloar \cite{treloar44}} 
& \multicolumn{2} {c|} {Treloar \cite{treloar44}} \\ \hline \hline
%%%%%%%%%%%%%%%%%%%%%%%%%%%%%%%%%%%%%%%%%%%%%%%%%%%%%%%%%%%%%%%%%%%%%%%%
  $\lambda$ & P & $\lambda$ & P & $\lambda$ & P 
& $\lambda$ & P & $\lambda$ & P & $\lambda$ & P  \\
  \;[-]\; & [MPa] & \;[-]\; & [MPa] & \;[-]\; & [MPa] 
& \;[-]\; & [MPa] & \;[-]\; & [MPa] & \;[-]\; & [MPa]  \\ \hline \hline
%%%%%%%%%%%%%%%%%%%%%%%%%%%%%%%%%%%%%%%%%%%%%%%%%%%%%%%%%%%%%%%%%%%%%%%%
1.00  &  0.00  &  1.00  &  0.00  &  1.00  &  0.00  &  1.00  &  0.00  &  1.00  &  0.00  &  1.00  &  0.00 \\
1.01  &  0.00  &  1.04  &  0.09  &  1.05  &  0.06  &  1.11  &  0.17  &  1.02  &  0.15  &  1.04  &  0.17 \\
1.13  &  0.14  &  1.08  &  0.16  &  1.13  &  0.16  &  1.23  &  0.29  &  1.08  &  0.30  &  1.23  &  0.40 \\
1.23  &  0.24  &  1.12  &  0.24  &  1.20  &  0.24  &  1.57  &  0.54  &  1.16  &  0.48  &  1.48  &  0.63 \\
1.41  &  0.33  &  1.15  &  0.26  &  1.33  &  0.33  &  2.12  &  0.80  &  1.37  &  0.74  &  2.52  &  1.03 \\
1.61  &  0.43  &  1.21  &  0.33  &  1.45  &  0.42  &  2.73  &  1.03  &  1.57  &  0.92  &  3.51  &  1.49 \\
1.89  &  0.52  &  1.32  &  0.44  &  1.86  &  0.59  &  3.36  &  1.30  &  1.96  &  1.17  &  4.33  &  1.90 \\
2.17  &  0.59  &  1.43  &  0.51  &  2.40  &  0.77  &  3.95  &  1.57  &  2.46  &  1.49  &  5.07  &  2.36 \\
2.45  &  0.68  &  1.70  &  0.66  &  2.99  &  0.95  &  4.39  &  1.79  &  2.79  &  1.78  &  5.74  &  2.74 \\
3.06  &  0.87  &  1.95  &  0.77  &  3.50  &  1.13  &  5.29  &  2.29  &  3.14  &  2.04  &  6.24  &  3.22 \\
3.62  &  1.06  &  2.50  &  0.97  &  3.98  &  1.29  &  6.11  &  2.80  &  3.45  &  2.33  &  6.36  &  3.63 \\
4.06  &  1.24  &  3.04  &  1.26  &  4.39  &  1.48  &  6.54  &  3.75  &  3.60  &  2.53  &  6.65  &  4.49 \\
4.82  &  1.60  &  3.44  &  1.47  &  4.72  &  1.65  &  6.95  &  5.27  &  3.86  &  2.96  &  6.91  &  5.34 \\
5.41  &  1.95  &  3.76  &  1.73  &  4.99  &  1.82  &  7.43  &  7.73  &  4.11  &  3.24  &  7.06  &  6.23 \\
5.79  &  2.30  &  4.03  &  1.97  &    &    &  7.76  &  \!\!\!10.21  &  4.60  &  4.24  &  7.26  &  7.00 \\
6.23  &  2.68  &  4.26  &  2.23  &    &    &    &    &  5.06  &  6.15  &  7.42  &  7.89 \\
6.46  &  3.03  &  4.45  &  2.45  &    &    &    &    &  5.28  &  6.99  &  7.56  &  9.18 \\
6.67  &  3.40  &    &    &    &    &    &    &  5.42  &  8.18  &  7.83  &  \!\!\!10.90 \\
6.96  &  3.78  &    &    &    &    &    &    &  5.59  &  9.87  &    &   \\
7.14  &  4.16  &    &    &    &    &    &    &  5.67  &  \!\!\!11.59  &    &   \\
7.25  &  4.49  &    &    &    &    &    &    &    &    &    &   \\
7.36  &  4.86  &    &    &    &    &    &    &    &    &    &   \\
7.49  &  5.24  &    &    &    &    &    &    &    &    &    &   \\
7.60  &  5.60  &    &    &    &    &    &    &    &    &    &   \\
7.69  &  6.33  &    &    &    &    &    &    &    &    &    &   \\ \hline
%%%%%%%%%%%%%%%%%%%%%%%%%%%%%%%%%%%%%%%%%%%%%%%%%%%%%%%%%%%%%%%%%%%%%%%%
\end{tabular}
\end{table} \\[6.pt]
%%%%%%%%%%%%%%%%%%%%%%%%%%%%%%%%%%%%%%%%%%%%%%%%%%%%%%%%%%%%%%%%%%%%%%%%
Table~\ref{tab01} summarizes our benchmark data for {\it{single-mode training}} from uniaxial tension experiments for rubber at 20$^{\circ}$ and 50$^{\circ}$ \cite{treloar44}, for gum stock and tread stock \cite{mooney40}, and for polymeric foam and rubber \cite{blatz62}. For comparison, we have converted all reported stresses from their initial units 
[kg/cm$^2$] \cite{treloar44},
[kg/$2.5\cdot3.2$mm$^2$] \cite{mooney40}, and
[Psi] \cite{blatz62} into the unified unit [MPa].
%%%
Table~\ref{tab02} summarizes our benchmark data for {\it{multi-mode training}} from uniaxial tension, equibiaxial tension, and pure shear experiments
for rubber at 20$^{\circ}$ and 50$^{\circ}$ \cite{treloar44}. 
For comparison, we have multiplied the equibiaxial stresses by their stretches and converted all reported stresses from their initial unit [kg/cm$^2$] into the unified unit [MPa].\\[6.pt]
%%%%%%%%%%%%%%%%%%%%%%%%%%%%%%%%%%%%%%%%%%%%%%%%%%%%%%%%%%%%%%%%%%%%%%%%
\noindent
{\bf{\sffamily{Classical Neural Networks can describe data well but cannot predict beyond the training regime.}}} %Effect of network depth and breadth.}}}  
%%%%%%%%%%%%%%%%%%%%%%%%%%%%%%%%%%%%%%%%%%%%%%%%%%%%%%%%%%%%%%%%%%%%%%%%
Figure~\ref{fig07} illustrates the effect of network depth and breadth  
for six classical fully connected feed forward Neural Networks with one and two layers and two, four, and eight nodes. 
%%%%%%%%%%%%%%%%%%%%%%%%%%%%%%%%%%%%%%%%%%%%%%%%%%%%%%%%%%%%%%%%%%%%%%%%
%\\[6.pt]
%%%%%%%%%%%%%%%%%%%%%%%%%%%%%%%%%%%%%%%%%%%%%%%%%%%%%%%%%%%%%%%%%%%%%%%%
The number of network weights and biases increases with increasing number of layers and nodes: 
The simplest model with one hidden layer and two nodes has 
$n_{\scas{w}} = 2+2 = 4$ weights and 
$n_{\scas{b}} = 2+1 = 3$ biases 
and a total number of $n_{\theta}=7$ network parameters;
the most complex model with two hidden layers and eight nodes has 
$n_{\scas{w}} = 8+8\times8+8 = 80$ weights and 
$n_{\scas{b}} = 8+8+1 =17$ biases 
and a total number of $n_{\theta}=97$ network parameters.
For this example, for all nodes, we use the hyperbolic tangent activation function according to Figure \ref{fig02}. Specifically, the network with two layers and two nodes uses the set of equations (\ref{neuralnetwork_tanh}).  
The networks learn the approximation of the Piola stress $P(\lambda)$ as a function of the stretch $\lambda$ using the uniaxial tension data $\hat{P}$ for rubber at 20$^{\circ}$ \cite{treloar44} from Tables \ref{tab01} and \ref{tab02}.
The dots illustrate the training data $\hat{P}$ and the color-coded areas highlight the contributions of the color-coded nodes to the final stress function $P(\lambda)$.
%%% result %%%
First and foremost, all six networks robustly approximate the stress $P(\lambda)$ as a function of the stretch $\lambda$ with virtually no error compared to the dots of the experimental data $\hat{P}$. In general, the cost of training a Neural Network increases with the number of nodes per layer and with the number of layers. Similar to a mesh refinement in a finite element analysis, in the spirit of h-adaptivity, we expect the approximation to improve with increasing network breadth and depth. 
The dots in Figure~\ref{fig07} indicate that the behavior of rubber under uniaxial tension is nonlinear, but monotonic and fairly smooth \cite{treloar44}. 
As a result, all six networks perform exceptionally well at {\it{describing}} or {\it{interpolating}} the data within the training regime of $1 \le \lambda \le 8$, even the simplest network with only one layer and two nodes. However, all six networks do a poor job at {\it{predicting}} or {\it{extrapolating}} the behavior outside the training regime for $\lambda > 8$.  \\[6.pt]
%%%%%%%%%%%%%%%%%%%%%%%%%%%%%%%%%%%%%%%%%%%%%%%%%%%%%%%%%%%%%%%%%%%%%%%%
\noindent
{\bf{\sffamily{Classical Neural Networks perform well for big data but tend to overfit sparse data.}}}
%%%%%%%%%%%%%%%%%%%%%%%%%%%%%%%%%%%%%%%%%%%%%%%%%%%%%%%%%%%%%%%%%%%%%%%%
Figure \ref{fig08} illustrates the performance of classical Neural Networks for different uniaxial tension data. For this example, we use a fully connected feed forward Neural Network with one layer, eight nodes, 16 weights, nine biases, and the hyperbolic tangent activation function from Figure \ref{fig02} for all nodes. 
%%%%%%%%%%%%%%%%%%%%%%%%%%%%%%%%%%%%%%%%%%%%%%%%%%%%%%%%%%%%%%%%%%%%%%%%
%\\[6.pt]
%%%%%%%%%%%%%%%%%%%%%%%%%%%%%%%%%%%%%%%%%%%%%%%%%%%%%%%%%%%%%%%%%%%%%%%%
The network learns the approximation of the Piola stress $P(\lambda)$ as a function of the stretch $\lambda$ using the uniaxial tension data $\hat{P}$ for
rubber at 20$^{\circ}$ and 50$^{\circ}$ \cite{treloar44}, 
gum stock and tread stock \cite{mooney40}, and
polymeric foam and rubber \cite{blatz62} from Table \ref{tab01}.
The dots illustrate the training data $\hat{P}$ and the color-coded areas highlight the contributions of the color-coded nodes to the final stress function $P(\lambda)$.
%%% result %%%
In general, our observations agree with Figure \ref{fig07} and suggest that classical Neural Networks robustly interpolate uniaxial tension data for rubber for all six experiments. However, for the example of gum stock with only seven data points and $n_{\theta}=25$ network parameters, we observe oscillations in the approximated stress function $P(\lambda)$ in the center region between $2.4 \le \lambda \le 6.4$, where we only have one data point. These oscillations are a result of {\it{negative weights}} in the final output layer that make the approximated function {\it{non-convex}}. While this single example is by no means a rigorous mathematical proof, it supports the general notion that classical Neural Networks {\it{fit big data well}} but {\it{tend to overfit sparse data}}.\\[6.pt]
%%%%%%%%%%%%%%%%%%%%%%%%%%%%%%%%%%%%%%%%%%%%%%%%%%%%%%%%%%%%%%%%%%%%%%%%
\noindent
{\bf{\sffamily{Classical Neural Networks perform well for multi-mode data but provide no physical insight.}}}
%%%%%%%%%%%%%%%%%%%%%%%%%%%%%%%%%%%%%%%%%%%%%%%%%%%%%%%%%%%%%%%%%%%%%%%%
Figure \ref{fig09} illustrates the performance of classical Neural Networks for multi-mode data, trained {\it{individually}} for each mode. 
Similar to the previous example, we use a fully connected feed forward Neural Network with one layer, eight nodes, 16 weights, nine biases, and the hyperbolic tangent activation function from Figure \ref{fig02} for all nodes. 
%%%%%%%%%%%%%%%%%%%%%%%%%%%%%%%%%%%%%%%%%%%%%%%%%%%%%%%%%%%%%%%%%%%%%%%%
%\\[6.pt]
%%%%%%%%%%%%%%%%%%%%%%%%%%%%%%%%%%%%%%%%%%%%%%%%%%%%%%%%%%%%%%%%%%%%%%%%
The network learns the approximation of the Piola stress $P(\lambda)$ as a function of the stretch $\lambda$ and trains {\it{individually}} on the uniaxial tension, equibiaxial tension, and pure shear data for rubber at 20$^{\circ}$ and 50$^{\circ}$ \cite{treloar44} from Table \ref{tab02}.
The dots illustrate the training data $\hat{P}$ and the color-coded areas highlight the contributions of the color-coded nodes to the final stress function $P(\lambda)$.
%%% result %%%
The network performs robustly on all six training sets and generates stress approximations $P(\lambda)$ that fit the stress-stretch data well, even for the S-shaped curves and in the presence of pronounced stretch stiffening. For all six cases, the loss function rapidly decreases by four orders of magnitude within less than 20,000 epochs and the error between model $P(\lambda)$ and data $\hat{P}$ is virtually invisible from the graphs. 
The full color spectrum in each graph suggests that all eight nodes contribute to the final stress approximation and that all weights between the last hidden layer and the output layer are non-zero. We conclude that we can robustly learn the $n_{\theta}=25$ network weights and biases from multi-modal training data; yet, these parameters have no physical meaning and do not contribute to {\it{interpreting}} or {\it{explaining}} the physics of rubber under uniaxial tension, equibiaxial tension, or pure shear.\\[6.pt]
%%%%%%%%%%%%%%%%%%%%%%%%%%%%%%%%%%%%%%%%%%%%%%%%%%%%%%%%%%%%%%%%%%%%%%%%
\noindent
{\bf{\sffamily{Constitutive Artificial Neural Networks describe and predict well and prevent overfitting.}}}
%%%%%%%%%%%%%%%%%%%%%%%%%%%%%%%%%%%%%%%%%%%%%%%%%%%%%%%%%%%%%%%%%%%%%%%%
Figure \ref{fig10} demonstrates the performance of our new class of Constitutive Artificial Neural Networks for different uniaxial tension data. For this example, we use the feed forward Constitutive Artificial Neural Network from Figure \ref{fig05}
with two layers, eight nodes, and twelve weights using the custom-designed activation functions from Figure \ref{fig04}.
%%%%%%%%%%%%%%%%%%%%%%%%%%%%%%%%%%%%%%%%%%%%%%%%%%%%%%%%%%%%%%%%%%%%%%%%
%\\[6.pt]
%%%%%%%%%%%%%%%%%%%%%%%%%%%%%%%%%%%%%%%%%%%%%%%%%%%%%%%%%%%%%%%%%%%%%%%%
The network learns the approximation of the free energy as a function of the invariants $\psi(I_1,I_2)$, where pre-processing generates the invariants as functions of the stretch $I_1(\lambda)$, $I_2(\lambda)$, and post-processing generates the stress as a function of the free energy $P(\psi)$. The network trains on the uniaxial tension data 
$\hat{P}$ for
rubber at 20$^{\circ}$ and 50$^{\circ}$ \cite{treloar44}, 
gum stock and tread stock \cite{mooney40}, and
polymeric foam and rubber \cite{blatz62} from Table \ref{tab01}.
The dots illustrate the training data $\hat{P}$ and the color-coded areas highlight the contributions of the color-coded nodes to the final stress function $P(\lambda)$.
%%% result %%%
First and foremost, similar to the classical Neural Network in Figure \ref{fig08}, the new Constitutive Artificial Neural Network in Figure \ref{fig10} performs robustly on all six training sets and learns stress functions $P(\lambda)$ that approximate the stress-stretch data well, even for S-shaped curves and in the presence of pronounced stretch stiffening. For all six cases, the loss function rapidly decreases by four orders of magnitude within less than 10,000 epochs and the error between model $P(\lambda)$ and data $\hat{P}$ is virtually invisible from the graphs. In contrast to the Neural Network example in Figure \ref{fig07} where the learned stresses flatline abruptly outside the training regime, all six stress approximations continue smoothly beyond the initial training regime. In contrast to the gum stock example with only seven data points in Figure \ref{fig08}, the Constitutive Artificial Neural Network generates smooth non-oscillatory stresses $P(\lambda)$, even in regions with sparse data. These observations suggest that our new Constitutive Artificial Neural Networks succeed at {\it{describing}}, {\it{predicting}}, and {\it{preventing overfitting}}, even in regions where data are sparse. \\[6.pt]
%%%%%%%%%%%%%%%%%%%%%%%%%%%%%%%%%%%%%%%%%%%%%%%%%%%%%%%%%%%%%%%%%%%%%%%%
\noindent
{\bf{\sffamily{Constitutive Artificial Neural Networks generate non-unique solutions for insufficiently rich data.}}}
%%%%%%%%%%%%%%%%%%%%%%%%%%%%%%%%%%%%%%%%%%%%%%%%%%%%%%%%%%%%%%%%%%%%%%%%
Figure~\ref{fig11} illustrates the effect of the initial conditions for the feed forward Constitutive Artificial Neural Network from Figure~\ref{fig05}
with two layers, eight nodes, and twelve weights, and the custom-designed activation functions from Figure~\ref{fig04}. 
%%%%%%%%%%%%%%%%%%%%%%%%%%%%%%%%%%%%%%%%%%%%%%%%%%%%%%%%%%%%%%%%%%%%%%%%
%\\[6.pt]
%%%%%%%%%%%%%%%%%%%%%%%%%%%%%%%%%%%%%%%%%%%%%%%%%%%%%%%%%%%%%%%%%%%%%%%%
For this example, we initialize the twelve network weights with six different sets of randomly generated numbers and compare their contributions to the final stress approximation $P(\lambda)$ as an indicator for the magnitude of the learned weights. The dots indicate the uniaxial tension data $\hat{P}$ for 
rubber at 20$^{\circ}$ \cite{treloar44} from Table \ref{tab01}, and the color-coded areas highlight the contributions of the color-coded nodes to the stress functions $P(\lambda)$ for the six different sets of initial conditions.
%%% results %%%
First and most importantly, within less than 10,000 epochs, all six sets of initial conditions robustly converge towards a set of weights that reduce the loss function by more than four orders of magnitude and interpolate the training equally data well. 
Interestingly, in contrast to the classical Neural Network graphs in Figures \ref{fig08} and \ref{fig09}, none of the six graphs in Figure \ref{fig11} covers the full color spectrum. This suggests that only a subset of the eight nodes of the last hidden layer contribute to the final stress approximation, while most of the weights between the last hidden layer and the output layer train to zero. For example, the fourth graph approximates the stress exclusively in terms of the third and fifth terms, $[\,I_1-3\,]^2$ and $[I_2-3]$, whereas the fifth graph uses the first, sixths, and eights terms, $[I_1-3]$, $[\rm{exp}([I_2-3])-1]$, and , $[\rm{exp}([I_2-3]^2)-1]$. From comparing the curves and the colored stress contributions in all six graphs, we conclude that the selection of weights that best approximate the stress-stretch relation is {\it{non-unique}}. While this is also true and well-known for classical Neural Networks, it is unfortunate for Constitutive Artificial Neural Networks since we attempt to correlate the network weights to constitutive parameters with a clear physical interpretation. It seems natural to ask whether this non-uniqueness is an inherent property of the Constitutive Artificial Neural Network itself or rather a result of insufficiently rich training data.  \\[6.pt]
%%%%%%%%%%%%%%%%%%%%%%%%%%%%%%%%%%%%%%%%%%%%%%%%%%%%%%%%%%%%%%%%%%%%%%%%
\noindent
{\bf{\sffamily{Constitutive Artificial Neural Networks are a natural generalization of existing constitutive models.}}}
%%%%%%%%%%%%%%%%%%%%%%%%%%%%%%%%%%%%%%%%%%%%%%%%%%%%%%%%%%%%%%%%%%%%%%%%
Figure~\ref{fig12} illustrates the performance of Constitutive Artificial Neural Networks
for multi-mode data, trained {\it{individually}} for each mode.
Similar to the previous two examples, we use the feed forward Constitutive Artificial Neural Network from Figure \ref{fig05} with two layers, eight nodes, and twelve weights using the custom-designed activation functions from Figure \ref{fig04}.
%%%%%%%%%%%%%%%%%%%%%%%%%%%%%%%%%%%%%%%%%%%%%%%%%%%%%%%%%%%%%%%%%%%%%%%%
%\\[6.pt]
%%%%%%%%%%%%%%%%%%%%%%%%%%%%%%%%%%%%%%%%%%%%%%%%%%%%%%%%%%%%%%%%%%%%%%%%
The network learns the approximation of the free energy as a function of the invariants $\psi(I_1,I_2)$ and trains {\it{individually}} on the uniaxial tension, equibiaxial tension, and pure shear data for rubber at 20$^{\circ}$ and 50$^{\circ}$ \cite{treloar44} from Table \ref{tab02}.
%%% result %%%
Similar to the classical Neural Network in Figure \ref{fig09}, the Constitutive Artificial Neural Network in Figure \ref{fig12} performs robustly on all six training sets and generates stress functions $P(\lambda)$ that approximate the stress-stretch data $\hat{P}$  well, even for the S-shaped curves and in the presence of pronounced stretch stiffening. 
Similar to the previous example, none of the six graphs in Figure \ref{fig12} covers the full color spectrum and most of the weights between the last hidden layer and the output layer train to zero.
Interestingly, some of the non-zero terms correlate well with the widely-used constitutive models for rubber elasticity: 
The dominant dark red $[\,I_1-3\,]$ term for the 20$^{\circ}$ equibiaxial tension data correlates well with the classical {\it{neo Hooke model}} \cite{treloar48} in equation (\ref{neohooke}), 
the dominant green $[\,I_1-2\,]$ term for the 20$^{\circ}$ pure shear data correlates well with the {\it{Blatz Ko model}} \cite{blatz62} in equation (\ref{blatzko}), 
the interacting $[\,I_1-1\,]$ and $[\,I_1-2\,]$ terms for the 20$^{\circ}$ and 50$^{\circ}$ pure shear data correlate well with the {\it{Mooney Rivlin model}} \cite{mooney40,rivlin48} in equation (\ref{mooney}), and
the dominant $[\rm{exp}([I_1-3])]$ term for the 50$^{\circ}$ uniaxial and equibiaxial tension data correlates well with the {\it{Demiray model}} \cite{demiray72} in equation (\ref{demiray}).
This suggests that Constitutive Artificial Neural Networks are a {\it{generalization}} of existing constitutive models that naturally self-select terms from subsets of well-known  constitutive models that best explain the data. \\[6.pt]
%%%%%%%%%%%%%%%%%%%%%%%%%%%%%%%%%%%%%%%%%%%%%%%%%%%%%%%%%%%%%%%%%%%%%%%%
\noindent
{\bf{\sffamily{Constitutive Artificial Neural Networks 
identify a single unique model and parameter set for sufficient data.}}}
%%%%%%%%%%%%%%%%%%%%%%%%%%%%%%%%%%%%%%%%%%%%%%%%%%%%%%%%%%%%%%%%%%%%%%%%
Figure \ref{fig13} illustrates the performance of Constitutive Artificial Neural Networks
for multi-mode data, trained {\it{simultaneously}} for all three modes.
Similar to the previous examples, we use the feed forward Constitutive Artificial Neural Network from Figure \ref{fig05} with two layers, eight nodes, and twelve weights using the custom-designed activation functions from Figure \ref{fig04}.
%%%%%%%%%%%%%%%%%%%%%%%%%%%%%%%%%%%%%%%%%%%%%%%%%%%%%%%%%%%%%%%%%%%%%%%%
%\\[6.pt]
%%%%%%%%%%%%%%%%%%%%%%%%%%%%%%%%%%%%%%%%%%%%%%%%%%%%%%%%%%%%%%%%%%%%%%%%
The network learns the approximation of the free energy as a function of the invariants $\psi(I_1,I_2)$ and trains {\it{simultaneously}} on the uniaxial tension, equibiaxial tension, and pure shear data for rubber at 20$^{\circ}$ and 50$^{\circ}$ \cite{treloar44} from Table \ref{tab02}.
%%% result %%%
Overall, the network trains robustly and uniquely for multi-mode data, both for the 20$^{\circ}$ and the 50$^{\circ}$ training sets. It is insensitive to the initial conditions and repeatedly converges towards the same set of weights to reduce the loss function by more than four orders of magnitude within less than 10,000 epochs.
Similar to the other Constitutive Artificial Neural Network examples, and in contrast to the classical Neural Network, the final approximation uses only a subset of non-zero weights, while most of the weights are zero. 
Compared to the individual single-mode training in Figure \ref{fig12}, the simultaneous multi-mode training in Figure \ref{fig13} seeks to approximate all three deformation modes simultaneously at the cost of a perfect fit: While the stress approximation $P(\lambda)$ slightly underestimates the training stress $\hat{P}$ in equibiaxial tension, it slightly overestimates the training stress $\hat{P}$ in the stiffening region in uniaxial tension and pure shear. 
%but(!) it is biased by the number of data points, e.g., more data in UT, then ET fit is less precise, we can fix this, e.g., by weighting terms in the loss function \\
%right balance of model complexity and data.
Most importantly though, the Constitutive Artificial Neural Network robustly identifies one {\it{unique model and parameter set}} for rubber at 20$^{\circ}$ and one set for rubber at 50$^{\circ}$. 
%%%%
For the low-temperature regime, the free energy reduces to a three-term function in terms of the first invariant
%similar to the neo Hooke model in equation (\ref{neohooke}), 
and the linear exponentials of the first and second invariants,
%similar to the Demiray model in equation (\ref{demiray}),
%%%%%%%%%%%%%%%%%%%%%%%%%%%%%%%%%%%%%%%%%%%%%%%%%%%%%%%%%%%%%%%%%%%%%%%%
\beq
  \psi(I_1,I_2)
= \frac{1}{2} \, \mu_1 \,[\,I_1 - 3\,]
+ \frac{1}{2} \frac{a_1}{b_1} \,[ \, \exp (\,  b_1 [\, I_1 -3 \,])   - 1\,] 
+ \frac{1}{2} \frac{a_2}{b_2} \,[ \, \exp (\,  b_2 [\, I_2 -3 \,])   - 1\,] \,.
\label{rubber20}
\eeq
%%%%%%%%%%%%%%%%%%%%%%%%%%%%%%%%%%%%%%%%%%%%%%%%%%%%%%%%%%%%%%%%%%%
It introduces five network weights that translate into five physically meaningful parameters with well-defined physical units, 
the shear modulus,
$\mu_1 = 2 \, w_{1,1} w_{2,1} = 0.2370$\,MPa, 
the stiffness-like parameters 
$a_1   = 2 \, w_{1,2} w_{2,2} = 0.0582$\,MPa and
$a_2   = 2 \, w_{1,6} w_{2,6} = 0.0013$\,MPa,
and the unit-less exponential coefficients
$b_1 = w_{1,2} = 0.0387$ and
$b_2 = w_{1,6} = 0.0022$.
%%%%%%
%%derived from
%%$\frac{1}{2} \, \mu_1 = w_{1,1} w_{2,1} = 0.1185$\,MPa, 
%%the stiffness-like parameters 
%%$\frac{1}{2} \,{a_1}/{b_1} = w_{2,2} = 0.7525$\,MPa and
%%$\frac{1}{2} \,{a_2}/{b_2} = w_{2,6} = 0.2918$\,MPa,
%%and the exponential coefficients
%%$b_1 = w_{1,2} = 0.0387$ and
%%$b_2 = w_{1,6} = 0.0022$.
%%%%%%
For the high-temperature regime, the free energy reduces to a three-term function in terms of the first and second invariants
%similar to the Mooney Rivlin model in equation (\ref{neohooke}), 
and the linear exponential of the first invariant,
%similar to the Demiray model in equation (\ref{demiray}),
%%%%%%%%%%%%%%%%%%%%%%%%%%%%%%%%%%%%%%%%%%%%%%%%%%%%%%%%%%%%%%%%%%%%%%%%%
\beq 
  \psi(I_1,I_2)
= \frac{1}{2} \, \mu_1 \,[\,I_1 - 3\,]
+ \frac{1}{2} \, \mu_2 \,[\,I_1 - 3\,]
+ \frac{1}{2} \frac{a_1}{b_1} \,[ \, \rm{exp} (\,  b_1 [\, I_1 -3 \,] \,)   - 1\,] \,.
\label{rubber50}
\eeq
%%%%%%%%%%%%%%%%%%%%%%%%%%%%%%%%%%%%%%%%%%%%%%%%%%%%%%%%%%%%%%%%%%%%
It introduces four network weights that translate into four physically meaningful parameters with well-defined physical units, 
the shear moduli,
$\mu_1 = 2 \, w_{1,1} w_{2,1} = 0.2830$\,MPa and 
$\mu_2 = 2 \, w_{1,5} w_{2,5} = 0.0141$\,MPa, 
the stiffness-like parameter 
$a_1   = 2 \, w_{1,2} w_{2,2} = 0.0434$\,MPa, and
the unit-less exponential coefficient
$b_1 = w_{1,2} = 0.0541$.
%%%%
%derived from
%$\frac{1}{2} \, \mu = w_{1,1} w_{2,1} = 0.1415$
%$\frac{1}{2} \, \mu_2 = w_{5,1} w_{5,2} = 0.0070$ 
%$b_1 = w_{1,2} = 0.0541$  
%$\frac{1}{2} \,{a_1}/{b_1} = w_{2,2} = 0.4009$
%%%%
This example suggests that the non-uniqueness of the fit in Figure \ref{fig11} is not an inherent problem of Constitutive Artificial Neural Networks per se, but rather a problem of insufficiently rich data to appropriately train the network. With multi-mode data from uniaxial tension, biaxial tension, and pure shear, our Constitutive Artificial Neural Network {\it{trains robustly}} and {\it{uniquely and simultaneously learns both model and parameters}}. Interestingly, the training autonomously selects a subset of weights that activate the relevant terms to the free energy function, while the remaining weights train to zero. This suggests that Constitutive Artificial Neural Networks are capable of identifying a free energy function and its material parameters--out of a broad spectrum of functions and parameters--to best explain the data.
%%%%%%%%%%%%%%%%%%%%%%%%%%%%%%%%%%%%%%%%%%%%%%%%%%%%%%%%%%%%%%%%%%%%%%%%
\section{Discussion}\label{sec_discussion}
%%%%%%%%%%%%%%%%%%%%%%%%%%%%%%%%%%%%%%%%%%%%%%%%%%%%%%%%%%%%%%%%%%%%%%%%
%%%%%%%%%%%%%%%%%%%%%%%%%%%%%%%%%%%%%%%%%%%%%%%%%%%%%%%%%%%%%%%%%%%%%%%%
%\bfsffff*{New method to quickly visually assess the effect of individual terms.}
%%%%%%%%%%%%%%%%%%%%%%%%%%%%%%%%%%%%%%%%%%%%%%%%%%%%%%%%%%%%%%%%%%%%%%%%
\noindent
{\bf{\sffamily{Constitutive\,Artificial\,Neural\,Networks 
simultaneously learn both model and parameters.}}}
For decades, chemical, physical, and material scientists alike have been modeling the hyperelastic response of rubber under large deformations \cite{james43,mooney40,treloar44,blatz62,ogden72}. They have proposed numerous competing constitutive models to best characterize the behavior of artificial and biological polymers and calibrated the model parameters in response to different modes of mechanical loading  \cite{boyce88,dal09,demiray72,gent96,holzapfel96,holzapfel00,miehe04,reese98,steinmann12}. Here we propose a radically different approach towards constitutive modeling and abandon the common strategy to {\it{first}} select a constitutive model and {\it{then}} tune its parameters by fitting the model to data. Instead, we propose a family of Constitutive Artificial Neural Networks that {\it{simultaneously learn both}} the constitutive model and its material parameters.\\[6.pt]
%%%
{\bf{\sffamily{Classical Neural Networks ignore the underlying physics.}}}
In the most general form, constitutive equations in solid mechanics are tensor-valued tensor functions that define a second order stress tensor, in our case the Piola stress, as a function of a second order deformation or strain measure, in our case the deformation gradient \cite{truesdellnoll65,truesdell69}. Classical Neural Networks are universal function approximators that learn these functions \cite{mcculloch43}, in our case the stress, from training data, in our case experimentally measured stress-strain data, by minimizing a loss function, in our case the mean squared error between model and data stress. Neural Networks have advanced as a powerful technology to interpolate or describe big data; yet, they fail to extrapolate or predict scenarios beyond the training regime \cite{alber19}. They are an excellent choice when we have no information about the underlying data, but in constitutive modeling, they entirely ignore our prior knowledge and thermodynamic considerations \cite{peng21}. \\[6.pt]
%%%
{\bf{\sffamily{Constitutive\,Artificial\,Neural\,Networks 
include kinematical, thermodynamical, and physical constraints.}}}
The general idea of this manuscript is to design a new family of Neural Networks that inherently satisfy common kinematical, thermodynamical, and physical constrains while, at the same time, constraining the design space of all admissible functions to make the network robust and reliable, even in the presence of small training data. Our approach is to {\it{reverse-engineer}} Constitutive Artificial Neural Networks that are, by design, a generalization of widely used and commonly accepted constitutive models with well-defined physical parameters \cite{mahnken22,steinmann12}. Towards this goal we revisit the non-linear field theories of mechanics \cite{antman05,truesdellnoll65,truesdell69}
and suggest to constrain 
the network {\it{output}} to enforce thermodynamic consistency;
the network {\it{input}} to enforce material objectivity, and, if desired, material symmetry and incompressibility;
the {\it{activation functions}} to implement physically reasonable constitutive restrictions; and the network {\it{architecture}} to ensure polyconvexity. \\[6.pt]
%%%
{\bf{\sffamily{Constitutive\,Artificial\,Neural\,Networks 
are a generalization of popular constitutive models.}}}
We prototype the design of Constitutive Artificial Neural Networks for the example of an isotropic perfectly incompressible feed forward network with two hidden layers and twelve weights that takes the scalar-valued first and second invariants of the deformation gradient, $[\,I_1-3\,]$ and $[\,I_2-3\,]$, as input and approximates the scalar-valued free energy function, $\psi(I_1,I_2)$, as output. The first layer generates the first and second powers, $(\,\circ\,)$ and $(\,\circ\,)^2$, of the input and the second layer applies the identity and the exponential, $(\,\circ\,)$ and $(\rm{exp}(\alpha (\circ))-1)$, to these powers. This results in eight individual subfunctions that additively feed into the final free energy function $\psi$ from which we derive the Piola stress, $\ten{P} = \partial \psi / \partial \ten{F}$, following standard arguments of thermodynamics. We demonstrate that the approximated free energy function of our network is a {\it{generalization}} of popular constitutive models with the neo Hooke \cite{treloar48}, Blatz Ko \cite{blatz62}, Mooney Rivlin \cite{mooney40,rivlin48}, Yeoh \cite{yeoh93}, and Demiray \cite{demiray72} models as special cases. Most importantly, through a direct comparison with these models, the twelve weights of the network gain a clear physical interpretation.\\[6.pt]
%%%
{\bf{\sffamily{Classical Neural Networks can interpolate robustly, but fail to extrapolate.}}}
In a side-by-side comparison with a classical Neural Network, we demonstrate the features of our new Constitutive Artificial Neural Network for several classical benchmark data sets for rubber in uniaxial tension \cite{blatz62,mooney40,treloar44}, equibiaxial tension \cite{treloar44}, and pure shear \cite{treloar44}. Both methods robustly identify functions that approximate the data well and reduce the error between model and data within less than 10,000 epochs: The classical Neural Network, without any prior knowledge of the underlying physics, directly learns the stress as a function of the deformation gradient, $\ten{P}(\ten{F})$, while the Constitutive Artificial Neural Network learns the free energy as a function of the strain invariants, $\psi(I_1,I_2)$. Our results in Figure~\ref{fig07} support the general notion that classical Neural Networks {\it{describe}} or {\it{interpolate}} data well, but cannot {\it{predict}} or {\it{extrapolate}} the behavior outside the training regime \cite{alber19}. We also confirm in Figure \ref{fig08} that they {\it{fit big data well}}, but tend to {\it{overfit sparse data}} \cite{peng21}. To quickly assess the importance of the individual nodes, we color-code their outputs and visually compare their contributions to the final output layer. From the color spectrum in Figure \ref{fig09}, we conclude that classical Neural Networks tend to activate all nodes of the final layer with non-zero weights, but that these weights have no physical meaning and do not contribute to {\it{interpret}} or {\it{explain}} the underlying physics. \\[6.pt]
%%%
{\bf{\sffamily{Constitutive\,Artificial\,Neural\,Networks 
robustly learn both model and parameters,\,even for sparse data.}}}
Our new family of Constitutive Artificial Neural Network addresses the limitations of conventional classical Neural Networks by including thermodynamic considerations by design. Figure \ref{fig10} suggests that they are both {\it{descriptive}} and {\it{predictive}}, {\it{without overfitting}} the data. From the reduced color spectra in Figures \ref{fig11} and \ref{fig12}, we conclude that our networks self-select subsets of activation functions, while most of their weights remain zero. Figure \ref{fig12} also shows that, for insufficiently rich data, the network still approximates the overall function $\psi(I_1,I_2)$ robustly, but the distribution of the individual contributions of the $I_1$ and $I_2$ terms is non-unique. Enriching the training data by multi-mode data from uniaxial tension, equibiaxial tension, and pure shear in Figure \ref{fig13} eliminates these non-uniqueness. This suggests that, when trained with sufficiently rich data, Constitutive Artificial Neural Networks {\it{simultaneously learn both a unique model and parameter set}}. \\[6.pt]
%%%
{\bf{\sffamily{Constitutive Artificial Neural Networks enable automated model discovery.}}}
For the example of rubber in the high and low temperature regimes, our new Constitutive Artificial Neural Network discovers two three-term models in terms of the first and second invariants, 
$\frac{1}{2} \mu_2 \,[\,I_1-3\,]$ and 
$\frac{1}{2} \mu_2 \,[\,I_2-3\,]$, similar to the classical Mooney Rivlin model \cite{mooney40,rivlin48}, and in terms of their linear exponentials, 
$\frac{1}{2} a_1 [\,\rm{exp}(b_1\,[I_1-3])-1]/b_1$ and 
$\frac{1}{2} a_2 [\,\rm{exp}(b_2\,[I_2-3])-1]/b_2$, 
similar to the Demiray model \cite{demiray72}. 
The non-zero network weights take the interpretation of the shear moduli, $\mu_1$ and $\mu_2$, stiffness-like parameters, $a_1$ and $a_2$, and exponential coefficients, $b_1$ and $b_2$ of these models. Since the network autonomously self-selects the model and parameters that best approximate the data, the human user no longer needs to decide which model to choose. This could have enormous implications, for example in finite element simulations: Instead of selecting a specific material model from a library of available models, finite element solvers could be built around a single generalized model, the Constitutive Artificial Neural Network {\it{autonomously discovers the model}} from data, populates the model parameters, and activates the relevant terms.\\[6.pt]
%%%
{\bf{\sffamily{Current limitations and future applications.}}}
In the present work, we have shown the application of Constitutive Artificial Neural Networks for the special case of perfectly incompressible isotropic materials according to Figure \ref{fig05}. It is easy to see that the general concept in Figure \ref{fig03} extends naturally to {\it{compressible}} or {\it{nearly incompressible}} materials with other symmetry classes,  {\it{transversely isotropic}} or {\it{orthotropic}}, simply by expanding the network input to other sets of strain invariants. A more involved extension would be to consider {\it{history-dependent, inelastic}} materials, for example by replacing the feed forward architecture through a {\it{long short-term memory network}} with feedback connections \cite{bhouri21}, while still keeping the same overall network input, output, activation functions, and basic architecture. In parallel, we could revisit the network architecture in Figure \ref{fig03} by expressing the free energy as a truncated infinite series of products of powers of the invariants, which would result in a {\it{fully connected feed forward}} network architecture. One limitation we foresee for these more complex networks, is that the majority of weights might no longer train to zero. If the network learns a large set of non-zero weights, and with them, activates too many terms that feed into the final free energy function, we could reduce the model to the most relevant terms by {\it{network pruning}}, a neurologically inspired process in which the network gradually self-eliminates less relevant connections from its overall architecture \cite{budday15}. Of course, we could also always enforce certain weights to zero, recover a popular subclasses of models, and use the Constitutive Artificial Neural Network for a plain inverse analysis and parameter identification. Finally, one important extension would be to embed the network in a Bayesian framework to supplement the analysis with {\it{uncertainty quantification}}. Instead of simple point estimates for the network parameters, a Bayesian Constitutive Artificial Neural Network would learn parameter distributions with means and credible intervals. In contrast to classical Bayesian Neural Networks, here, these distributions would have a clear physical interpretation, since our network weights have a well-defined physical meaning. 
%%%%%%%%%%%%%%%%%%%%%%%%%%%%%%%%%%%%%%%%%%%%%%%%%%%%%%%%%%%%%%%%%%%%%%%%
\section{Conclusion}\label{sec_conclusion}
%%%%%%%%%%%%%%%%%%%%%%%%%%%%%%%%%%%%%%%%%%%%%%%%%%%%%%%%%%%%%%%%%%%%%%%%
\noindent
Constitutive Artificial Neural Networks are a new family of neural networks that 
satisfy kinematical, thermodynamical, and physical constraints by design, and, at the same time, constrain the space of admissible functions to train robustly, even for space data. In contrast to classical Neural Networks, they can describe, predict, and explain data and reduce the risk of overfitting. Constitutive Artificial Neural Networks integrate more than a century of knowledge in continuum mechanics and modern machine learning to create Neural Networks with specific network input, output, activation functions, and architecture to a priori guarantee thermodynamic consistency, material objectivity, material symmetry, physical restrictions, and polyconvexity. The resulting network is a generalization of widely used popular constitutive models with network weights that have a clear physical interpretation. When trained with sufficiently rich data, Constitutive Artificial Neural Networks can simultaneously learn both a unique model and set of parameters, while most of the network weights train to zero. This suggests that Constitutive Artificial Neural Networks have the potential to enable automated model discovery and could induce a paradigm shift in constitutive modeling, from user-defined to automated model selection and parameterization. 
%%%%%%%%%%%%%%%%%%%%%%%%%%%%%%%%%%%%%%%%%%%%%%%%%%%%%%%%%%%%%%%%%%%
%{\bf{Features:}}\\
%+ network is both descriptive {\it{and}} predictive \\
%+ inherently satisfies kinematical, thermodynamical, and physical constraints \\
%+ allows us to quickly visually assess the effect of individual terms \\
%+ has common constitutive models as special cases \\
%+ method autonomously an simultaneously self-selects the model {\it{and}} the parameters that best explain the data\\
%+ method is robust and stable, learns a unique stress function, and, for sufficiently rich data, also a unique model and parameter set \\[6.pt]
%%%
%{\bf{Show:}} If we have such a generalized network and we train it (with pruning, where we successively delete connections with small weights), do we recover the Holzapfel model for arterial data, the Gent model for rubber data etc? That would be super cool! \\[6.pt]
%{\bf{Comment on... possible future directions...}}\\
%physics constraint $>$ constrain parameters to remain positive \\
%value terms evenly $>$ data weighting \\
%systematic model evaluation, compare parameters to literature $>$ paul \\
%%%%%%%%%%%%%%%%%%%%%%%%%%%%%%%%%%%%%%%%%%%%%%%%%%%%%%%%%%%%%%%%%%%
\section*{Data availability}
\noindent
Our source code, data, and examples are available at 
https:/\!/github.com/LivingMatterLab/CANN.

%%%%%%%%%%%%%%%%%%%%%%%%%%%%%%%%%%%%%%%%%%%%%%%%%%%%%%%%%%%%%%%%%%%
\section*{Acknowledgments}
\noindent
We thank Oliver Weeger, Wolfgang Ehlers, and Paul Steinmann for asking the right questions
and Greg Bronevetsky, Serdar G\"oktepe, and Andreas Menzel for helping us find the right answers.
This work was supported 
by a DAAD Fellowship to Kevin Linka and
by the Stanford School of Engineering Covid-19 Research and Assistance Fund and Stanford Bio-X IIP seed grant to Ellen Kuhl.
%%%%%%%%%%%%%%%%%%%%%%%%%%%%%%%%%%%%%%%%%%%%%%%%%%%%%%%%%%%%%%%%%%%
\begin{small}

\end{small}
%%%%%%%%%%%%%%%%%%%%%%%%%%%%%%%%%%%%%%%%%%%%%%%%%%%%%%%%%%%%%%%%%%%
%%%%%%%%%%%%%%%%%%%%%%%%%%%%%%%%%%%%%%%%%%%%%%%%%%%%%%%%%%%%%%%%%%%
\end{document}